\documentclass[acmtog]{acmart}
\usepackage{amsmath}
\usepackage{booktabs} 
\usepackage{caption}
\usepackage{subcaption}
\usepackage{graphicx}
\usepackage{tikz}
\usepackage{tabularx}
\usepackage{hyperref}
\usetikzlibrary{spy}

\usepackage[ruled]{algorithm2e} 

\SetAlFnt{\small}
\SetAlCapFnt{\small}
\SetAlCapNameFnt{\small}
\SetAlCapHSkip{0pt}

\hypersetup{draft}   

\acmJournal{TOG}

\acmVolume{42} 
\acmNumber{4} 
\acmArticle{1}
\acmMonth{8} 
\acmPrice{15.00}

\begin{document}
\title{Pyramid Texture Filtering}

\author{Qing Zhang}
\orcid{0000-0001-5312-2800}
\affiliation{%
  \institution{Sun Yat-sen University}
  \city{Guangzhou}
  \country{China}}
\email{zhangqing.whu.cs@gmail.com}

\author{Hao Jiang}
\orcid{0009-0002-6398-2323}
\affiliation{%
	\institution{Sun Yat-sen University}
	\city{Guangzhou}
	\country{China}}
\email{jiangh69@mail2.sysu.edu.cn}

\author{Yongwei Nie}
\orcid{0000-0002-8922-3205}
\affiliation{%
	\institution{South China University of Technology}
	\city{Guangzhou}
	\country{China}}
\email{nieyongwei@scut.edu.cn}

\author{Wei-Shi Zheng} 
\authornote{Corresponding author.}
\orcid{0000-0001-8327-0003}
\affiliation{%
	\institution{Sun Yat-sen University}
	\city{Guangzhou}
	\country{China}}
\email{wszheng@ieee.org}

\begin{abstract}
We present a simple but effective technique to smooth out textures while preserving the prominent structures. Our method is built upon a key observation---the coarsest level in a Gaussian pyramid often naturally eliminates textures and summarizes the main image structures. This inspires our central idea for texture filtering, which is to progressively upsample the very low-resolution coarsest Gaussian pyramid level to a full-resolution texture smoothing result with well-preserved structures, under the guidance of each fine-scale Gaussian pyramid level and its associated Laplacian pyramid level. We show that our approach is effective to separate structure from texture of different scales, local contrasts, and forms, without degrading structures or introducing visual artifacts. We also demonstrate the applicability of our method on various applications including detail enhancement, image abstraction, HDR tone mapping, inverse halftoning, and LDR image enhancement. Code is available at \url{https://rewindl.github.io/pyramid_texture_filtering/}.
\end{abstract}

%
%



\begin{teaserfigure}
	\centering
	\begin{subfigure}[c]{0.196\textwidth}
		\centering
		\includegraphics[width=1.38in]{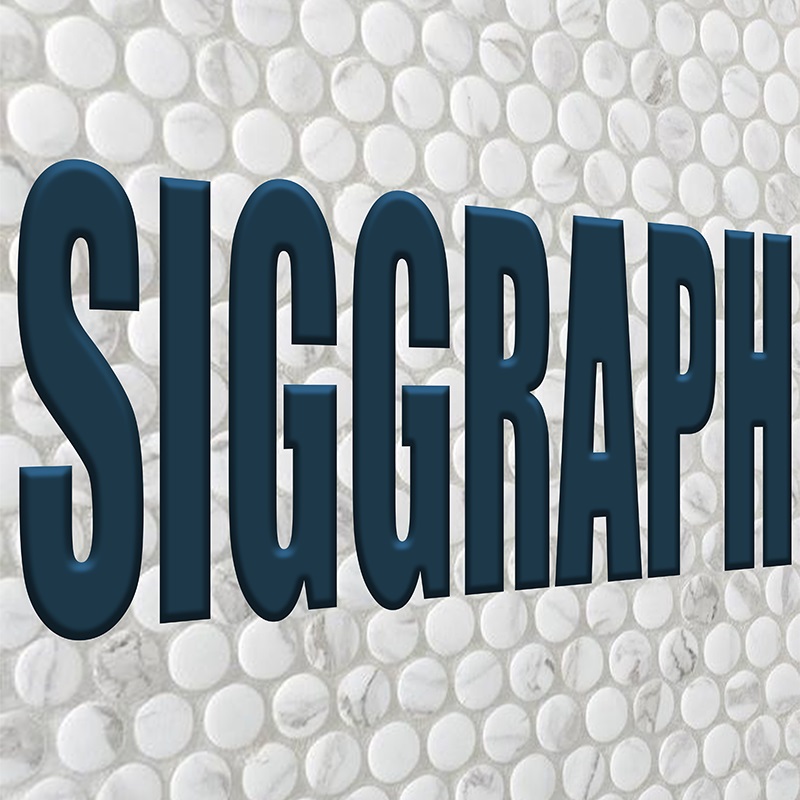}  \\ \vspace{2pt}
		\includegraphics[width=1.38in]{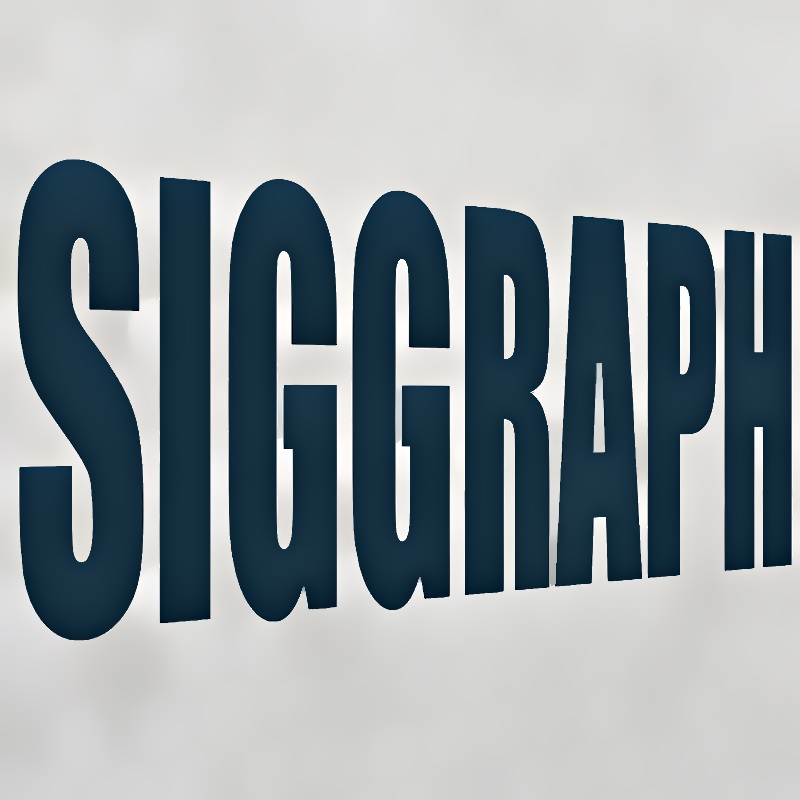}
	\end{subfigure}
	\begin{subfigure}[c]{0.196\textwidth}
		\centering
		\includegraphics[width=1.38in]{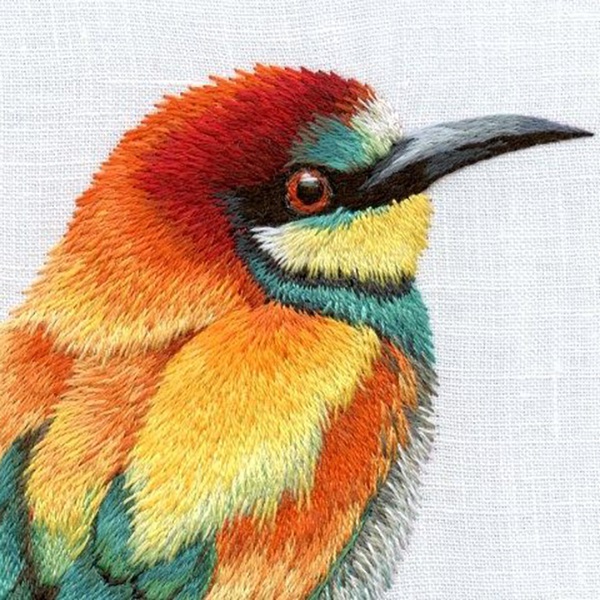}  \\ \vspace{2pt}
		\includegraphics[width=1.38in]{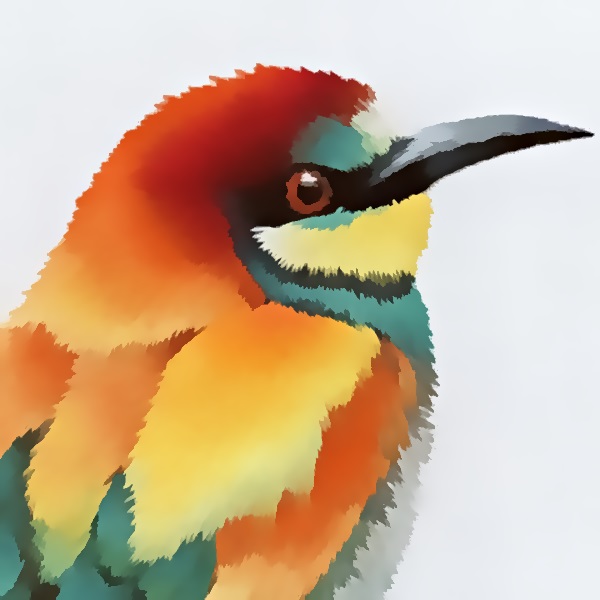}
	\end{subfigure}
	\begin{subfigure}[c]{0.196\textwidth}
		\centering
		\includegraphics[width=1.38in]{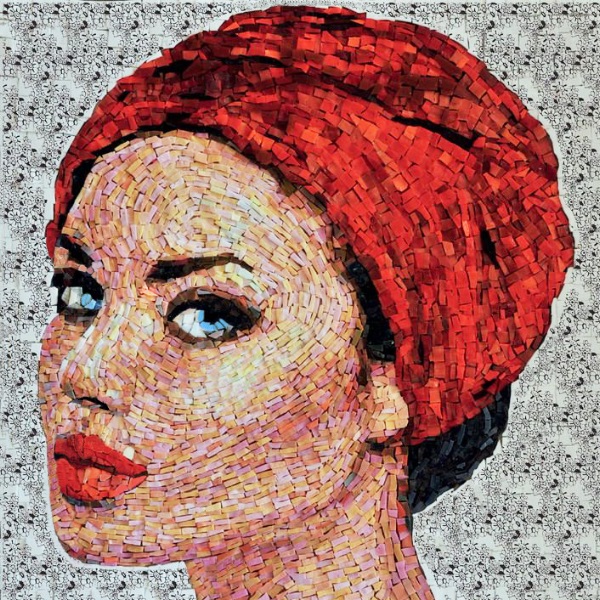}  \\ \vspace{2pt}
		\includegraphics[width=1.38in]{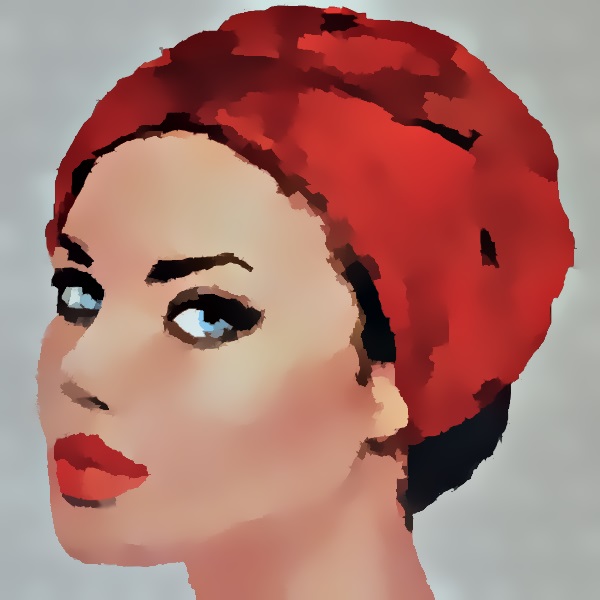}
	\end{subfigure}
	\begin{subfigure}[c]{0.196\textwidth}
		\centering
		\includegraphics[width=1.38in]{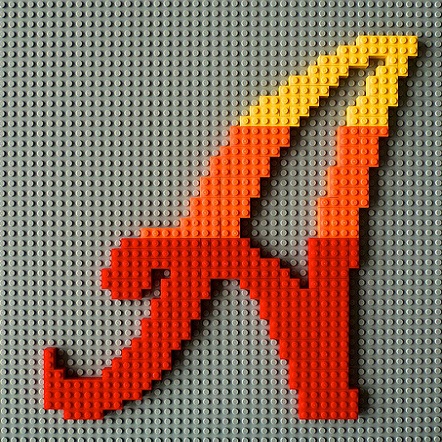}  \\ \vspace{2pt}
		\includegraphics[width=1.38in]{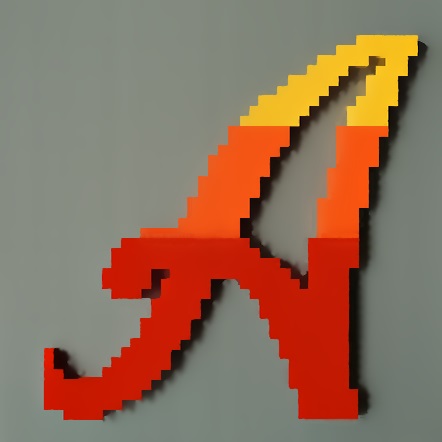}
	\end{subfigure}
	\begin{subfigure}[c]{0.196\textwidth}
		\centering
		\includegraphics[width=1.38in]{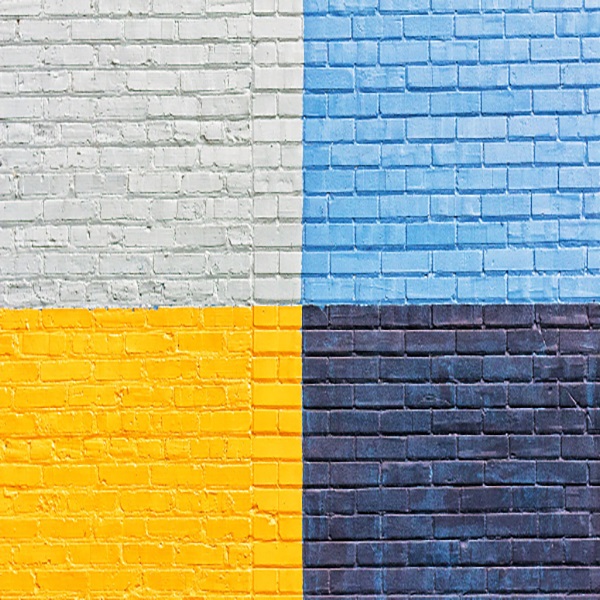}  \\ \vspace{2pt}
		\includegraphics[width=1.38in]{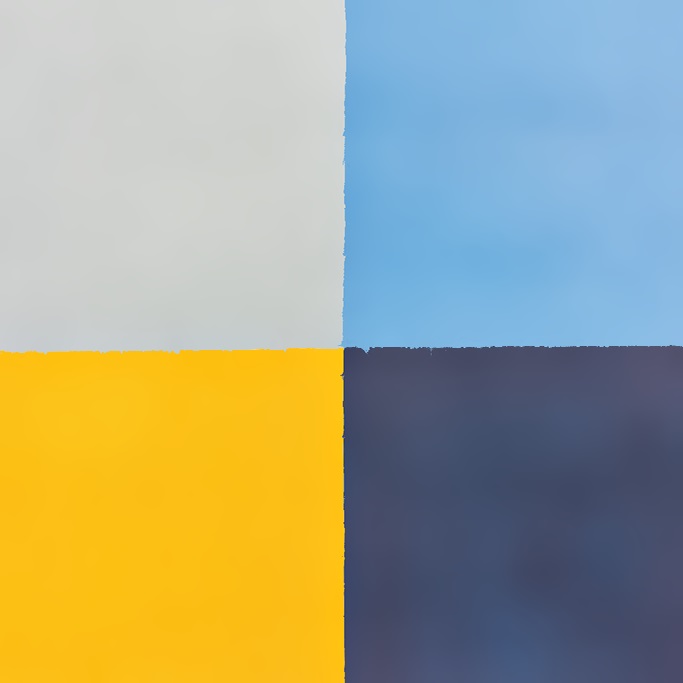}
	\end{subfigure}
	\vspace{-2mm}
	\caption{We demonstrate texture filtering (also referred to as structure-preserving filtering) based on Gaussian and Laplacian pyramids, which, unlike previous work, does not rely on any explicit measures to distinguish texture from structure, but can effectively deal with diverse types of textures. Top: input images. Bottom: our results. The 2nd-5th images are from TrishBurr, Simona Proto Art, \cite{xu2012structure}, and Flickr user Kyle Eroche. } 
\label{fig:teaser}
\end{teaserfigure}

\maketitle

\section{Introduction}

Texture smoothing is a fundamental problem in computational photography and image analysis. It aims to remove the distracting fine-scale textures while maintaining the large-scale image structures. This smoothing operation has received considerable research attention, since it not only benefits scene understanding but also enables a wide variety of image manipulation applications such as image abstraction, detail enhancement, and HDR tone mapping.

Despite being widely studied, texture smoothing remains a challenge as it requires to distinguish texture from structure, which is a non-trivial problem. The reason is that texture and structure are often very similar in terms of basic visual elements such as intensity, gradient, and local contrast, making the two indistinguishable. It is notable that, although closely related and conceptually similar, edge-preserving smoothing approaches \cite{tomasi1998bilateral,farbman2008edge,fattal2009edge,kass2010smoothed,paris2011local,xu2011image,gastal2011domain,bi20151,liu2021generalized} are not optimal solutions for texture smoothing, because their goal is to retain salient edges regardless of whether they are originated from texture or structure components.

Previous texture smoothing methods primarily formulate filtering or optimization frameworks based on hand-crafted texture-structure separation measures \cite{subr2009edge,xu2012structure,karacan2013structure,bao2013tree,cho2014bilateral,zhang2015segment}. However, they may struggle when presented with textures that cannot be effectively distinguished by their measures, or tend to remove textures at the cost of degrading image structures. More recently, some learning-based approaches that enable texture removal have also been proposed \cite{lu2018deep,fan2018image}, but they may not generalize well to textures not present in their training datasets.

In this paper, we present a novel texture smoothing method that can effectively remove different types of textures while preserving the main structures. Unlike previous methods that focus on designing texture-structure separation measures based on certain local image statistics, we argue that the most discriminative difference between texture and structure is the scale, since we observe that the coarsest Gaussian pyramid level representing large-scale image information often naturally eliminate textures without destroying structures (see Figure~\ref{fig:observation}). This observation inspires us to perform texture smoothing by upsampling the coarsest Gaussian pyramid level, which is, however, cannot accomplished by existing upsampling methods as they are unable to produce result with sharp structures from a very low-resolution input (see Figure~\ref{fig:ablation}(b) and (c)). The key to the success of our method is the proposed pyramid-guided structure-aware upsampling, which is iteratively performed under the guidance of each fine-scale Gaussian pyramid level and its associated Laplacian pyramid level, until the coarsest level is upsampled to a full-resolution texture smoothing result. 

As shown in Figure~\ref{fig:teaser}, our method, while very simple in key idea, is surprisingly effective in texture removal and can faithfully preserve image structures without introducing visual artifacts. Besides, it is very easy to implement and allows to rapidly produce results. We also demonstrate that it enables various image manipulation applications. Our main contributions are summarized as follows:
\begin{itemize}
	\item We find that the coarsest level in a Gaussian pyramid usually eliminates textures while preserving the main image structures, providing new cues for texture smoothing.
	\item We present a novel pyramid-based texture smoothing approach based on the above finding, which works by progressively upsampling the coarsest Gaussian pyramid level of a given image to the original full-resolution, without relying on any texture-structure separation measures.
	\item We develop pyramid-guided structure-aware upsampling to produce texture smoothing result with sharp structures from the coarsest Gaussian pyramid level in very low-resolution.
\end{itemize}

\section{Related Work}
\subsection{Edge-Preserving Image Smoothing}
The past decades have witnessed a wealth of work on edge-preserving smoothing. Among them, a large number of methods belong to the category of local filtering, the core idea of which is to perform weighted average over a local spatial neighborhood. Popular representatives in this category include anisotropic diffusion \cite{perona1990scale}, bilateral filter \cite{tomasi1998bilateral} and its fast approximations \cite{durand2002fast,paris2006fast,weiss2006fast,chen2007real}, edge-avoiding wavelets \cite{fattal2009edge}, local histogram filter \cite{kass2010smoothed}, geodesic filters \cite{criminisi2010geodesic,gastal2011domain,gastal2012adaptive}, and local Laplacian filer (LLF) \cite{paris2011local}. Note that although both our method and LLF utilize image pyramids, we aim to achieve texture smoothing by upsampling the coarsest level in Gaussian pyramid, while the goal of LLF is to enable edge-aware filtering through manipulation of Laplacian pyramid coefficients.

\begin{figure}
	\centering
	\begin{subfigure}[c]{0.156\textwidth}
		\centering
		\begin{tikzpicture}[
		spy using outlines={color=white, rectangle, magnification=3,
			every spy on node/.append style={rectangle},
			every spy in node/.append style={rectangle}}
		]
		\node[inner sep=0,outer sep=0]{\includegraphics[width=1.1in]{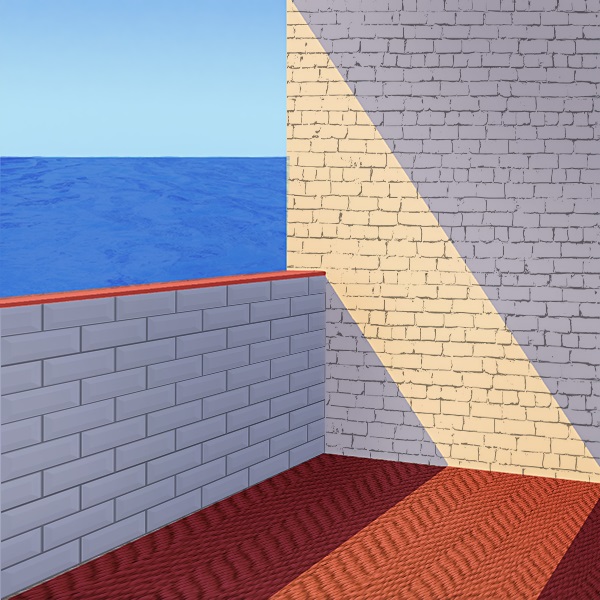}};
		\spy[width=0.6cm, height=0.6cm] on (0.1, -0.5) in node at (-1.06, 1.06);
		\end{tikzpicture} \\
		\vspace{-2mm}
		\caption{$G_0$ (Original image)}
	\end{subfigure}
	\begin{subfigure}[c]{0.156\textwidth}
		\centering
		\begin{tikzpicture}[
		spy using outlines={color=white, rectangle, magnification=3,
			every spy on node/.append style={rectangle,size=0cm},
			every spy in node/.append style={rectangle}}
		]
		\node[inner sep=0,outer sep=0]{\includegraphics[width=1.1in]{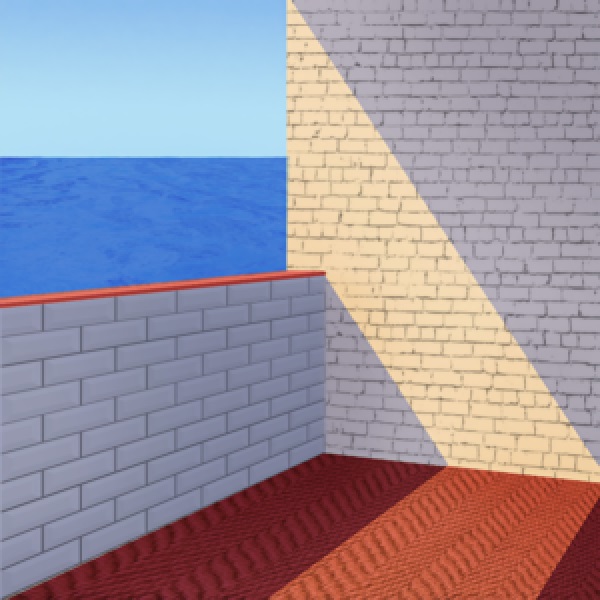}};
		\spy[width=0.6cm, height=0.6cm] on (0.1, -0.5) in node at (-1.06, 1.06);
		\end{tikzpicture} \\
		\vspace{-2mm}
		\caption{$G_1$ (1/2 resolution)}
	\end{subfigure}
	\begin{subfigure}[c]{0.156\textwidth}
		\centering
		\begin{tikzpicture}[
		spy using outlines={color=white, rectangle, magnification=3,
			every spy on node/.append style={rectangle,size=0cm},
			every spy in node/.append style={rectangle}}
		]
		\node[inner sep=0,outer sep=0]{\includegraphics[width=1.1in]{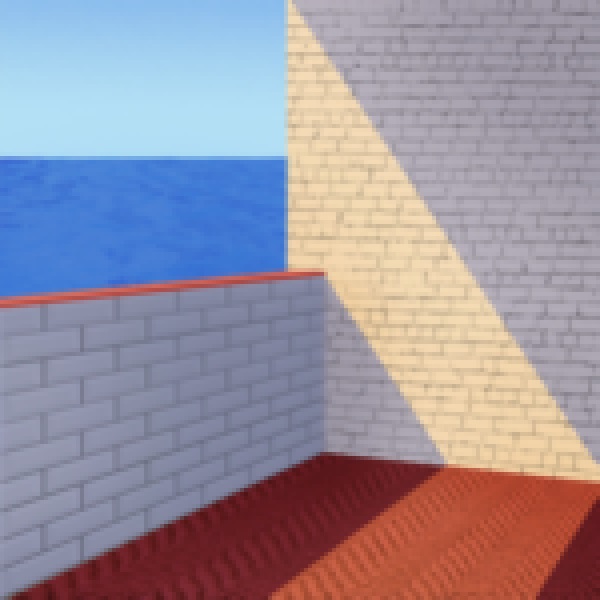}};
		\spy[width=0.6cm, height=0.6cm] on (0.1, -0.5) in node at (-1.06, 1.06);
		\end{tikzpicture} \\
		\vspace{-2mm}
		\caption{$G_2$ (1/4 resolution)}
	\end{subfigure} \\ \vspace{1mm}
	\begin{subfigure}[c]{0.156\textwidth}
		\centering
		\begin{tikzpicture}[
		spy using outlines={color=white, rectangle, magnification=3,
			every spy on node/.append style={rectangle,size=0cm},
			every spy in node/.append style={rectangle}}
		]
		\node[inner sep=0,outer sep=0]{\includegraphics[width=1.1in]{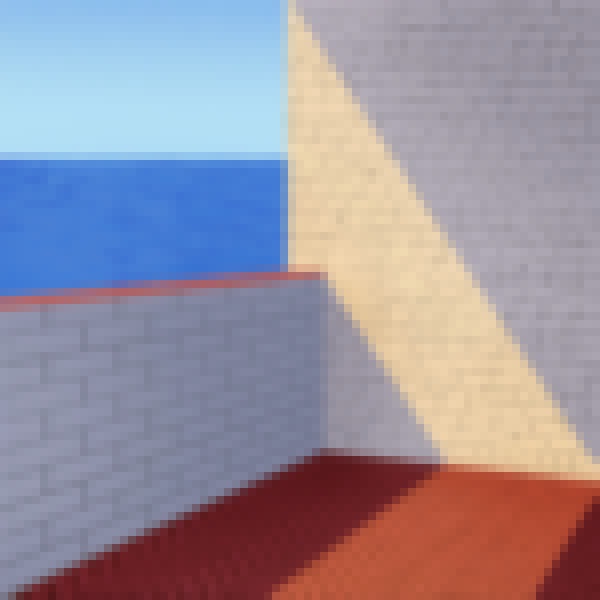}};
		\spy[width=0.6cm, height=0.6cm] on (0.1, -0.5) in node at (-1.06, 1.06);
		\end{tikzpicture} \\
		\vspace{-2mm}
		\caption{$G_3$ (1/8 resolution)}
	\end{subfigure}
	\begin{subfigure}[c]{0.156\textwidth}
		\centering
		\begin{tikzpicture}[
		spy using outlines={color=white, rectangle, magnification=3,
			every spy on node/.append style={rectangle,size=0cm},
			every spy in node/.append style={rectangle}}
		]
		\node[inner sep=0,outer sep=0]{\includegraphics[width=1.1in]{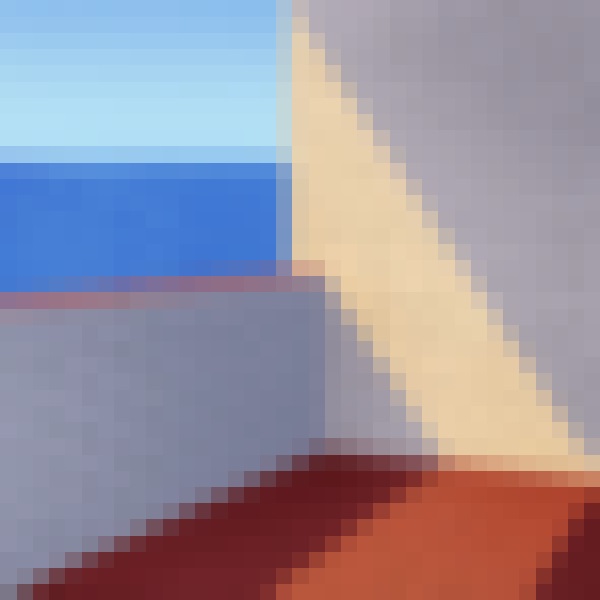}};
		\spy[width=0.6cm, height=0.6cm] on (0.1, -0.5) in node at (-1.06, 1.06);
		\end{tikzpicture} \\
		\vspace{-2mm}
		\caption{$G_4$ (1/16 resolution)}
	\end{subfigure}
	\begin{subfigure}[c]{0.156\textwidth}
		\centering
		\begin{tikzpicture}[
		spy using outlines={color=white, rectangle, magnification=3,
			every spy on node/.append style={rectangle,size=0cm},
			every spy in node/.append style={rectangle}}
		]
		\node[inner sep=0,outer sep=0]{\includegraphics[width=1.1in]{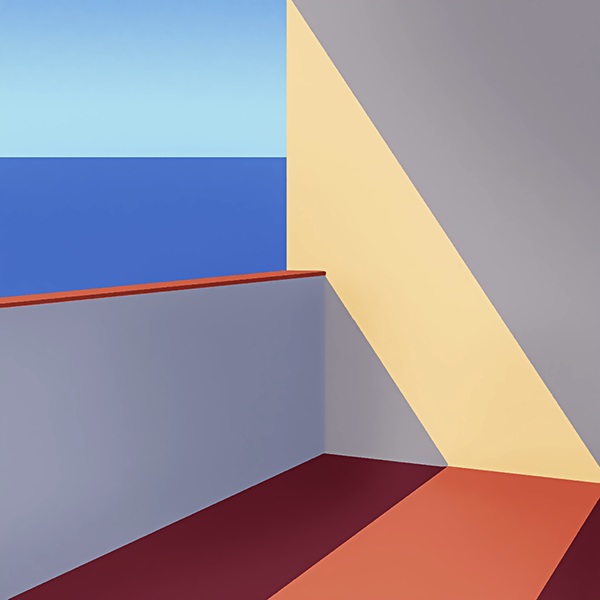}};
		\spy[width=0.6cm, height=0.6cm] on (0.1, -0.5) in node at (-1.06, 1.06);
		\end{tikzpicture} \\
		\vspace{-2mm}
		\caption{Ground truth}
	\end{subfigure}
	\vspace{-2mm}
	\caption{Our main observation. Given a Gaussian pyramid $\{G_{0},...,G_4\}$ showing all levels at original full-resolution ($600 \times 600$), we find that the coarsest level $G_4$ naturally eliminates textures while maintaining the main image structures. Note, to obtain the texture smoothing ground truth, the original image here is synthesized by adding textures to an existing structure-only image (i.e., the ground-truth). Image courtesy of Kieran Gabriel Prints.} 
	\label{fig:observation}
\end{figure}

Optimization-based methods, such as weighted least squares (WLS) \cite{farbman2008edge} and its more efficient variants \cite{min2014fast,liu2017semi,liu2020real}, $L_0$ gradient minimization \cite{xu2011image}, $L_1$ image transform \cite{bi20151}, are also commonly adopted solutions for edge-preserving smoothing. However, most of them involve relatively high computational complexity because of requiring solving large-scale linear systems, which limits their scalability and increases the difficulty of implementation.

\begin{figure*}
	\centering
	\includegraphics[width=6.9in]{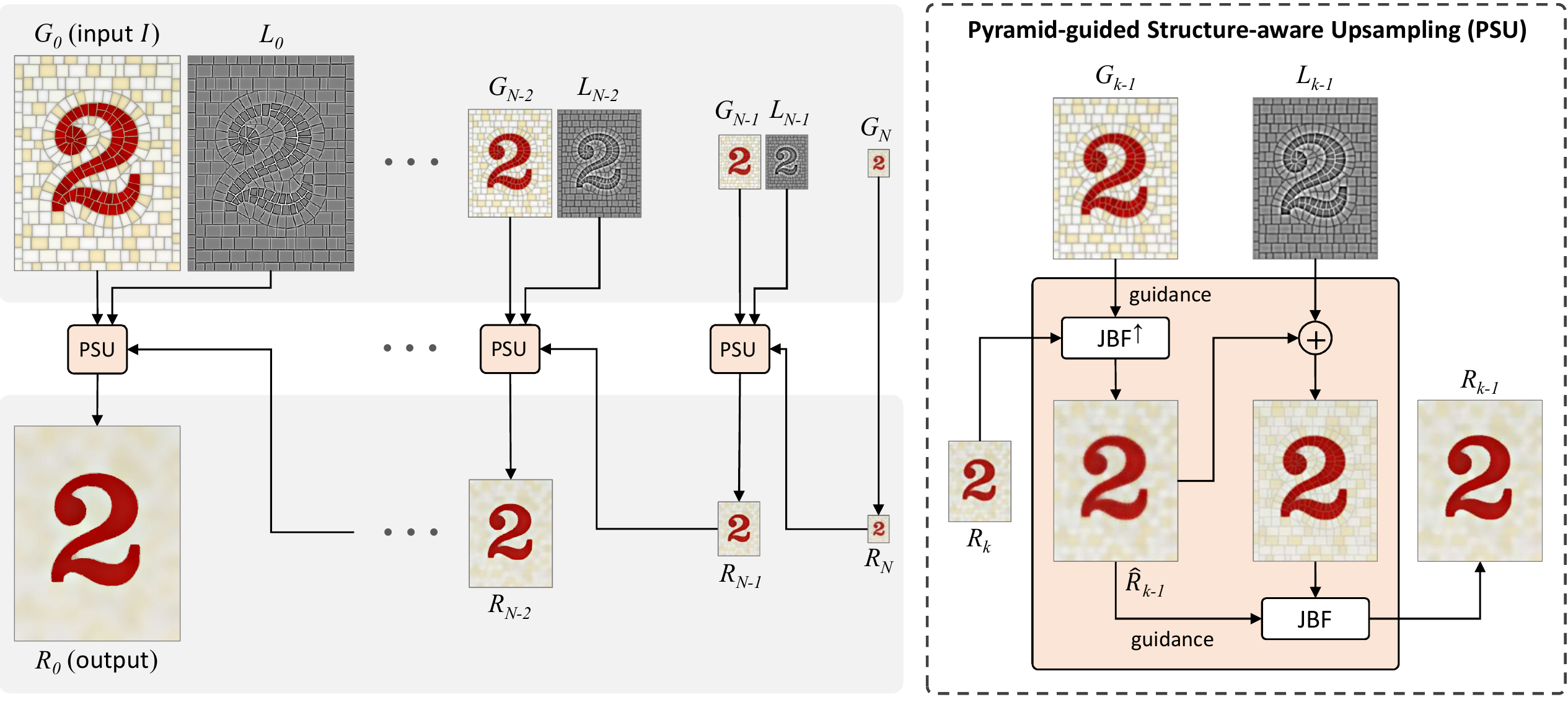}\\
    \vspace{-3mm}
	\caption{Overview of our approach. Given an input image $I$, we first build its Gaussian and Laplacian pyramids $\{G_{\ell}\}$ and $\{L_{\ell}\}$. Next, we upsample the coarsest Gaussian pyramid level $G_N$ ($G_N = R_N$) to an intermediate texture smoothing image $R_{N-1}$ at the previous finer scale. This is achieved by a pyramid-guided structure-aware upsampling (PSU) taking $G_{N-1}$ and $L_{N-1}$ as guidance. The resulting $R_{N-1}$ is then subjected to the same upsampling process guided by $G_{N-2}$ and $L_{N-2}$. The above upsampling cycle is repeated multiple times until a full-resolution texture smoothing image $R_0$ is finally obtained. $\textrm{JBF}$ refers to joint bilateral filtering, and the symbol $\uparrow$ in $\textrm{JBF}^{\uparrow}$ indicates increase in spatial resolution of the output. Image courtesy of Flickr user Mighty Tieton.} 
	\label{fig:overview}
\end{figure*}

Recently, deep learning has also been introduced to the field of edge-aware smoothing \cite{xu2015deep,liu2016learning,fan2018image}. As it is very difficult to obtain ground-truth smoothing results, current deep-learning-based methods either use results produced by existing smoothing algorithms as supervision, or apply deep network as an optimization solution.

\subsection{Texture Smoothing}
Despite the great success of edge-preserving smoothing research, texture smoothing was also advocated and widely studied, since the overall image structures, rather than fine-scale textures, are assumed to be crucial to visual perception. As the major difficulty of the problem lies in how to effectively distinguish texture from structure, most existing methods are built upon specially designed texture-structure separation measures, e.g., local extrema \cite{subr2009edge}, relative total variation \cite{xu2012structure,cho2014bilateral}, region covariance based patch similarity \cite{karacan2013structure,zhu2016non}, minimum spanning tree \cite{bao2013tree}, and segment graph \cite{zhang2015segment}, or rely on auxiliary contour information as guidance \cite{wei2018joint}. Besides, there are some scale-aware texture smoothing methods \cite{zhang2014rolling,du2016two,jeon2016scale} based on Gaussian filtering, while more recent works are mostly deep learning based \cite{lu2018deep,kim2018structure}.

Our work is complementary to existing texture smoothing methods in three aspects. First, we reveal that the multi-scale representations provided by standard image pyramids can be effective cues for texture smoothing. Second, we present a novel pyramid-based texture smoothing approach that neither relies on any texture-structure separation measures, nor requires additional contour information as input. Third, we show that our approach outperforms previous methods in terms of structure preservation and texture removal, and is effective to deal with previously challenging images with large-scale or high-contrast textures.

\subsection{Image Pyramids}
Image pyramids are useful multi-resolution representations for analyzing and manipulating images over a range of spatial scales \cite{burt1983laplacian}. Here we briefly describe Gaussian and Laplacian pyramids as we build our approach on top of them. Given an image $I$, its Gaussian pyramid is a set of low resolution versions $\{G_{\ell}\}$ (called levels) of the image in which small-scale image details gradually disappear. To construct $\{G_{\ell}\}$, the original image $I$, i.e., the finest level $G_0$, is repeatedly smoothed by a Gaussian filter and then subsampled to generate the sequence of levels, until a minimum resolution is reached at the coarsest level $G_N$. Laplacian pyramid is created from the Gaussian pyramid, with the goal of capturing image details that are present in one Gaussian pyramid level, but not present at the following coarser level. Specifically, levels of the Laplacian pyramid are defined as the differences between the successive levels of the Gaussian pyramid, $L_{\ell} = G_{\ell} - \textrm{upsample}(G_{\ell+1})$, where $\textrm{upsample}(\cdot)$ is an operation that upscales the resolution of a Gaussian pyramid level to that of its previous finer level. The last level $L_N$ in the Laplacian pyramid is not a difference image, but exactly $G_N$, based on which the original image $I$ can be reconstructed by recursively applying $G_{\ell} = L_{\ell} + \textrm{upsample}(G_{\ell+1})$ 
to collapse the Laplacian pyramid.

\begin{figure*}
	\centering
	\begin{subfigure}[c]{0.162\textwidth}
		\centering
		\includegraphics[width=\textwidth]{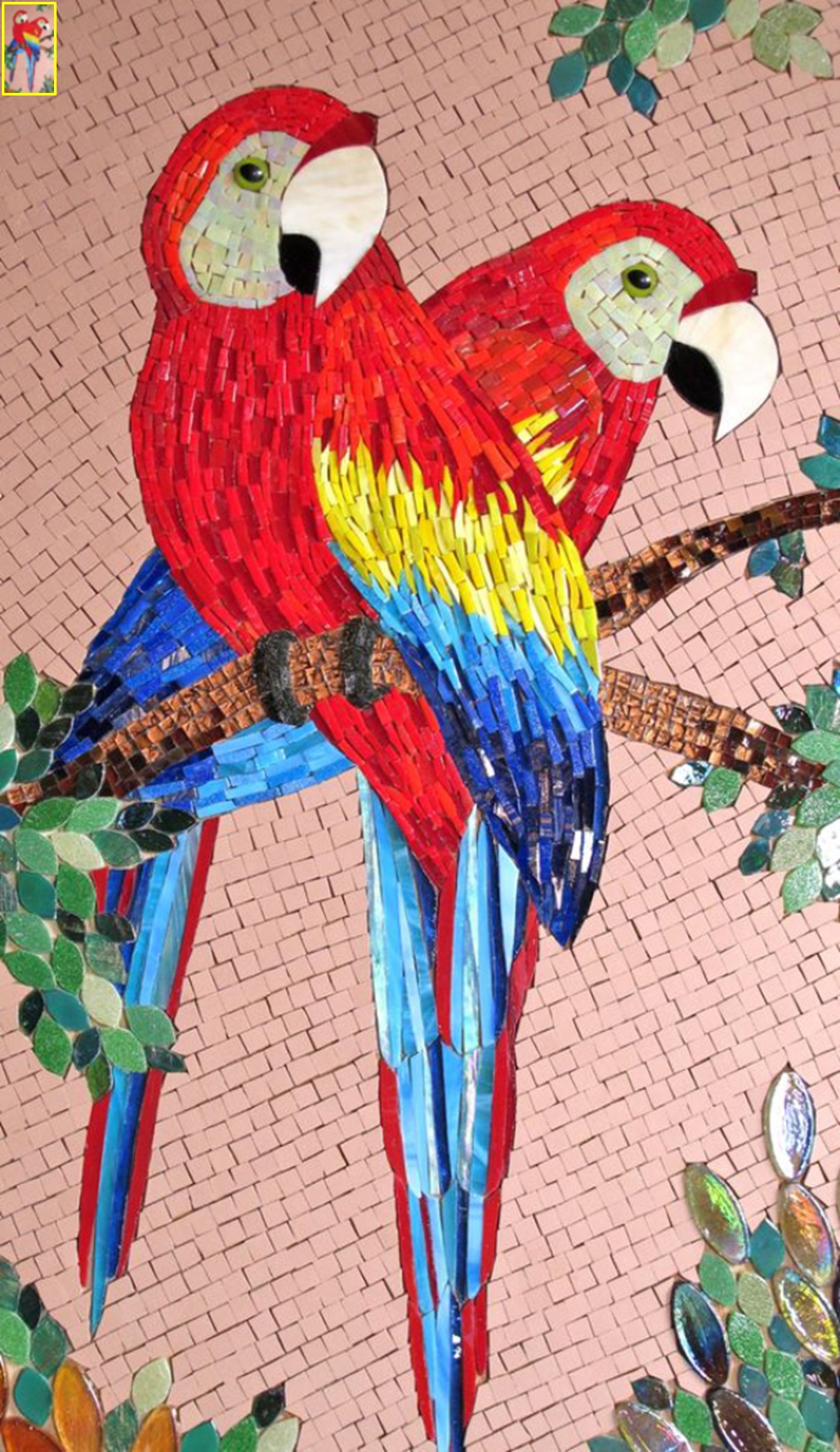} \\
        \vspace{-2mm}
		\caption{}
	\end{subfigure}
	\begin{subfigure}[c]{0.162\textwidth}
		\centering
		\includegraphics[width=\linewidth]{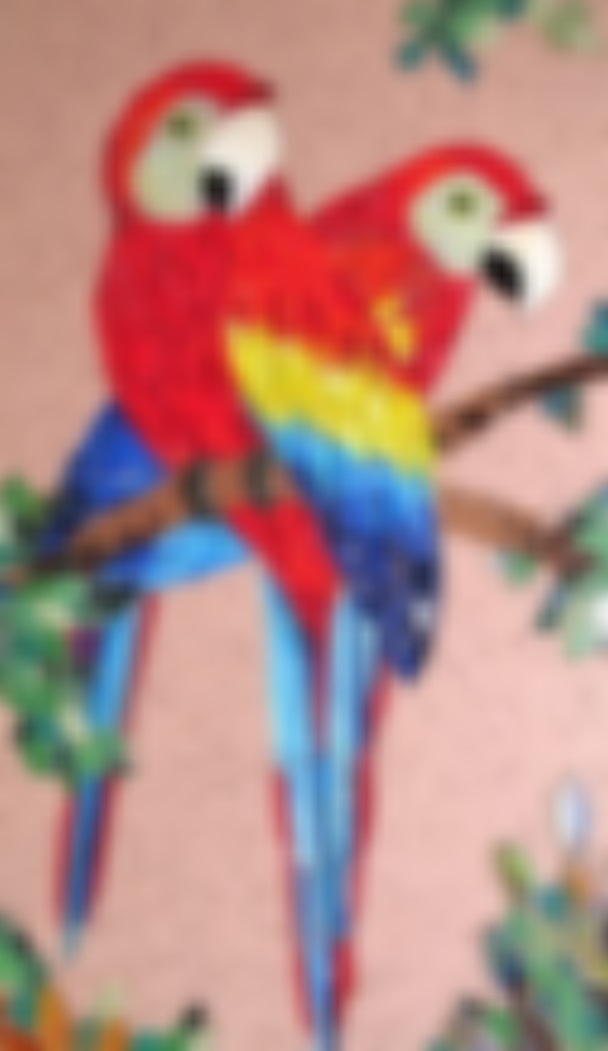} \\
        \vspace{-2mm}
		\caption{}
	\end{subfigure}
	\begin{subfigure}[c]{0.162\textwidth}
		\centering
		\includegraphics[width=\linewidth]{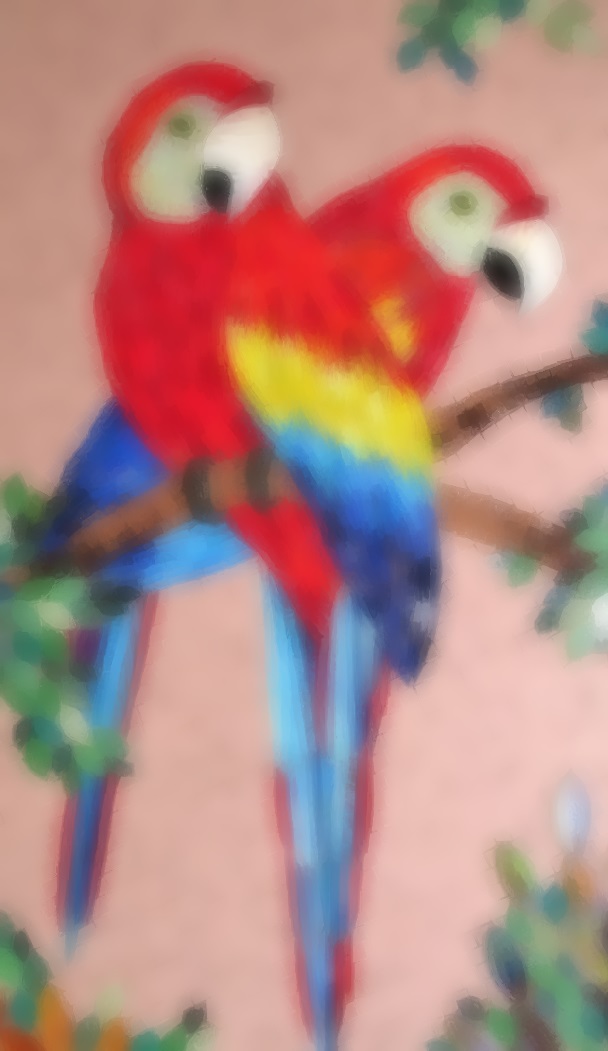} \\
        \vspace{-2mm}
		\caption{}
	\end{subfigure}
	\begin{subfigure}[c]{0.162\textwidth}
		\centering
		\includegraphics[width=\linewidth]{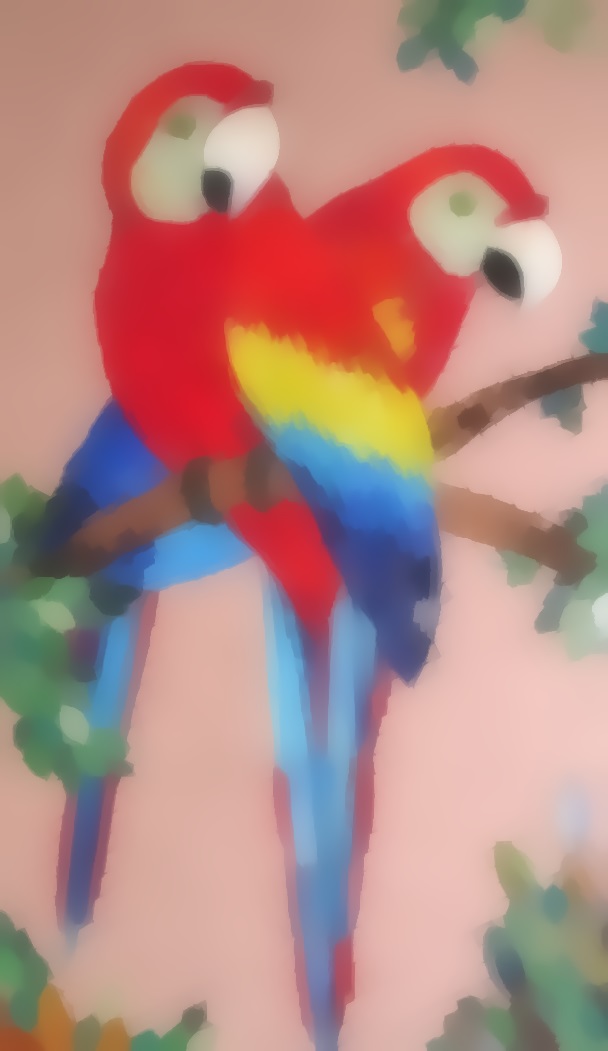} \\
        \vspace{-2mm}
		\caption{}
	\end{subfigure}
	\begin{subfigure}[c]{0.162\textwidth}
		\centering
		\includegraphics[width=\linewidth]{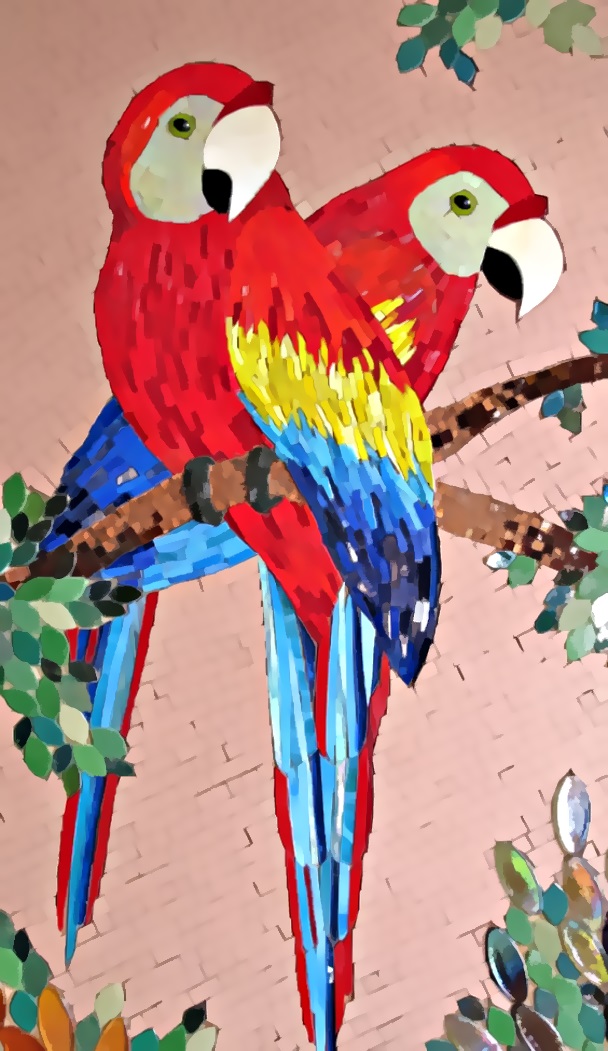} \\
        \vspace{-2mm}
		\caption{}
	\end{subfigure}
	\begin{subfigure}[c]{0.162\textwidth}
		\centering
		\includegraphics[width=\linewidth]{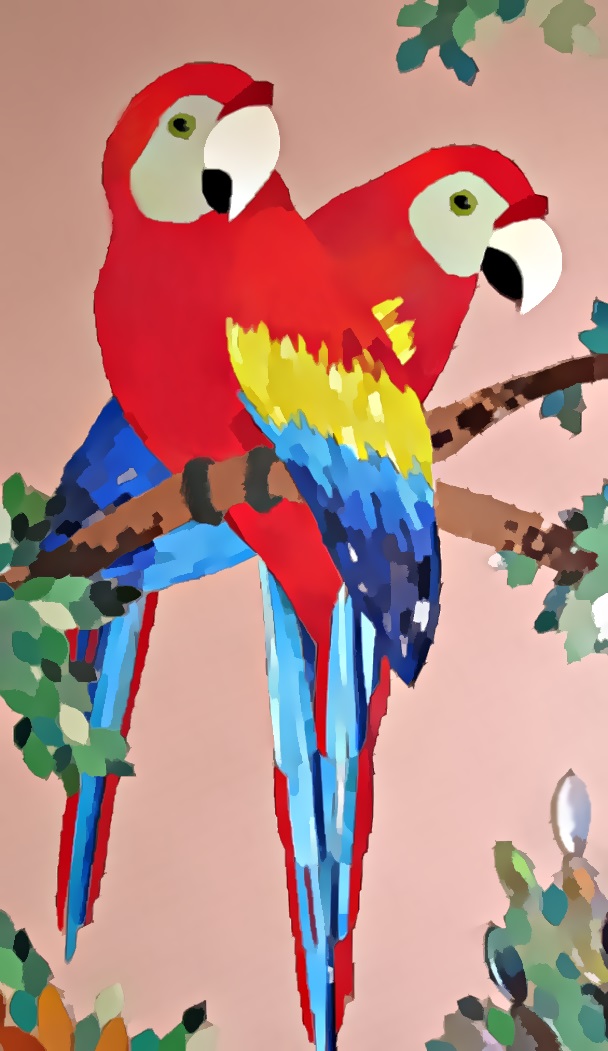} \\
        \vspace{-2mm}
		\caption{}
	\end{subfigure}
    \vspace{-2mm}
	\caption{Results using different upsampling strategies for the coarsest level in Gaussian pyramid. (a) Input image and its coarsest Gaussian pyramid level (shown in the top-left corner) with only 1/16 of the original full-resolution. (b) Bilinear upsampling. (c) Joint bilateral upsampling \cite{kopf2007joint} using the original image (a) as guidance. (d) Iterative joint bilateral upsampling using each fine-scale Gaussian pyramid level as guidance, i.e., our upsampling without the Laplacian pyramid based operation in Eq.~\eqref{equ:jbf}. (e) Our upsampling that replaces Eq.~\eqref{equ:jbf} with a different Laplacian pyramid usage $R_{k-1} = \hat{R}_{k-1} + \textrm{JBF}(L_{k-1}, \hat{R}_{k-1})$. (f) Our upsampling method. Image courtesy of Mosaics In Mind. }
	\label{fig:ablation} 
\end{figure*}

\section{Main Observation}
Figure~\ref{fig:observation} illustrates the observation that motivates the proposal of our method. It can be observed that as the scale of the Gaussian pyramid level becomes coarser, small-scale textures are gradually eliminated, finally yielding a structure-only coarsest Gaussian pyramid level. This inherent scale-separation property of Gaussian pyramid inspires us that it is feasible to perform texture smoothing by upsampling the coarsest Gaussian pyramid level to the original full-resolution. However, since the coarsest level in a Gaussian pyramid usually has very low-resolution, existing image upsampling strategies, e.g., bilinear upsampling and joint bilateral upsampling \cite{kopf2007joint}, are insufficient to produce high-quality texture smoothing results with sharp structures, as shown in Figure~\ref{fig:ablation}(b) and (c). To this end, we develop a novel pyramid-based upsampling approach, which will be introduced in the next section.

\section{Approach}
Given an input image $I$, and its Gaussian and Laplacian pyramids $\{G_{\ell}\}$ and $\{L_{\ell}\}$ with $\ell = N$ indicates the coarsest scale and $\ell=0$ corresponds to the finest scale ($G_0 = I$), our method aims to produce a texture filtered image $R$, by upsampling the coarsest Gaussian pyramid level $G_N$ to the original full-resolution. Figure~\ref{fig:overview} presents the overview of our approach. As shown, at the core of our method is the proposed pyramid-guided structure-aware upsampling, which is employed to iteratively upsample $G_N$ to a series of intermediate texture smoothing images $R_k$ at different scales $k$ ($0 \leq k \leq N-1$), until a full-resolution texture smoothing output $R_0$ ($R_0 = R$) with finest structure obtained. In the following, we first introduce the pyramid-guided structure-aware upsampling, and then elaborate the implementation details of our approach.

\subsection{Pyramid-Guided Structure-Aware Upsampling}

\paragraph{Background} We first summarize the joint bilateral filter (JBF) \cite{petschnigg2004digital,eisemann2004flash}, as we built our pyramid-guided structure-aware upsampling on top of it. Given an input image $I$, the joint bilateral filter computes an output image by replacing $I$ with a guidance image $\tilde{I}$ in range filter kernel of the bilateral filter \cite{tomasi1998bilateral}, which is expressed as:
\begin{equation} \label{equ:src_jbf}
\begin{split}
\textrm{JBF}(I, \tilde{I})_p &= \frac{1}{K_p}\sum_{q \in \Omega^d_p}g_{\sigma_s}(\|p-q\|)~g_{\sigma_r}(\|\tilde{I}_p - \tilde{I}_q\|)~I_q, \\
\textrm{with} \enspace K_p &= \sum_{q \in \Omega^d_p}g_{\sigma_s}(\|p-q\|)~g_{\sigma_r}(\|\tilde{I}_p - \tilde{I}_q\|),
\end{split}
\end{equation}
where $p$ and $q$ are pixel coordinates. $\Omega^d_p$ is a set of pixels in the $d \times d$ squared neighborhood centered at $p$. $g_{\sigma}(x) = \exp(-x^2/2\sigma^2)$ is a Gaussian kernel function with standard deviation $\sigma$. $\sigma_s$ and $\sigma_r$ control the spatial support and the sensitivity to edges, respectively. A well-known variant with similar formulation to Eq.~\eqref{equ:src_jbf} is joint bilateral upsampling (denoted as JBF$^\uparrow$ in paper) \cite{kopf2007joint}, which is designed to upsample a low-resolution image under the guidance of a high-resolution image in an edge-aware manner. 

\paragraph{Our upsampling} The dashed box in Figure~\ref{fig:overview} gives the workflow of our pyramid-guided structure-aware upsampling. As can be seen, its goal is to upsample an intermediate texture smoothing image $R_{k}$ at scale $k$ ($0 < k \leq N$) to a structure-refined fine-scale output $R_{k-1}$, with the aid of the Gaussian and Laplacian pyramid levels $G_{k-1}$ and $L_{k-1}$. Below we describe our upsampling algorithm in detail.

To produce $R_{k-1}$ from $R_k$, we start by upsampling $R_k$ to an initial output $\hat{R}_{k-1}$ at scale $k-1$ by performing joint bilateral upsampling using the Gaussian pyramid level $G_{k-1}$ as guidance:
\begin{equation} \label{equ:jbu}
\hat{R}_{k-1} = \textrm{JBF}^{\uparrow}(R_k, G_{k-1}).
\end{equation}
The process aims to make the upsampling output $\hat{R}_{k-1}$ share similar structure edges with $G_{k-1}$ that contains finer-scale structures. However, as shown in Figure~\ref{fig:overview}, the structure sharpness of $\hat{R}_{k-1}$ is still inferior to that of $G_{k-1}$, indicating that $\hat{R}_{k-1}$ loses certain amount of structure details compared to $G_{k-1}$, and iteratively performing the process in Eq.~\eqref{equ:jbu} will fail to generate texture smoothing output with sharp structures (see Figure~\ref{fig:ablation}(d)). Note, the above process will not introduce textures to the output (i.e., $\hat{R}_{k-1}$ is texture-free), as shown in Figure~\ref{fig:overview} and Figure~\ref{fig:ablation}(d). The reason is twofold. First, the starting input of the upsampling process is the coarsest level $G_N$ that barely contains textures. Second, as will be introduced later, we set a small neighborhood $d$ for the process to avoid mistaking texture details in the guidance image $G_{k-1}$ as structures. 

\begin{figure*}
    \centering
    \begin{subfigure}[c]{0.162\textwidth}
        \centering
        \includegraphics[width=\linewidth]{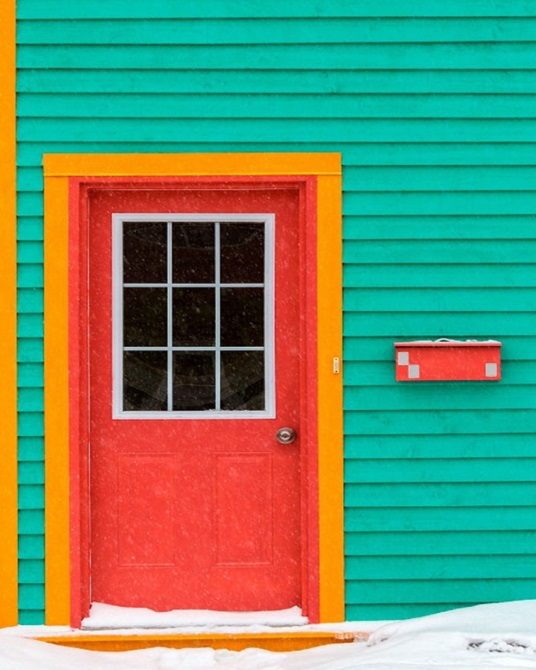} \\
        \vspace{-2mm}
        \caption{Input}
    \end{subfigure}
        \begin{subfigure}[c]{0.162\textwidth}
        \centering
        \includegraphics[width=\linewidth]{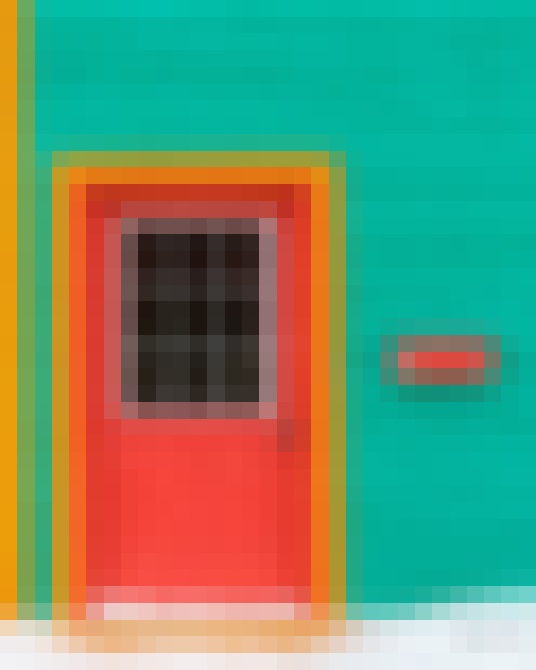} \\
        \vspace{-2mm}
        \caption{$R_4$ (coarsest level)}
    \end{subfigure}
        \begin{subfigure}[c]{0.162\textwidth}
        \centering
        \includegraphics[width=\linewidth]{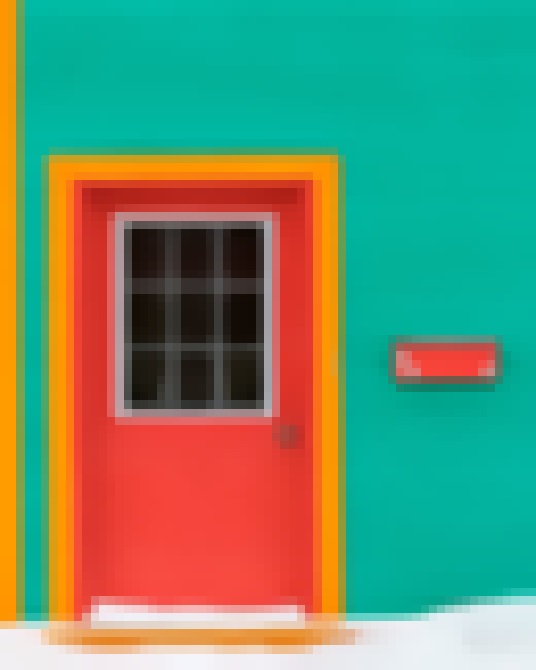} \\
        \vspace{-2mm}
        \caption{$R_3$}
    \end{subfigure}
        \begin{subfigure}[c]{0.162\textwidth}
        \centering
        \includegraphics[width=\linewidth]{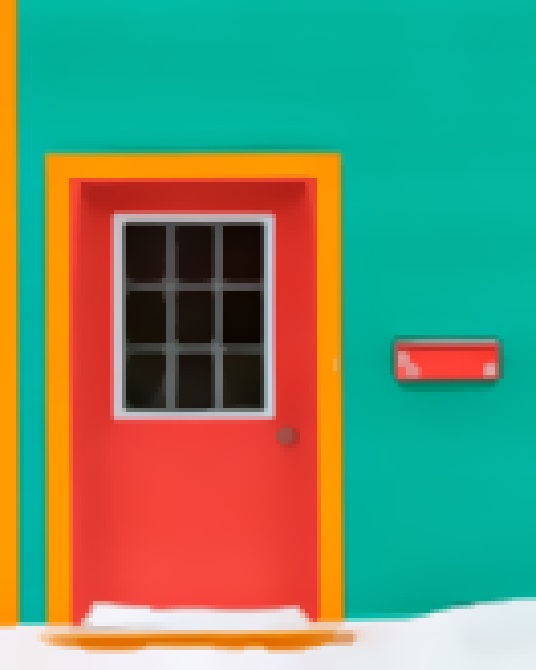} \\
        \vspace{-2mm}
        \caption{$R_2$}
    \end{subfigure}
        \begin{subfigure}[c]{0.162\textwidth}
        \centering
        \includegraphics[width=\linewidth]{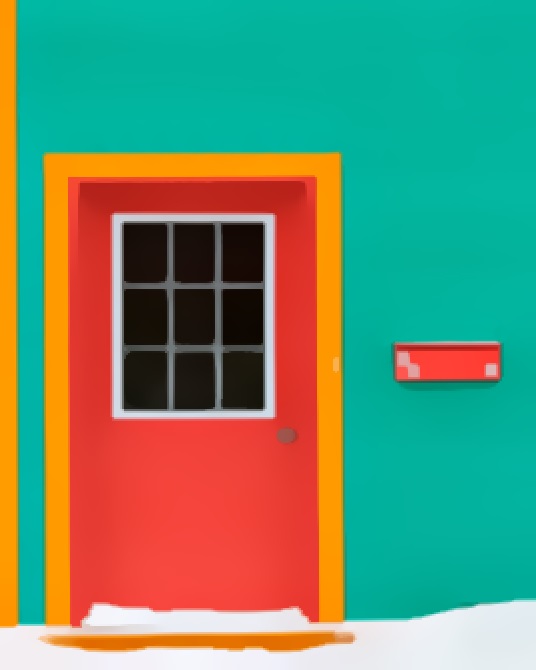} \\
        \vspace{-2mm}
        \caption{$R_1$}
    \end{subfigure}
        \begin{subfigure}[c]{0.162\textwidth}
        \centering
        \includegraphics[width=\linewidth]{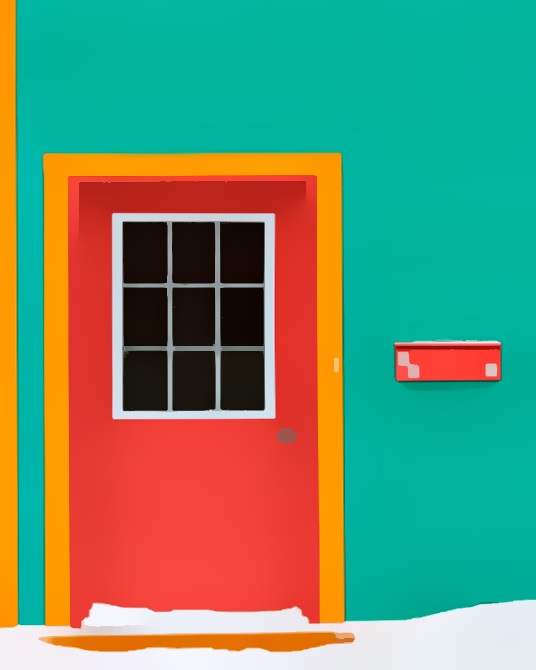} \\
        \vspace{-2mm}
        \caption{$R_0$ (final result)}
    \end{subfigure}
    \vspace{-2mm}
    \caption{Our texture smoothing images at different scales, where all intermediate results ($R_4$, $R_3$, $R_2$, $R_1$,) are showed at original full-resolution. Image courtesy of Pinterest user bianca. }
    \label{fig:results_diff_scales}  
\end{figure*}

\begin{figure*}
	\centering
	\captionsetup[subfigure]{labelformat=empty}
	\begin{subfigure}[c]{0.162\textwidth}
		\centering
		\includegraphics[width=\linewidth]{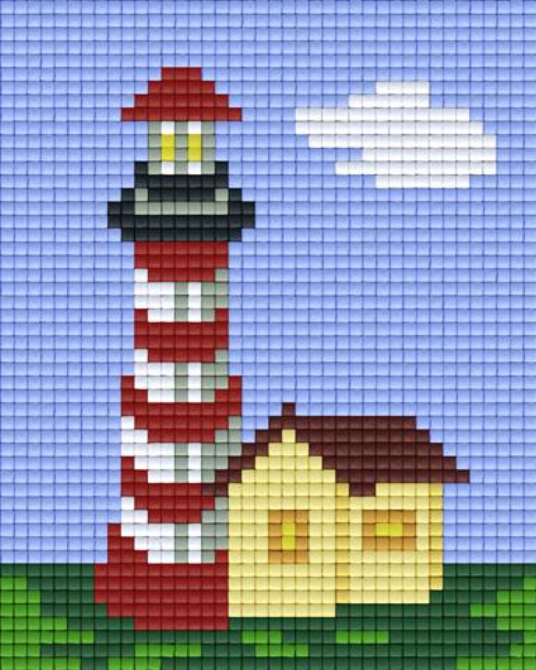}\\ \vspace{2pt}
		\includegraphics[width=\linewidth]{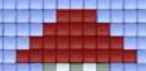}\\
		\vspace{-2mm}
		\caption{Input}
	\end{subfigure}
	\begin{subfigure}[c]{0.162\textwidth}
		\centering
		\includegraphics[width=\linewidth]{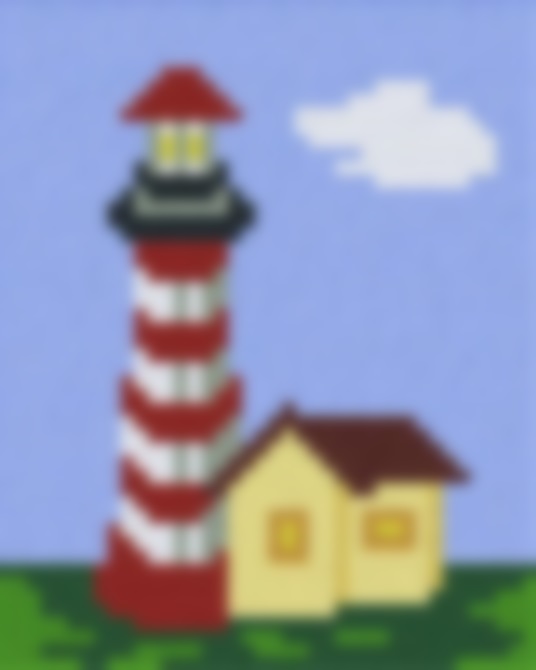}\\ \vspace{2pt}
		\includegraphics[width=\linewidth]{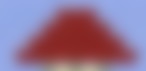}\\
		\vspace{-2mm}
		\caption{Gaussian filtering ($\sigma=7$)}
	\end{subfigure}
	\begin{subfigure}[c]{0.162\textwidth}
		\centering
		\includegraphics[width=\linewidth]{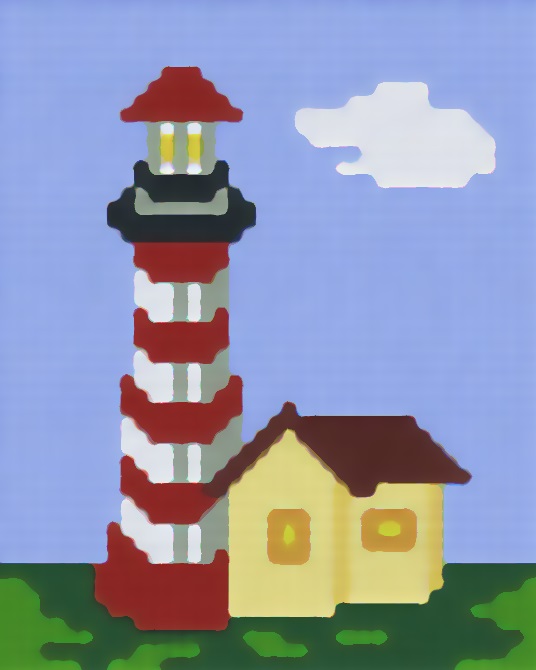}\\ \vspace{2pt}
		\includegraphics[width=\linewidth]{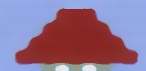}\\
		\vspace{-2mm}
		\caption{\cite{zhang2014rolling}}
	\end{subfigure}
	\begin{subfigure}[c]{0.162\textwidth}
		\centering
		\includegraphics[width=\linewidth]{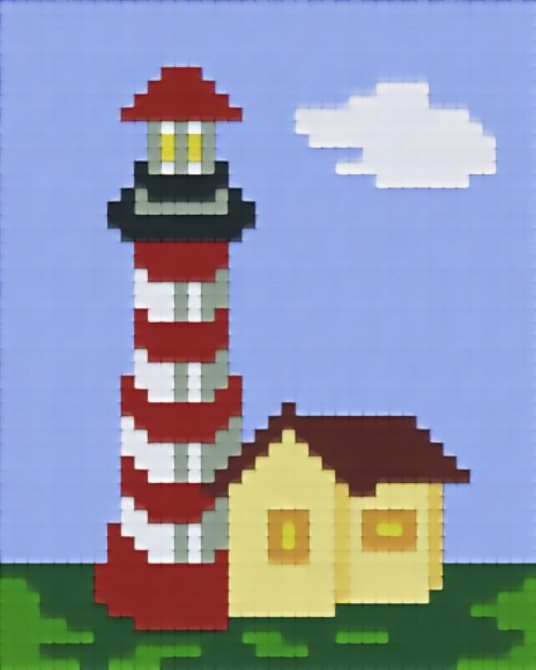}\\ \vspace{2pt}
		\includegraphics[width=\linewidth]{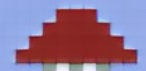}\\
		\vspace{-2mm}
		\caption{\cite{du2016two}}
	\end{subfigure}
	\begin{subfigure}[c]{0.162\textwidth}
		\centering
		\includegraphics[width=\linewidth]{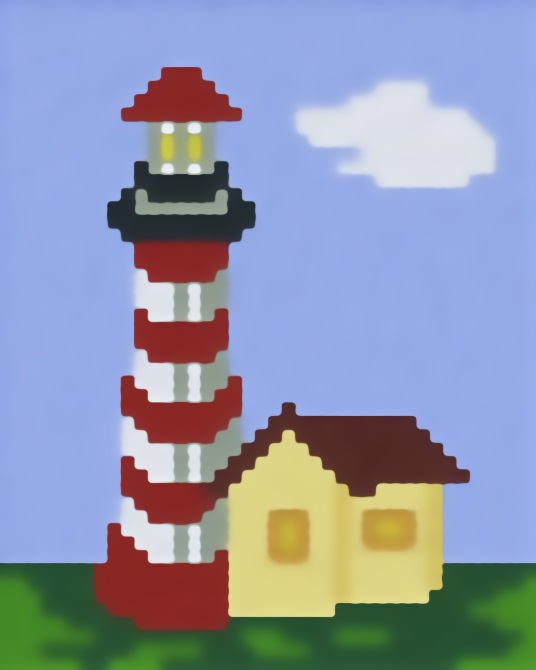}\\ \vspace{2pt}
		\includegraphics[width=\linewidth]{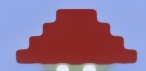}\\
		\vspace{-2mm}
		\caption{\cite{jeon2016scale}}
	\end{subfigure}
	\begin{subfigure}[c]{0.162\textwidth}
		\centering
		\includegraphics[width=\linewidth]{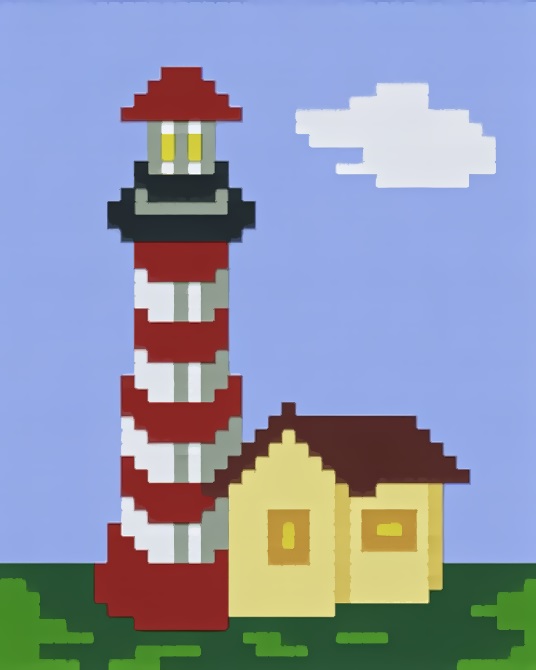}\\ \vspace{2pt}
		\includegraphics[width=\linewidth]{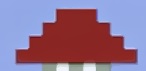}\\
		\vspace{-2mm}
		\caption{Ours}
	\end{subfigure}
	\vspace{-2mm}
	\caption{Comparison with previous scale-aware texture smoothing methods. As shown, previous scale-aware methods built upon Gaussian smoothing either tends to distort structures or may lead to noticeable texture residuals. Parameters: \cite{zhang2014rolling} ($\sigma_s=5$, $\sigma_r=0.03$, $N^{iter}=8$), \cite{du2016two} ($\sigma_s = 5$, $\sigma_r = 0.05$, $N^{iter}=5$), \cite{jeon2016scale} ($\sigma=5$, $\sigma_r=0.1$, $N^{iter}=5$), and ours ($\sigma_s=5$, $\sigma_r=0.03$). Image courtesy of Pixelhobby.}
	\label{fig:comp_scale_aware}  %
\end{figure*}

To further refine structures and obtain the resulting $R_{k-1}$ with sharper structures than $\hat{R}_{k-1}$, we introduce Laplacian pyramid into our method because it records exactly the image details that each Gaussian pyramid level loses compared to its next coarser level. Specifically, we propose to compute $R_{k-1}$ by smoothing out the unwanted texture details from the sum of $\hat{R}_{k-1}$ and $L_{k-1}$ via joint bilateral filtering with $\hat{R}_{k-1}$ as guidance:
\begin{equation} \label{equ:jbf}
R_{k-1} = \textrm{JBF}(\hat{R}_{k-1} + L_{k-1}, \hat{R}_{k-1}).
\end{equation}
The reason to use $\hat{R}_{k-1}$ as guidance is because it is texture-free and has similar structure edges to the sum of $\hat{R}_{k-1}$ and $L_{k-1}$, allowing effective texture removal and faithfully maintaining the desired structure residuals carried by $L_{k-1}$ (see Figure~\ref{fig:overview}). 

\begin{algorithm}[t]
	\SetAlgoLined
	\KwIn{image $I$, parameters ${\sigma}_s$ and ${\sigma}_r$}
	\KwOut{texture smoothing image $R$}
	
	build Gaussian and Laplacian pyramids $\{G_{\ell}\}$ and $\{L_{\ell}\}$ for image $I$
	
	$R_N \leftarrow$ the coarsest level $G_N$
	
	\For{$k = N-1:0$}{
		$\hat{R}_k = \textrm{JBF}^{\uparrow}(R_{k+1}, G_k)$
		
		$R_k= \textrm{JBF}(\hat{R}_k + L_k, \hat{R}_k)$
	
	}
	$R \leftarrow R_0$
	\caption{Pyramid Texture Filtering}
	\label{alg:ptf}
\end{algorithm}

\begin{figure}
	\centering
	\begin{subfigure}[c]{0.155\textwidth}
		\centering
		\includegraphics[width=\linewidth]{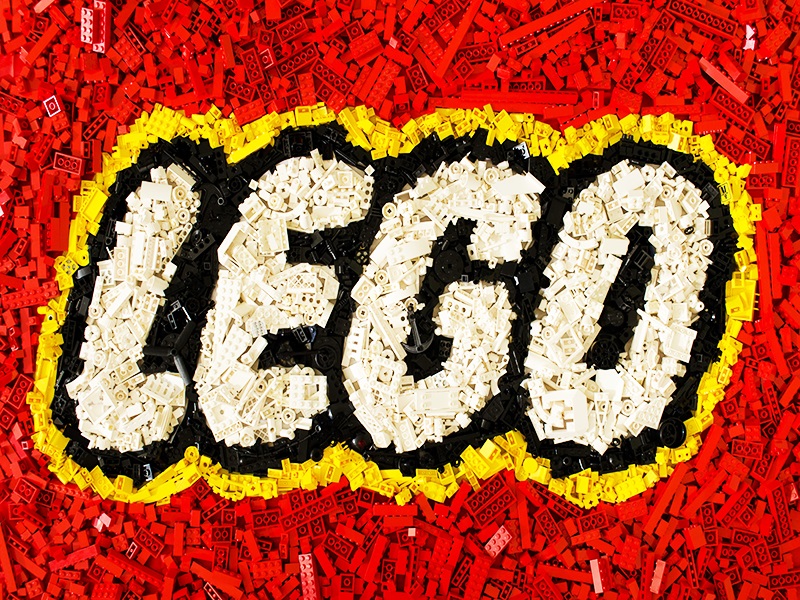} \\
		\vspace{-2mm}
		\caption{Input}
	\end{subfigure}
	\begin{subfigure}[c]{0.155\textwidth}
		\centering
		\includegraphics[width=\linewidth]{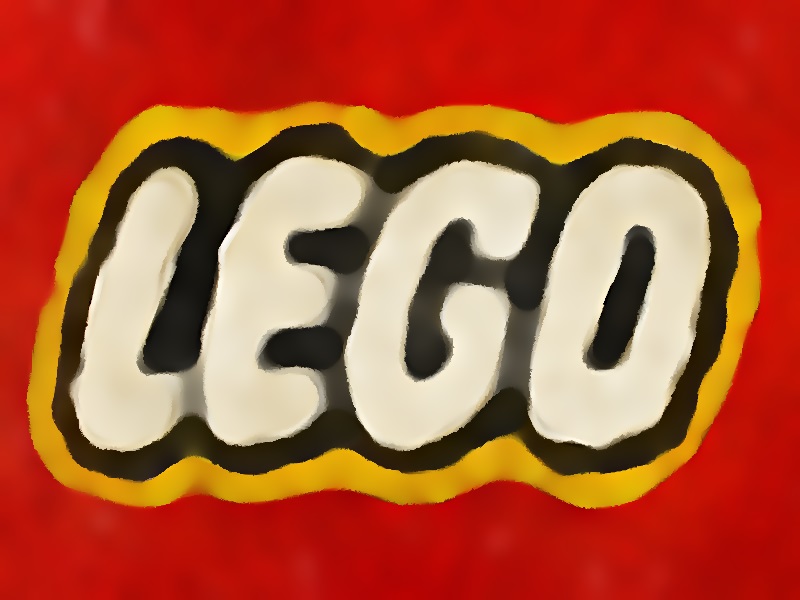} \\
		\vspace{-2mm}
		\caption{Without pyramids}
	\end{subfigure}
	\begin{subfigure}[c]{0.155\textwidth}
		\centering
		\includegraphics[width=\linewidth]{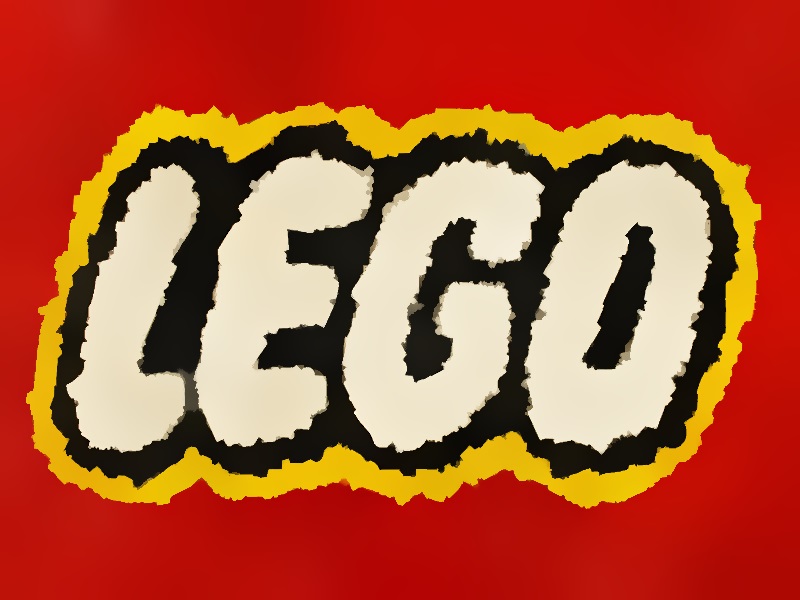} \\
		\vspace{-2mm}
		\caption{With pyramids}
	\end{subfigure}
	\vspace{-2mm}
	\caption{Comparison of results with and without using image pyramids. It is clear that image pyramids are beneficial to structure preservation. Image courtesy of Flickr user tikitikitembo.}
	\label{fig:no_pyramid}
	\vspace{-2mm}
\end{figure}

\begin{figure*}
	\centering
	\begin{subfigure}[c]{0.196\textwidth}
		\centering
		\includegraphics[width=\linewidth]{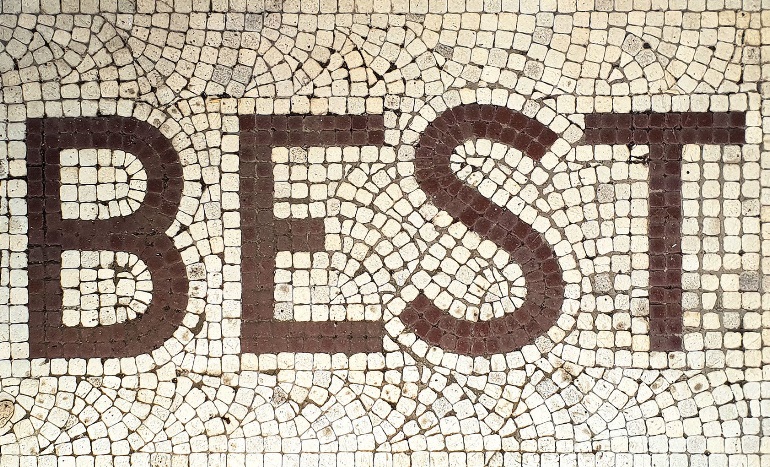} \\
		\vspace{-2mm}
		\caption{Input}
	\end{subfigure}
	\begin{subfigure}[c]{0.196\textwidth}
		\centering
		\includegraphics[width=\linewidth]{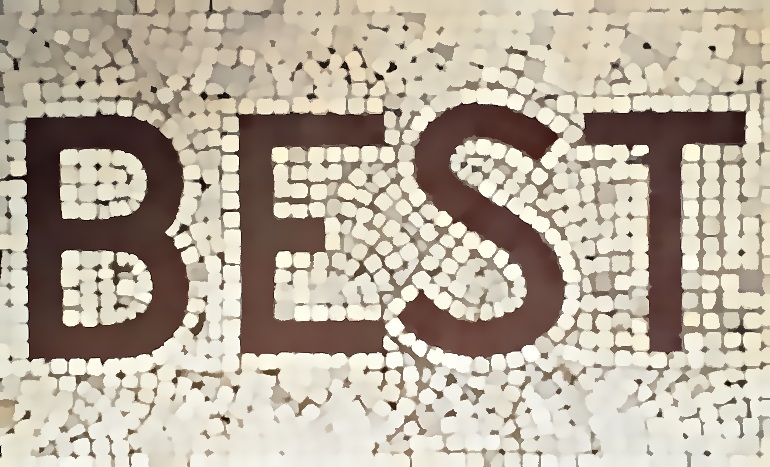} \\
		\vspace{-2mm}
		\caption{$\sigma_s=3$, $\sigma_r=0.05$}
	\end{subfigure}
	\begin{subfigure}[c]{0.196\textwidth}
		\centering
		\includegraphics[width=\linewidth]{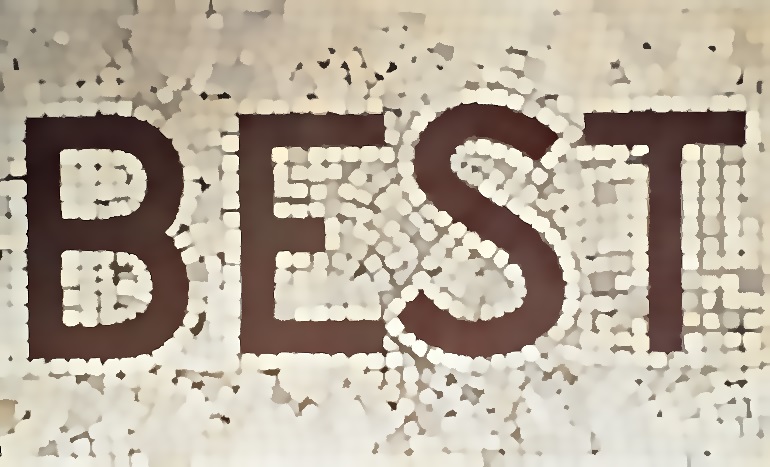} \\
		\vspace{-2mm}
		\caption{$\sigma_s=3$, $\sigma_r=0.07$}
	\end{subfigure}
	\begin{subfigure}[c]{0.196\textwidth}
		\centering
		\includegraphics[width=\linewidth]{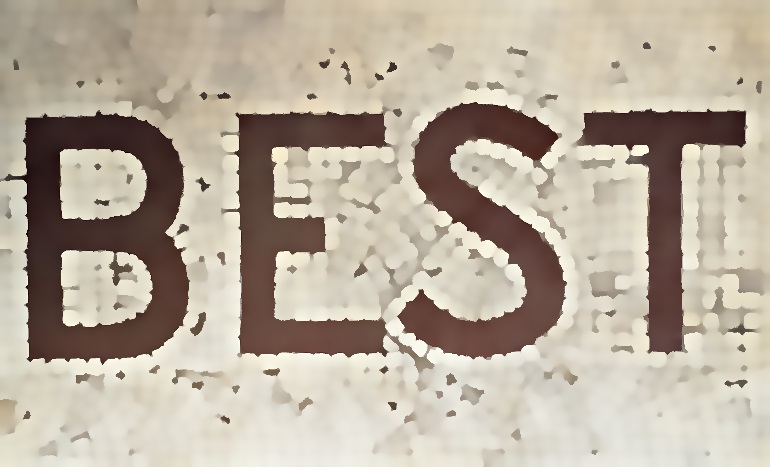} \\
		\vspace{-2mm}
		\caption{$\sigma_s=3$, $\sigma_r=0.09$}
	\end{subfigure}
	\begin{subfigure}[c]{0.196\textwidth}
		\centering
		\includegraphics[width=\linewidth]{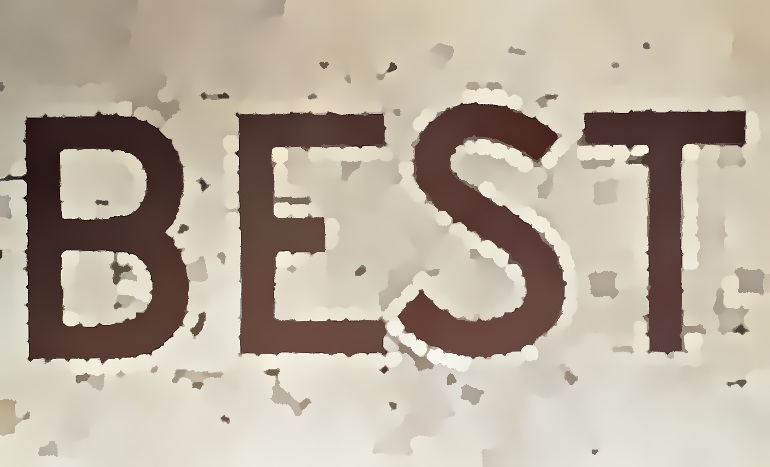} \\
		\vspace{-2mm}
		\caption{$\sigma_s=5$, $\sigma_r=0.05$}
	\end{subfigure}  \vspace{2pt} \\
	\begin{subfigure}[c]{0.196\textwidth}
		\centering
		\includegraphics[width=\linewidth]{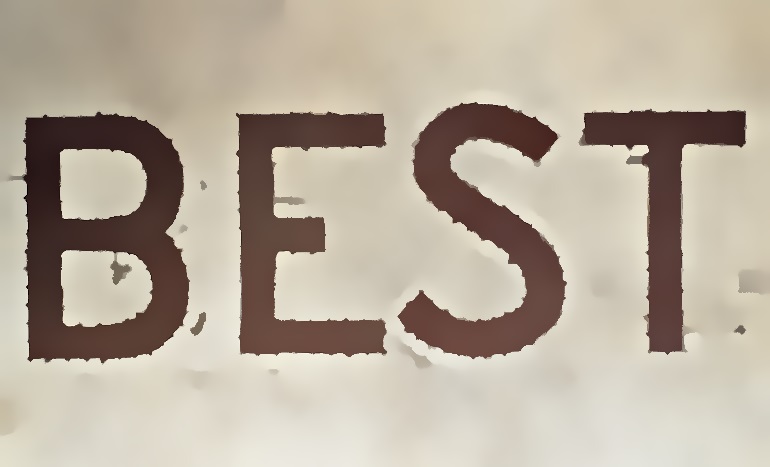} \\
		\vspace{-2mm}
		\caption{$\sigma_s=5$, $\sigma_r=0.07$}
	\end{subfigure}
	\begin{subfigure}[c]{0.196\textwidth}
		\centering
		\includegraphics[width=\linewidth]{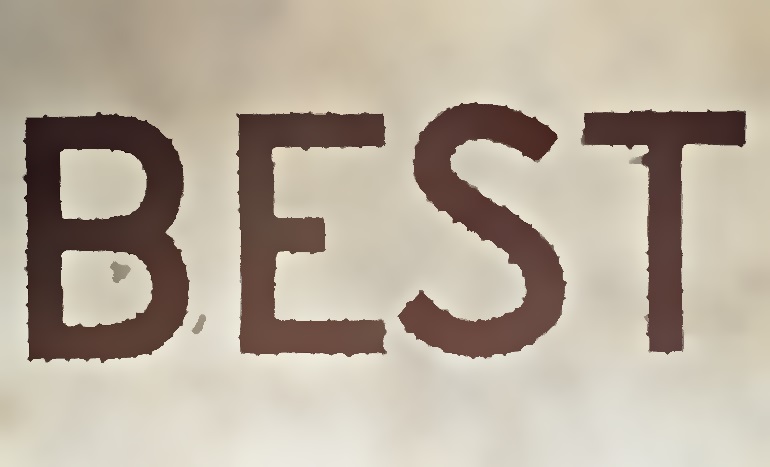} \\
		\vspace{-2mm}
		\caption{$\sigma_s=5$, $\sigma_r=0.09$}
	\end{subfigure}
	\begin{subfigure}[c]{0.196\textwidth}
		\centering
		\includegraphics[width=\linewidth]{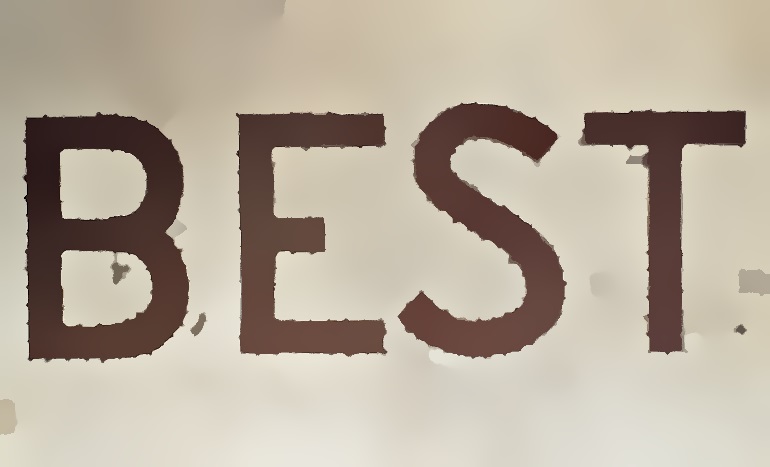} \\
		\vspace{-2mm}
		\caption{$\sigma_s=7$, $\sigma_r=0.05$}
	\end{subfigure}
	\begin{subfigure}[c]{0.196\textwidth}
		\centering
		\includegraphics[width=\linewidth]{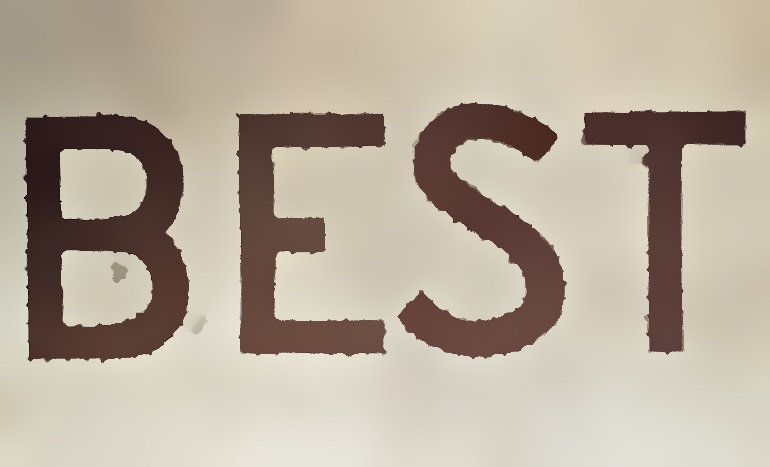} \\
		\vspace{-2mm}
		\caption{$\sigma_s=7$, $\sigma_r=0.07$}
	\end{subfigure}
	\begin{subfigure}[c]{0.196\textwidth}
		\centering
		\includegraphics[width=\linewidth]{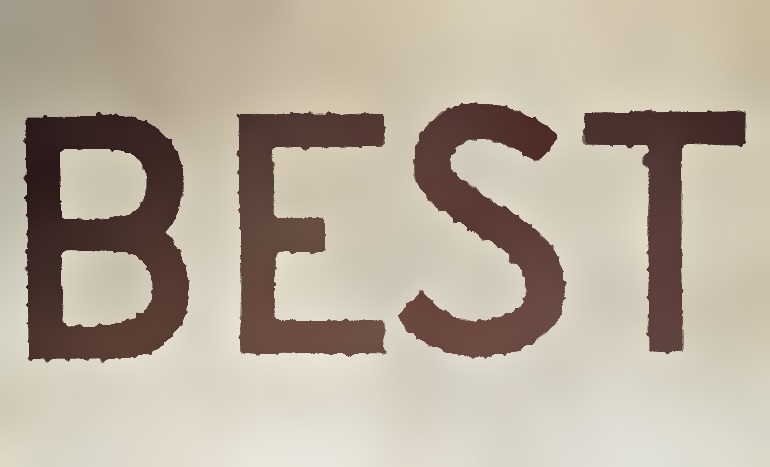} \\
		\vspace{-2mm}
		\caption{$\sigma_s=7$, $\sigma_r=0.09$}
	\end{subfigure}
	\vspace{-2mm}
	\caption{Results with varying $\sigma_s$ and $\sigma_r$. As shown, larger $\sigma_s$ and $\sigma_r$ have stronger ability in texture removal. Image courtesy of Flickr user Roger Marks.}
	\label{fig:varying_parameters}  
\end{figure*}

\begin{figure*}
	\centering
	\captionsetup[subfigure]{justification=centering}
	\begin{subfigure}[c]{0.163\textwidth}
		\centering
		\includegraphics[width=\linewidth]{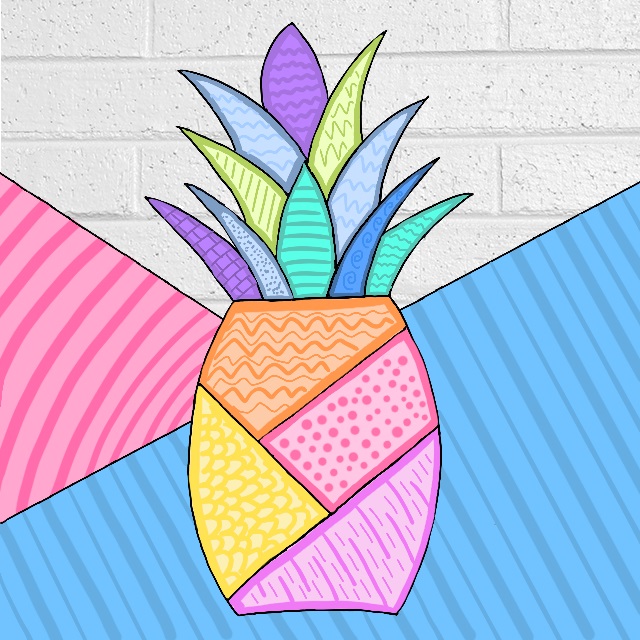}\\
		\vspace{-2mm}
		\caption{Input\\
			image size: $640 \times 640$}
	\end{subfigure}
	\begin{subfigure}[c]{0.163\textwidth}
		\centering
		\includegraphics[width=\linewidth]{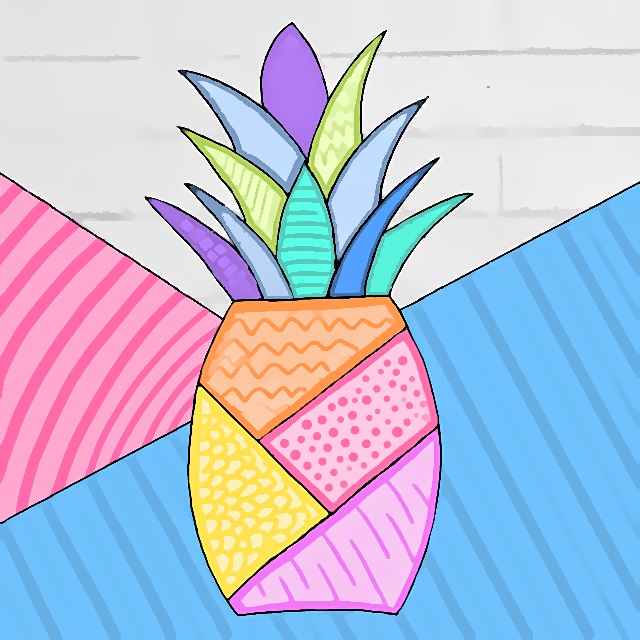}\\
		\vspace{-2mm}
		\caption{depth=2 \\ coarsest level: $320 \times 320$ }
	\end{subfigure}
	\begin{subfigure}[c]{0.163\textwidth}
		\centering
		\includegraphics[width=\linewidth]{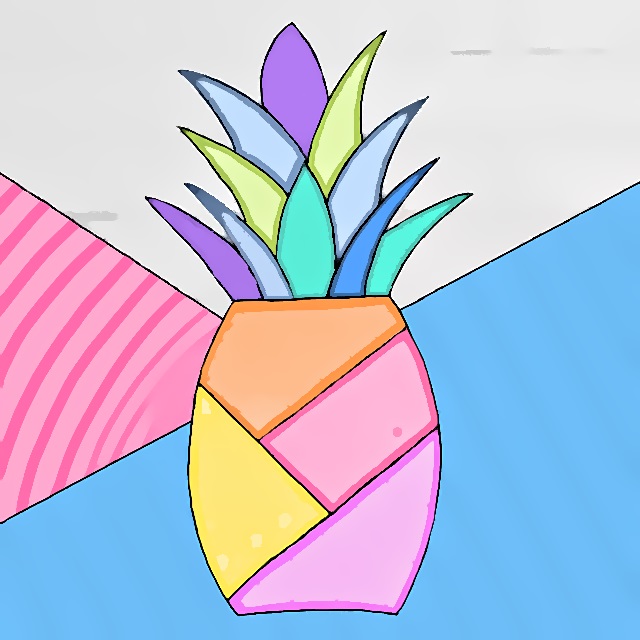}\\
		\vspace{-2mm}
		\caption{depth=3 \\ coarsest level: $160 \times 160$}
	\end{subfigure} 
	\begin{subfigure}[c]{0.163\textwidth}
		\centering
		\includegraphics[width=\linewidth]{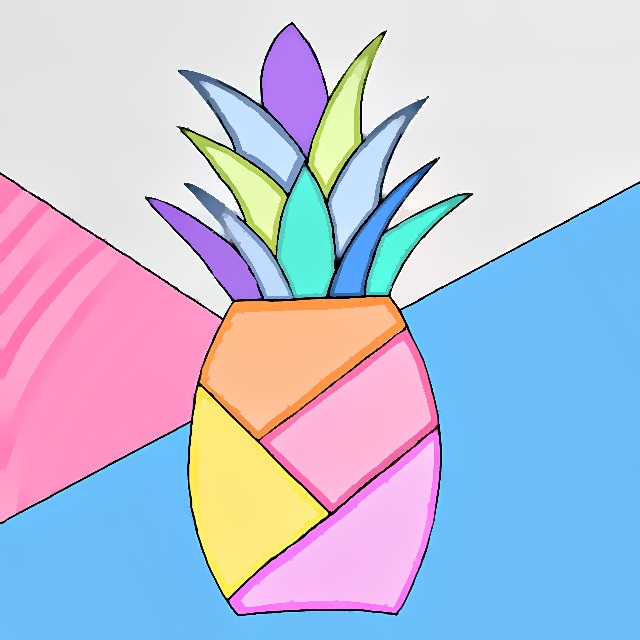}\\
		\vspace{-2mm}
		\caption{depth=4 \\ coarsest level: $80 \times 80$}
	\end{subfigure}
	\begin{subfigure}[c]{0.163\textwidth}
		\centering
		\includegraphics[width=\linewidth]{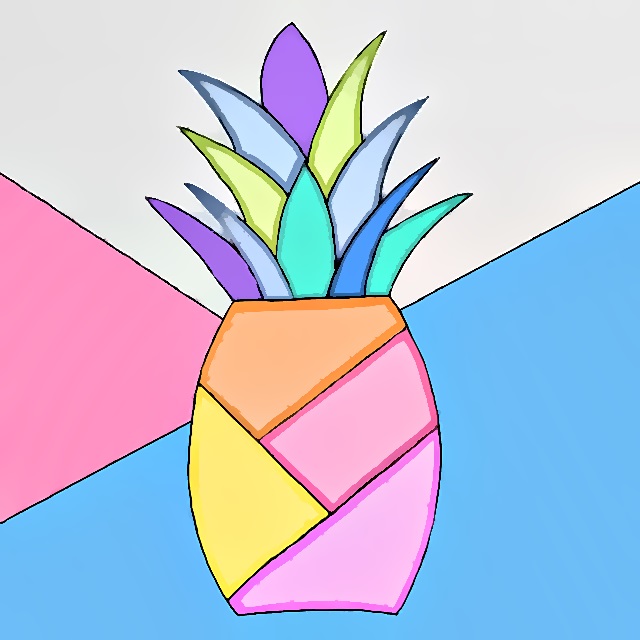}\\
		\vspace{-2mm}
		\caption{depth=5 \\ coarsest level: $40 \times 40$}
	\end{subfigure}
	\begin{subfigure}[c]{0.163\textwidth}
		\centering
		\includegraphics[width=\linewidth]{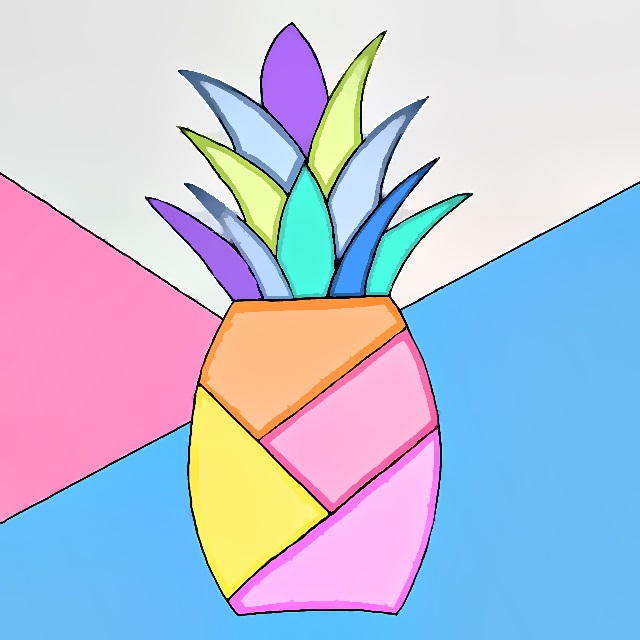}\\
		\vspace{-2mm}
		\caption{depth=6 \\ coarsest level: $20 \times 20$}
	\end{subfigure}
	\vspace{-2mm}
	\caption{Results with varying pyramid depths. Deeper pyramid with smaller coarsest level helps remove larger-scale textures. Image courtesy of Classy Chick.} 
	\label{fig:diff_pyramid_depth} 
\end{figure*}
	
Figure~\ref{fig:results_diff_scales} shows how the texture smoothing output varies with the iterative upsampling, where we can see that the image structures are gradually refined along with the upsampling, without introducing textures. As can be seen, although structures in the input image are not properly aligned with the severely blurred structures in the coarsest level $R_4$, especially the white wooden fence window, our method is also able to produce a high-quality texture smoothing output $R_0$ with structures almost as sharp as those in the input image. The reason is that our method does not directly deal with the severe structure misalignment between the original image and the coarsest level, but instead iteratively performs pyramid-guided structure-aware upsampling to gradually refine structures, which is obviously easier and more reliable because the structure misalignment between two adjacent pyramid levels is much weak. This also explains why naively upsampling the coarsest level through joint bilateral upsampling with the original image as guidance fails to generate result with sharp structures, as shown in Figure~\ref{fig:ablation}(c).

\paragraph{Parameters} As suggested by Eq.~\eqref{equ:src_jbf}, the success of our method is also related to the joint bilateral filtering parameters in Eqs.~\eqref{equ:jbu} and \eqref{equ:jbf} in our  upsampling at each scale, i.e., the spatial and range smoothing parameters $\sigma_s$ and $\sigma_r$ as well as the neighborhood size $d$. Note that although our upsampling at each scale involves six parameters, there are actually only two parameters, $\sigma_{s}$ and $\sigma_{r}$, to control throughout the entire iterative upsampling, as all other parameters can be adaptively adjusted according to the two parameters.

\begin{figure}
	\centering
	\begin{subfigure}[c]{0.155\textwidth}
		\centering
		\includegraphics[width=\linewidth]{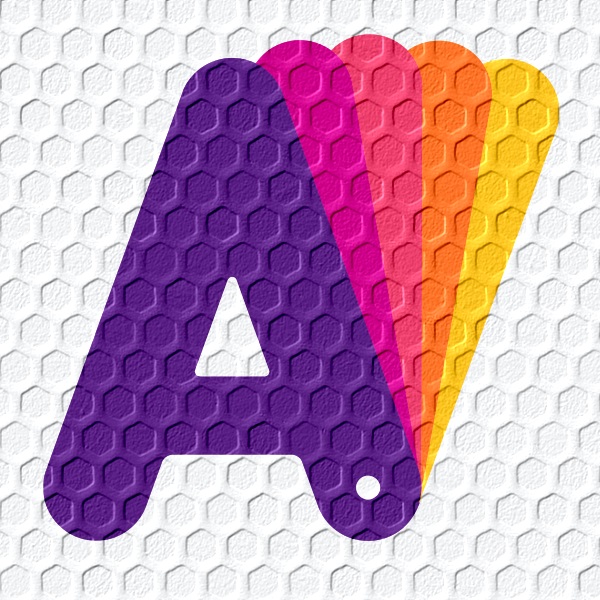} \\
		\vspace{-2mm}
		\caption{Input}
	\end{subfigure}
	\begin{subfigure}{0.155\textwidth}
		\centering
		\includegraphics[width=\linewidth]{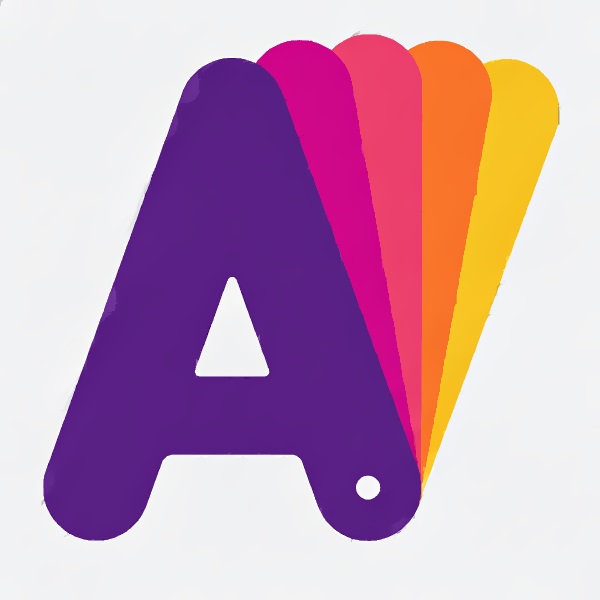} \\
		\vspace{-2mm}
		\caption{Single smoothing}
	\end{subfigure}
	\begin{subfigure}{0.155\textwidth}
		\centering
		\includegraphics[width=\linewidth]{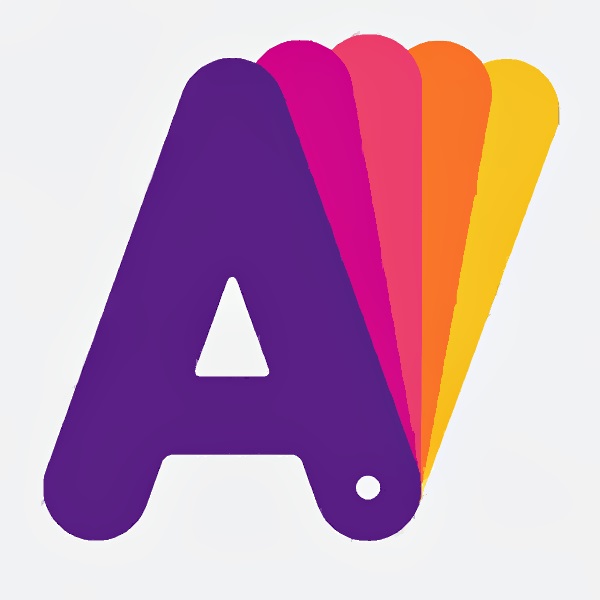}\\
		\vspace{-2mm}
		\caption{Smoothing twice}
	\end{subfigure}
	\vspace{-2mm}
	\caption{Comparison of single and multiple smoothing. Note, the result in (c) is generated with the result in (b) as input. We find that single smoothing is usually enough to remove textures, and smoothing for multiple times would not gain more improvement nor further degrade image structures. Source image \copyright Letters.} 
	\label{fig:multi_filtering}
\end{figure}

\begin{figure}
	\centering
	\begin{subfigure}[c]{0.236\textwidth}
		\centering
		\begin{tikzpicture}[
		spy using outlines={color=red, rectangle, magnification=3,
			every spy on node/.append style={rectangle},
			every spy in node/.append style={rectangle}}
		]
		\node[inner sep=0,outer sep=0]{\includegraphics[width=\linewidth]{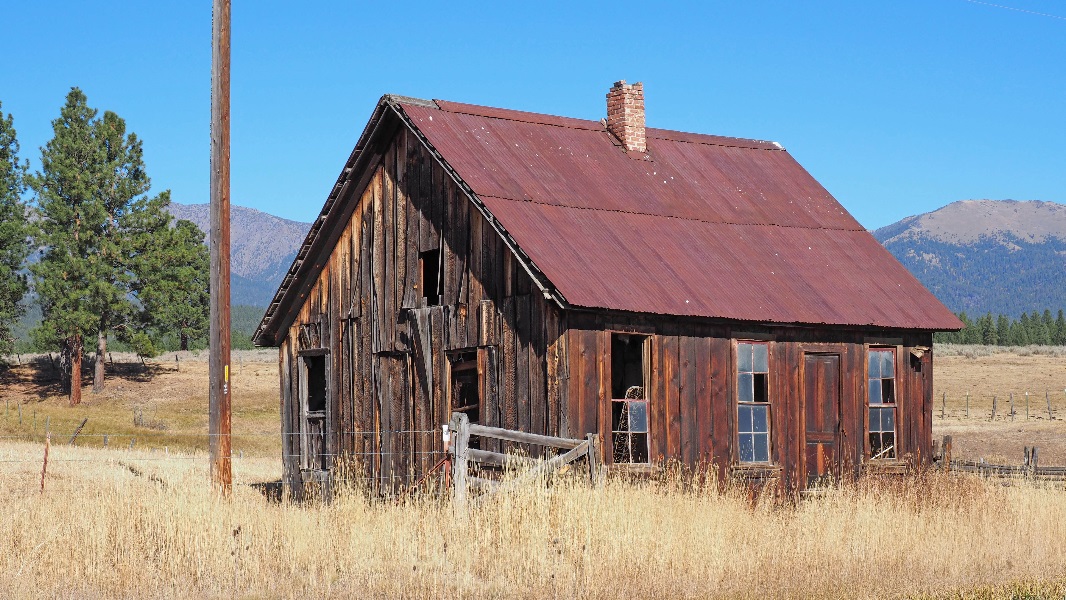}};
		\spy[width=1.0cm, height=0.8cm] on (0.35, 0.75) in node at (1.6, 0.8);
		\end{tikzpicture} \\ \vspace{2pt}
		\begin{tikzpicture}[
		spy using outlines={color=red, rectangle, magnification=3,
			every spy on node/.append style={rectangle},
			every spy in node/.append style={rectangle}}
		]
		\node[inner sep=0,outer sep=0]{\includegraphics[width=\linewidth]{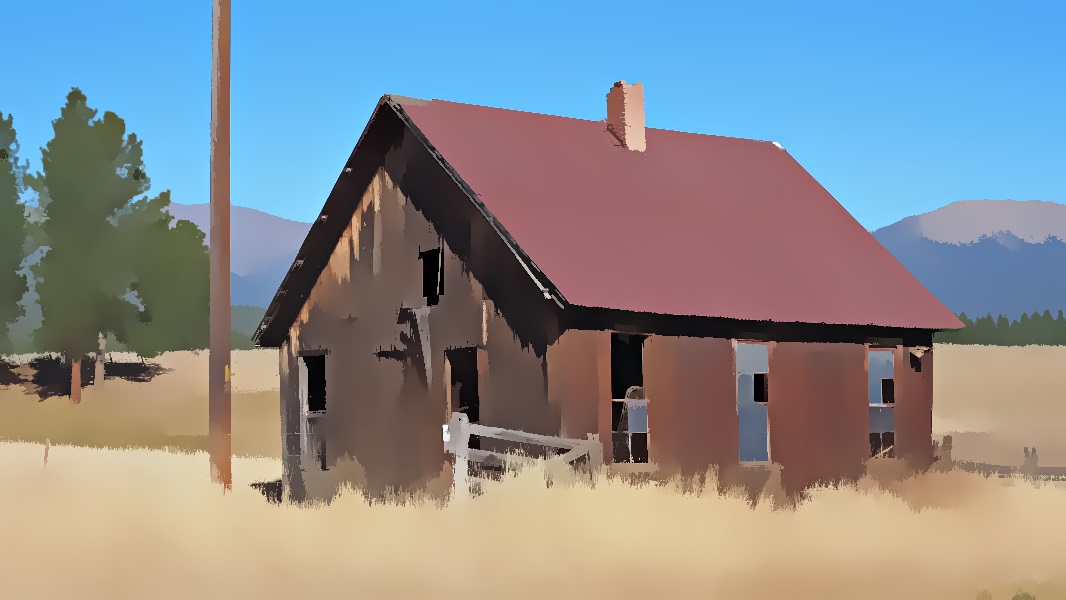}};
		\spy[width=1.0cm, height=0.8cm] on (0.35, 0.75) in node at (1.6, 0.8);
		\end{tikzpicture}  \\
		\vspace{-2mm}
		\caption{Clean input and our result }
	\end{subfigure}
	\begin{subfigure}[c]{0.236\textwidth}
		\centering
		\begin{tikzpicture}[
		spy using outlines={color=red, rectangle, magnification=3,
			every spy on node/.append style={rectangle},
			every spy in node/.append style={rectangle}}
		]
		\node[inner sep=0,outer sep=0]{\includegraphics[width=\linewidth]{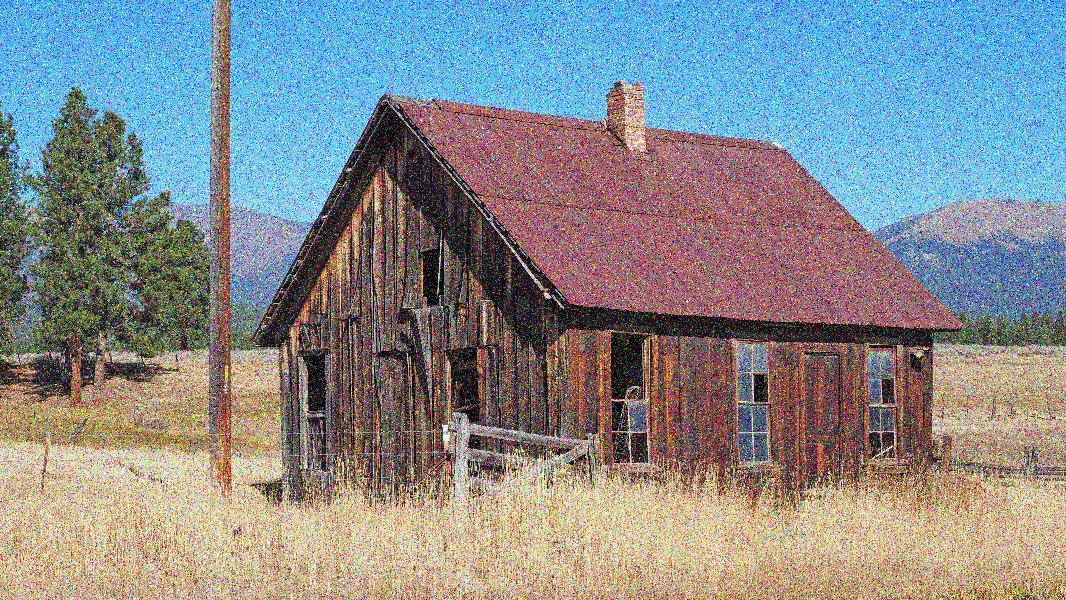}};
		\spy[width=1.0cm, height=0.8cm] on (0.35, 0.75) in node at (1.6, 0.8);
		\end{tikzpicture} \\ \vspace{2pt}
		\begin{tikzpicture}[
		spy using outlines={color=red, rectangle, magnification=3,
			every spy on node/.append style={rectangle},
			every spy in node/.append style={rectangle}}
		]
		\node[inner sep=0,outer sep=0]{\includegraphics[width=\linewidth]{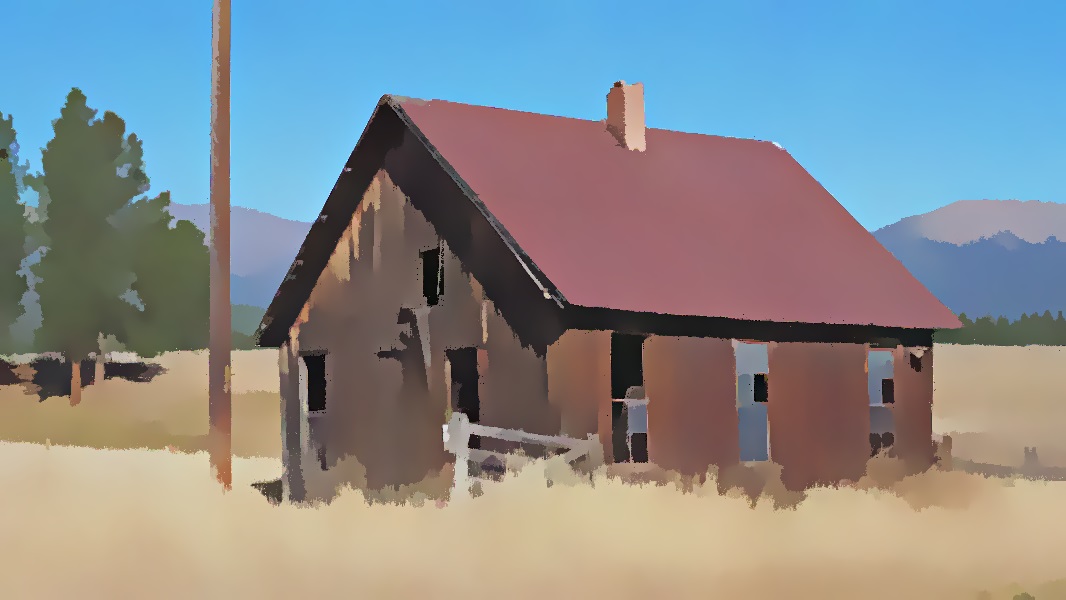}};
		\spy[width=1.0cm, height=0.8cm] on (0.35, 0.75) in node at (1.6, 0.8);
		\end{tikzpicture} \\
		\vspace{-2mm}
		\caption{Noisy input and our result}
	\end{subfigure}
	\vspace{-2mm}
	\caption{Effect of noise. Our method produces visually indistinguishable results for a clean image and its noisy counterpart. Image courtesy of Flickr user Eclectic Jack.} 
	\label{fig:noise}  %
\end{figure}

\begin{figure*}
	\centering
	\begin{subfigure}[c]{0.163\textwidth}
		\centering
		\includegraphics[width=\linewidth]{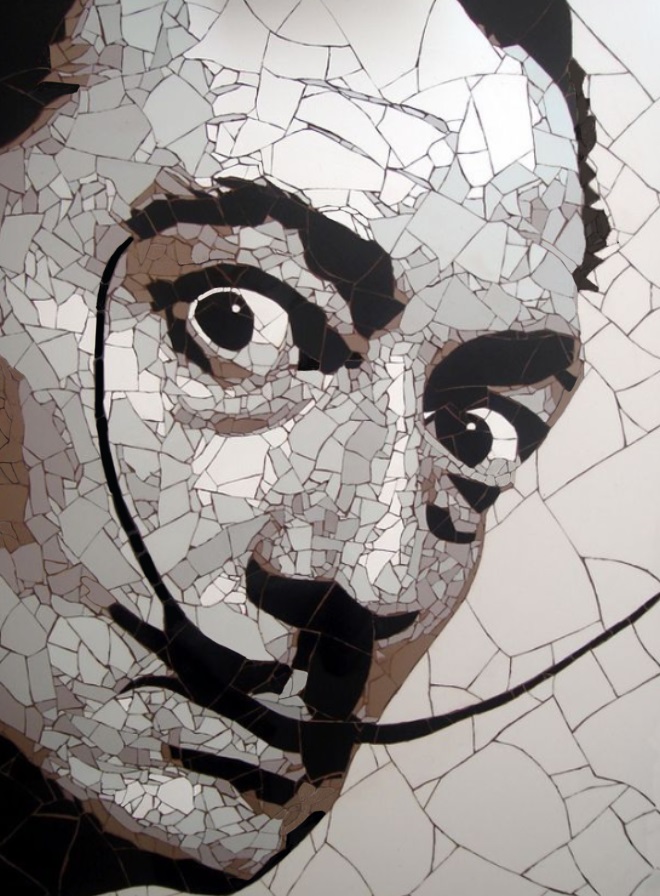}\\ \vspace{2pt}
		\includegraphics[width=\linewidth]{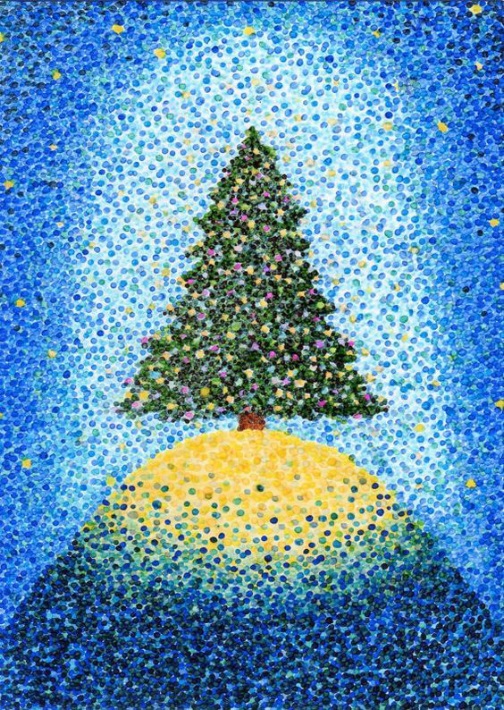} \\ \vspace{2pt}
		\includegraphics[width=\linewidth]{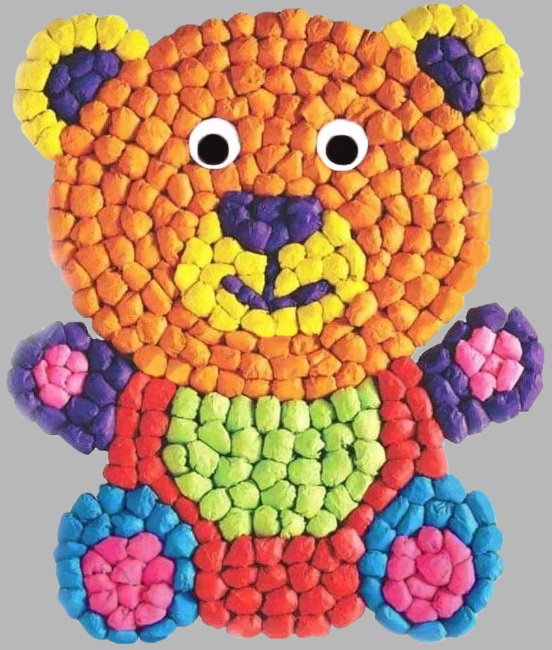} \\
		\vspace{-2mm}
		\caption{Input}
	\end{subfigure}
	\begin{subfigure}[c]{0.163\textwidth}
		\centering
		\includegraphics[width=\linewidth]{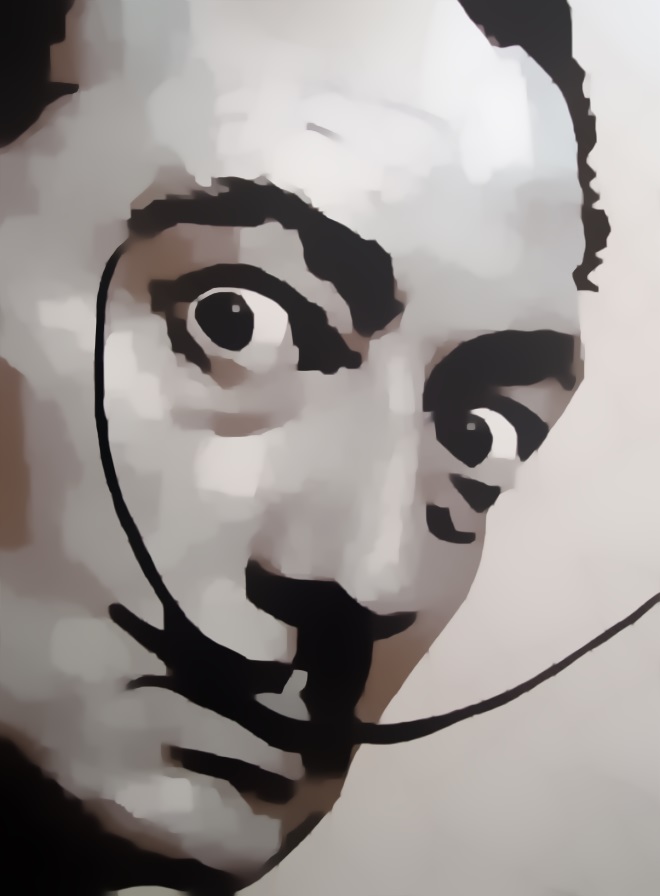}\\ \vspace{2pt}
		\includegraphics[width=\linewidth]{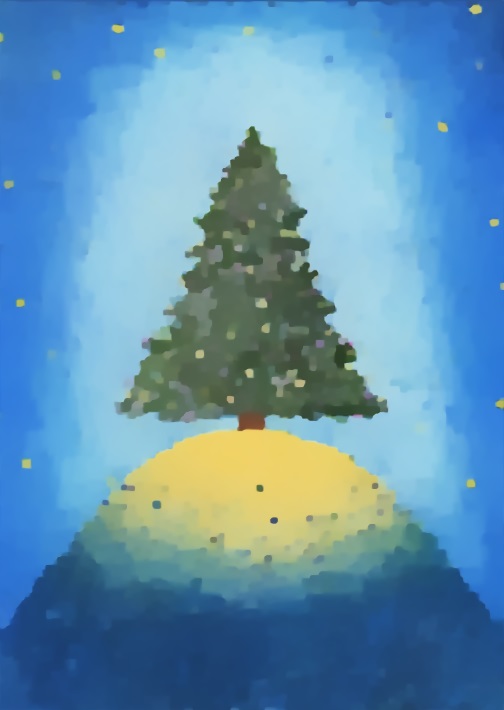} \\ \vspace{2pt}
		\includegraphics[width=\linewidth]{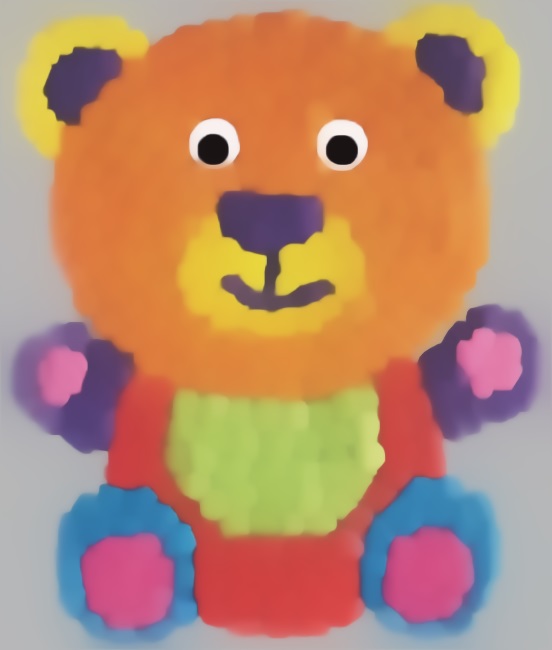} \\
		\vspace{-2mm}
		\caption{\cite{xu2012structure}}
	\end{subfigure}
	\begin{subfigure}[c]{0.163\textwidth}
		\centering
		\includegraphics[width=\linewidth]{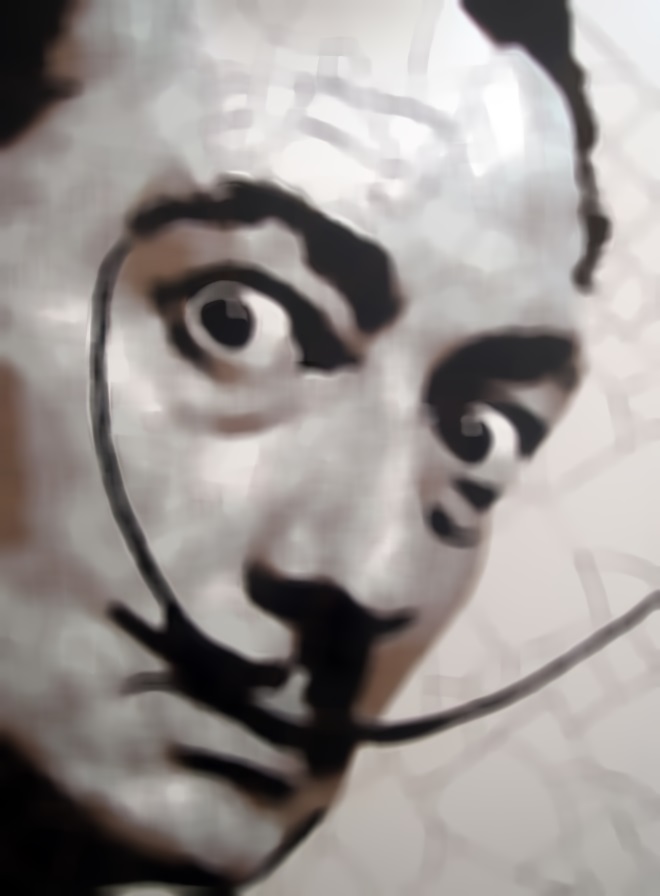}\\ \vspace{2pt}
		\includegraphics[width=\linewidth]{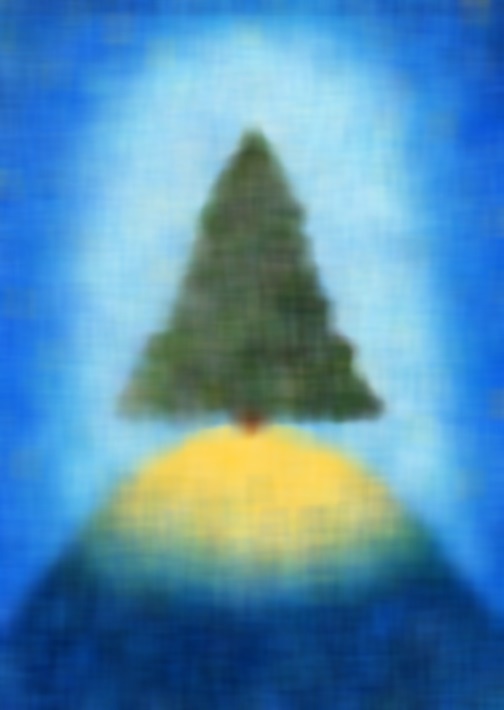} \\ \vspace{2pt}
		\includegraphics[width=\linewidth]{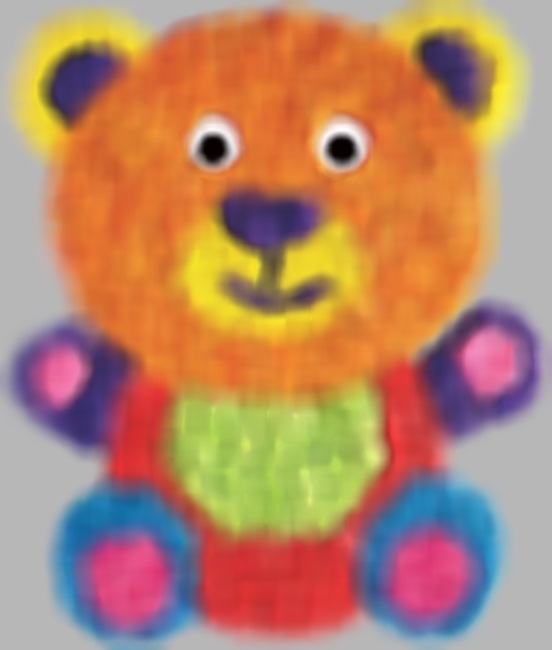} \\
		\vspace{-2mm}
		\caption{\cite{karacan2013structure}}
	\end{subfigure}
	\begin{subfigure}[c]{0.163\textwidth}
		\centering
		\includegraphics[width=\linewidth]{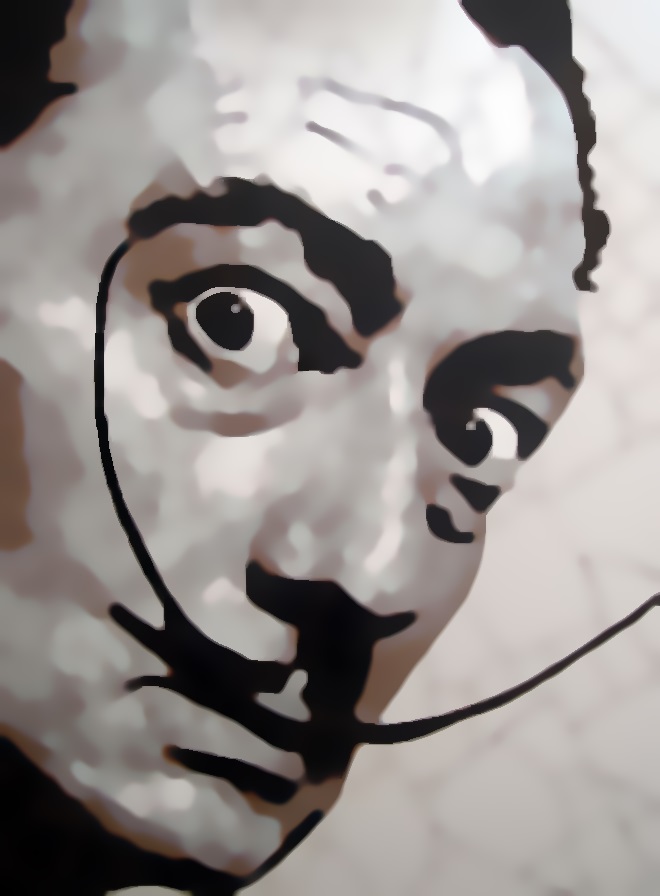}\\ \vspace{2pt}
		\includegraphics[width=\linewidth]{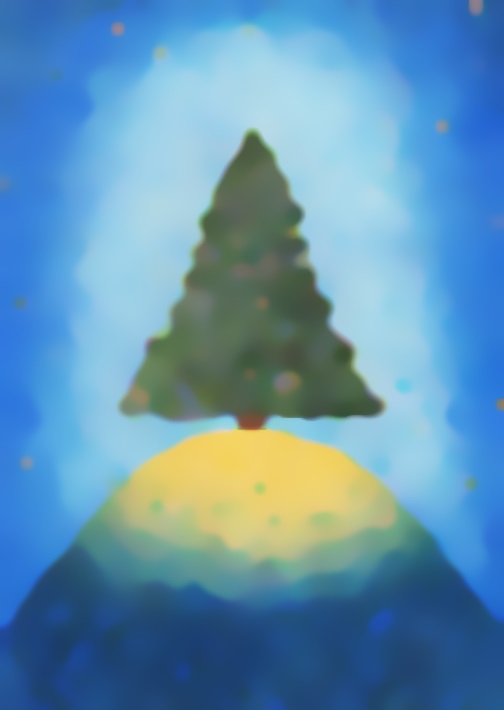} \\ \vspace{2pt}
		\includegraphics[width=\linewidth]{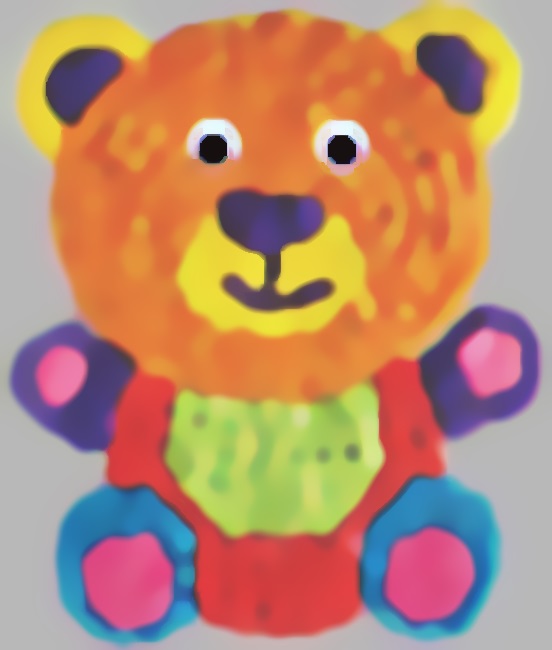}  \\
		\vspace{-2mm}
		\caption{\cite{cho2014bilateral}}
	\end{subfigure}
	\begin{subfigure}[c]{0.163\textwidth}
		\centering
		\includegraphics[width=\linewidth]{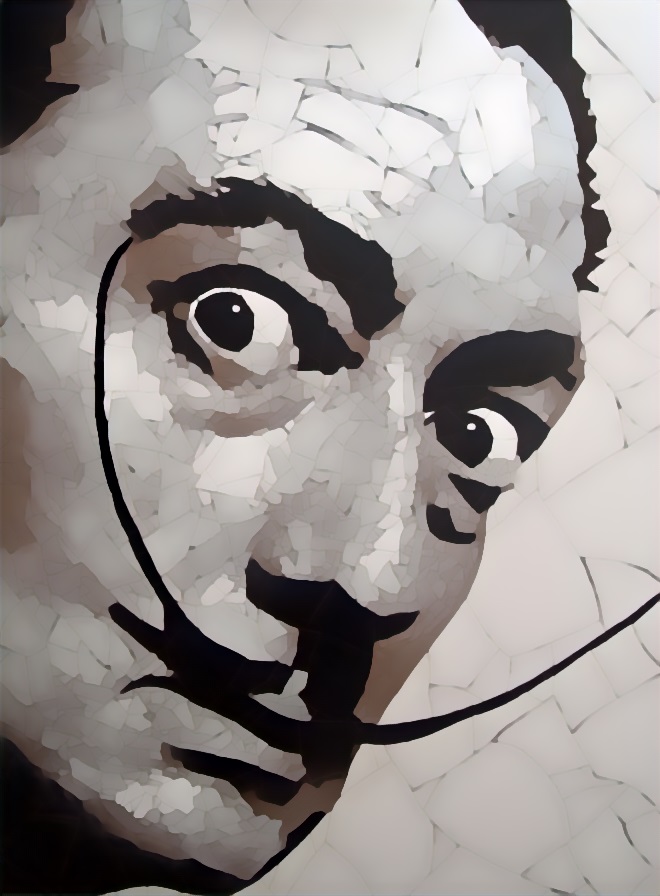}\\ \vspace{2pt}
		\includegraphics[width=\linewidth]{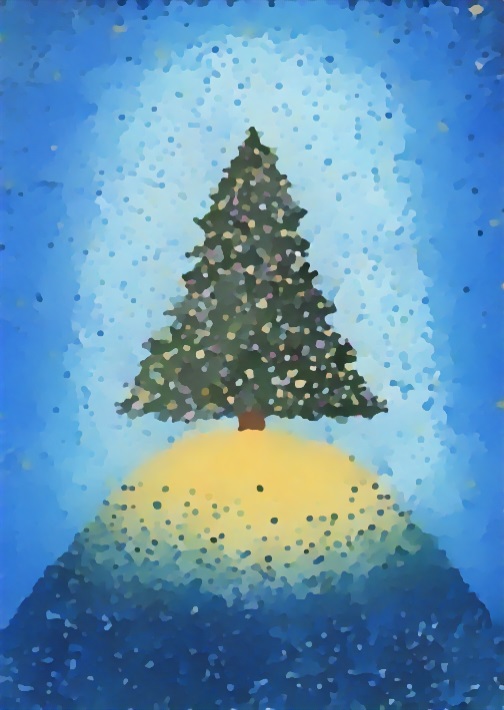} \\ \vspace{2pt}
		\includegraphics[width=\linewidth]{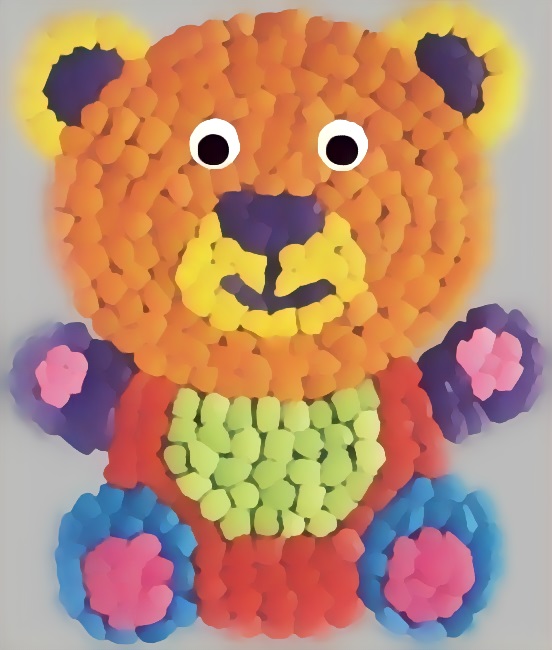} \\
		\vspace{-2mm}
		\caption{\cite{fan2018image}}
	\end{subfigure}
	\begin{subfigure}[c]{0.163\textwidth}
		\centering
		\includegraphics[width=\linewidth]{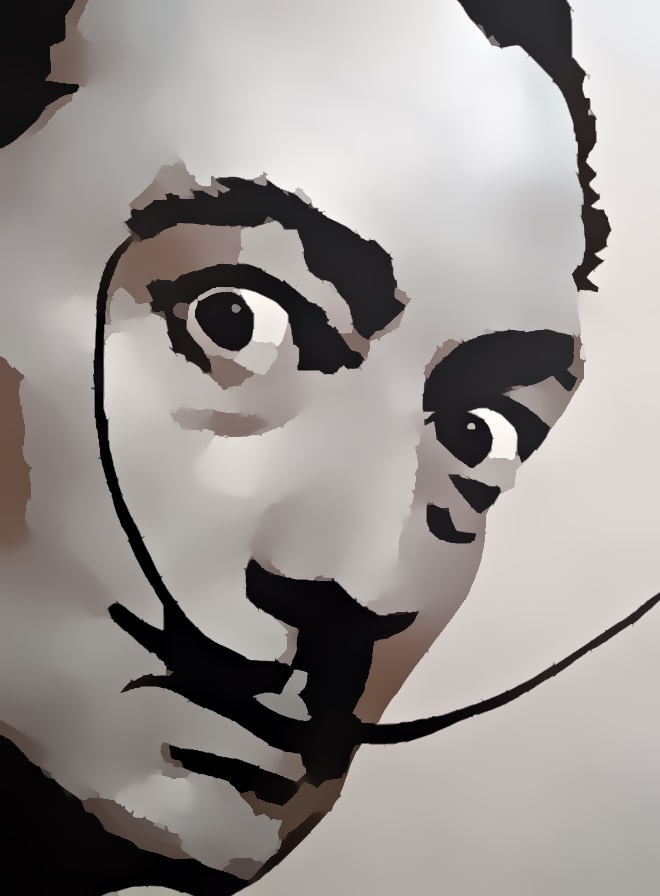}\\ \vspace{2pt}
		\includegraphics[width=\linewidth]{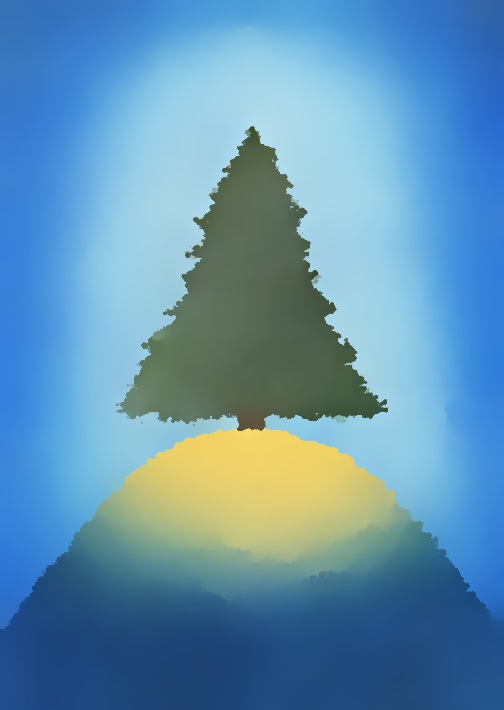} \\ \vspace{2pt}
		\includegraphics[width=\linewidth]{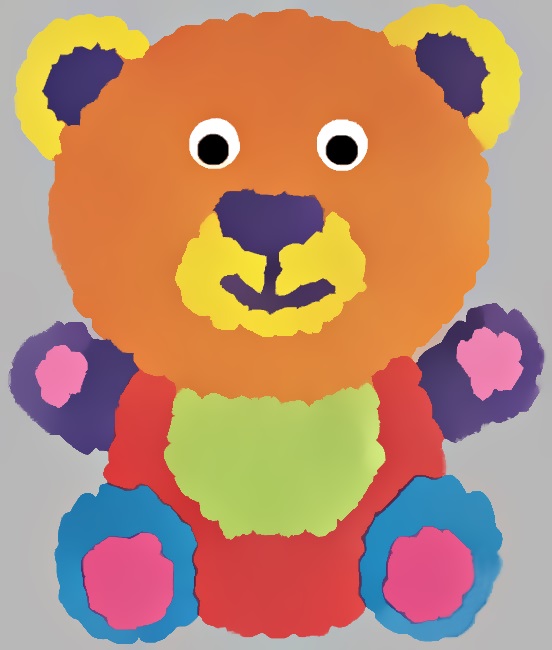} \\
		\vspace{-2mm}
		\caption{Ours}
	\end{subfigure}
	\vspace{-2mm}
	\caption{Comparison with previous methods on texture smoothing. Parameters: \cite{xu2012structure} \{1st: ($\lambda=0.03$, $\sigma=6$), 2nd: ($\lambda=0.04$, $\sigma=7$), 3rd: ($\lambda=0.04$, $\sigma=9$)\}, \cite{karacan2013structure} \{1st: ($k=13$, $\sigma=0.5$, Model 1), 2nd: ($k=11$, $\sigma=0.7$, Model 1), 3rd: ($k=11$, $\sigma=0.5$, Model 1)\}, \cite{cho2014bilateral} \{1st: ($k=9$, $n_{itr}=5$), 2nd: ($k=11$, $n_{itr}=7$), 3rd: ($k=13$, $n_{itr}=7$)\}, and our method \{1st: ($\sigma_s=7$, $\sigma_r=0.07$), 2nd: ($\sigma_s=5$, $\sigma_r=0.09$), 3rd: ($\sigma_s=9$, $\sigma_r=0.07$)\}. Results of \cite{fan2018image} are produced by a trained model released by the authors. Images courtesy of Ed Chapman, Flickr user Allison Blacker, and Snapdeal.} 
	\label{fig:compare_sota}
\end{figure*}

Specifically, we fix all the range parameters to $\sigma_{r}$ to ensure that image edges across different scales are always treated the same. Suppose that $\sigma_{s,k}$ denotes the spatial parameter utilized in both Eqs.~\eqref{equ:jbu} and \eqref{equ:jbf} in our upsampling at scale $k$ ($0 \leq k < N$) with $R_{k}$ as output, at the finest scale $k=0$ we set $\sigma_{s,0} = \sigma_{s}$. The spatial parameters of the subsequent coarse-scale ($k \geq 1$) upsampling are then set as $\sigma_{s,k} = \sigma_{s,0} / 2^k$ to adapt to scale changes. To strengthen structure refinement while maintaining good texture removal effect, we empirically set the neighborhood sizes (no less than $3 \times 3$) in Eqs.~\eqref{equ:jbu} and \eqref{equ:jbf} in our upsampling at scale $k$ as the odd values closest to $\max(\sigma_{s,k}, 3)$ and $\max(4\sigma_{s,k},3)$, respectively. Setting a small neighborhood for Eq.~\eqref{equ:jbu} is to reduce the number of pixels involved in the weighted average and avoid bringing back textures from the guided Gaussian pyramid level at a finer scale, while setting a large neighborhood for Eq.~\eqref{equ:jbf} aims to ensure that textures introduced by the Laplacian pyramid level are effectively removed. Note, as the guidance image $\hat{R}_k$ in Eq.~\eqref{equ:jbf} is texture-free, a large neighborhood here will not recover unwanted textures as in Eq.~\eqref{equ:jbu}.

\subsection{Implementation} \label{sec:implementation}
The implementation of our approach is summarized in Algorithm~\ref{alg:ptf}. We build standard image pyramids with a downsampling rate of 1/2 based on Gaussian smoothing with $5 \times 5$ kernel and a standard deviation of 1, and empirically limit the long axis of the coarsest Gaussian pyramid level to [32, 64) to set the pyramid depth. The down- and up-sampling operations in pyramid construction are achieved by bilinear interpolation. We normalize all pixel values to [0,1], and use $\sigma_s \in [3, 15]$ and $\sigma_r \in [0.02, 0.09]$ to produce all results in the paper. We find that $\sigma_s = 5$ and $\sigma_r = 0.07$ are good starting parameters for most images. As shown in Figure~\ref{fig:varying_parameters}, large $\sigma_s$ and $\sigma_r$ help increase the effectiveness in removing large-scale and high-contrast textures, respectively. Note, different from standard joint bilateral filtering in which the range smoothing weight is independently computed for each RGB color channel, a three-channel shared range weight based on the Euclidean distance between two RGB color vectors is adopted in our method for enhancing structure preservation (see the supplementary material for validation). Parameters for all images in the paper are given in the supplementary material.

\section{More Analysis}

\begin{figure*}
	\centering
	\captionsetup[subfigure]{labelformat=empty}
	\begin{subfigure}[c]{0.163\textwidth}
		\centering
		\includegraphics[width=1.14in]{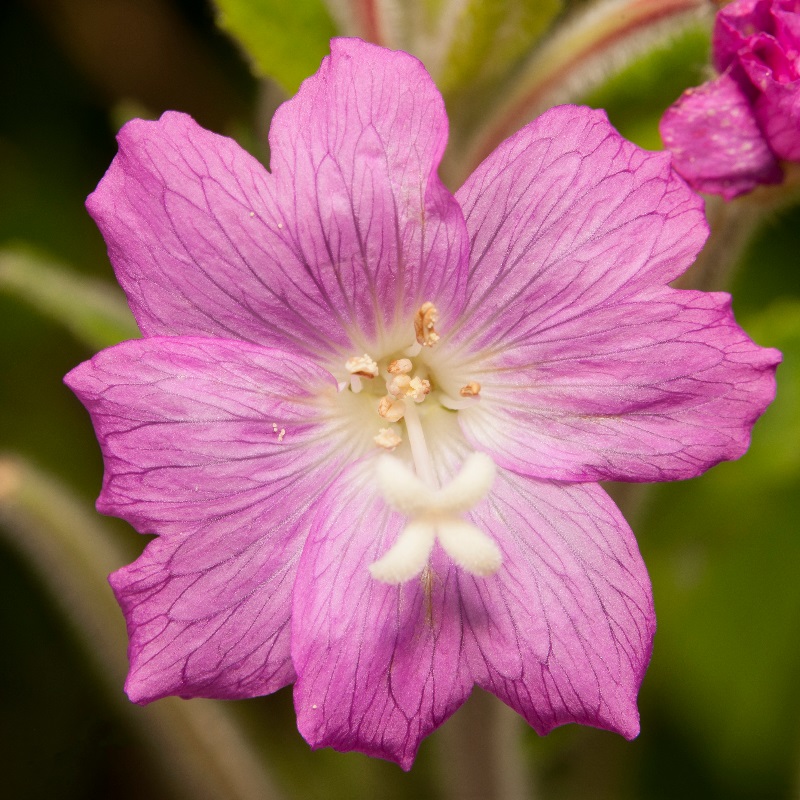}\\ 
		\vspace{-2mm}
		\caption{Input}
	\end{subfigure}
	\begin{subfigure}[c]{0.163\textwidth}
		\centering
		\includegraphics[width=1.14in]{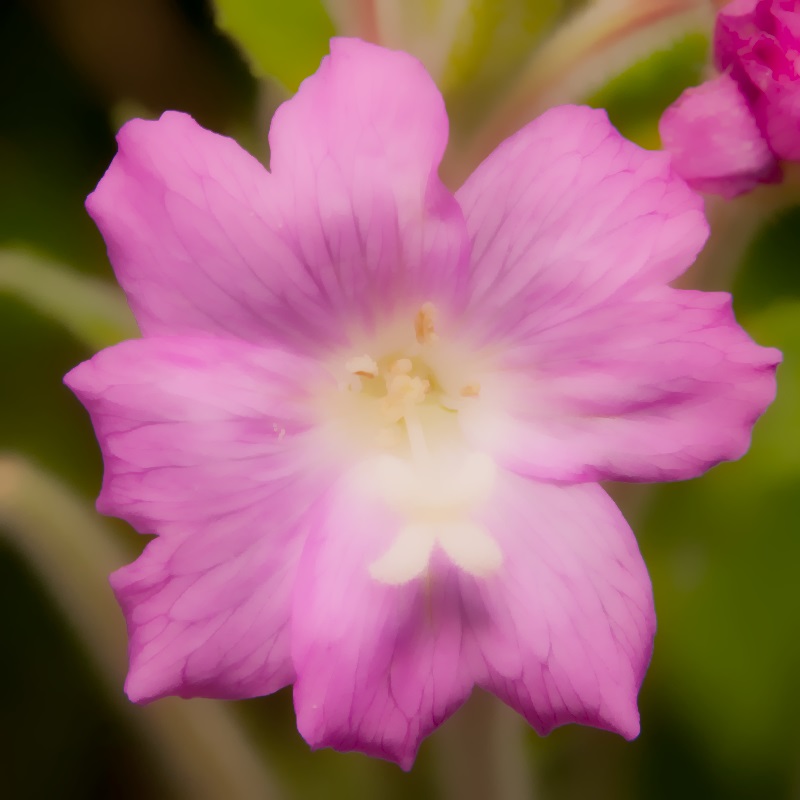} \\ \vspace{2pt}
		\includegraphics[width=1.14in]{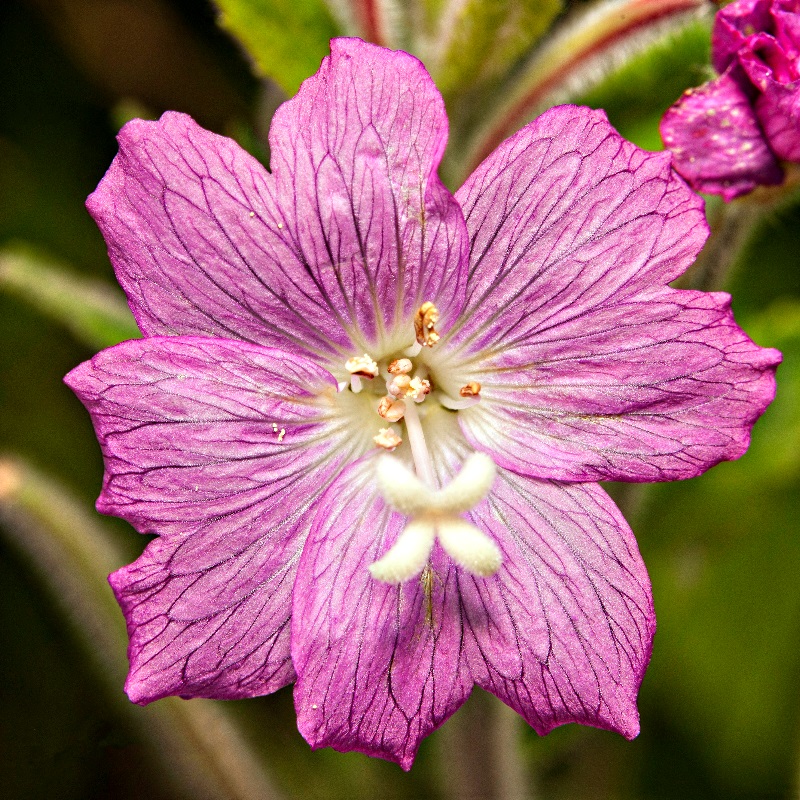} \\
		\vspace{-2mm}
		\caption{WLS ($\lambda=1$, $\alpha=1.2$)}
	\end{subfigure}
	\begin{subfigure}[c]{0.163\textwidth}
		\centering
		\includegraphics[width=1.14in]{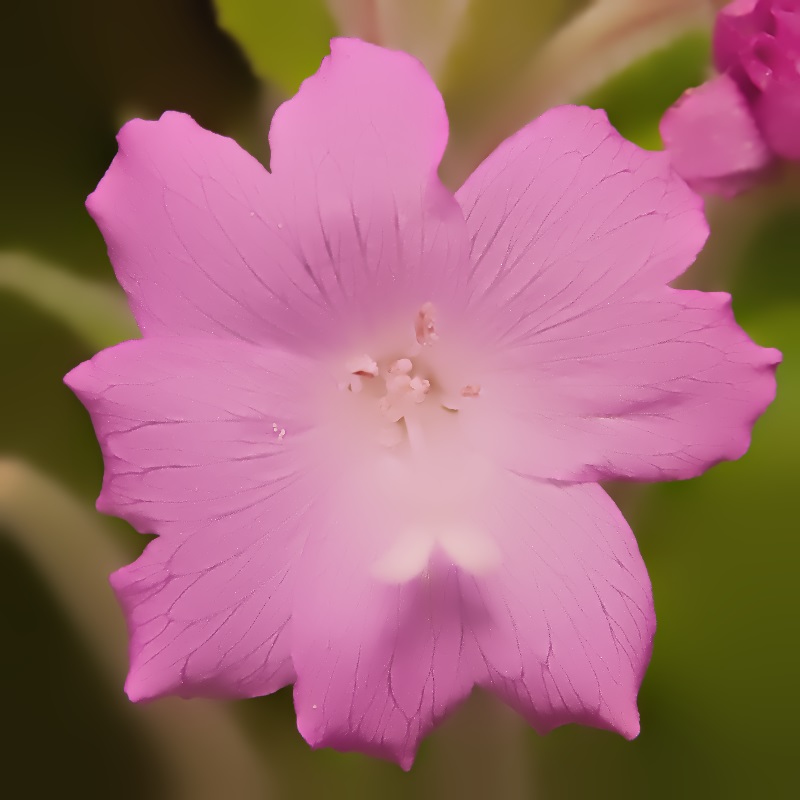}\\ \vspace{2pt}
		\includegraphics[width=1.14in]{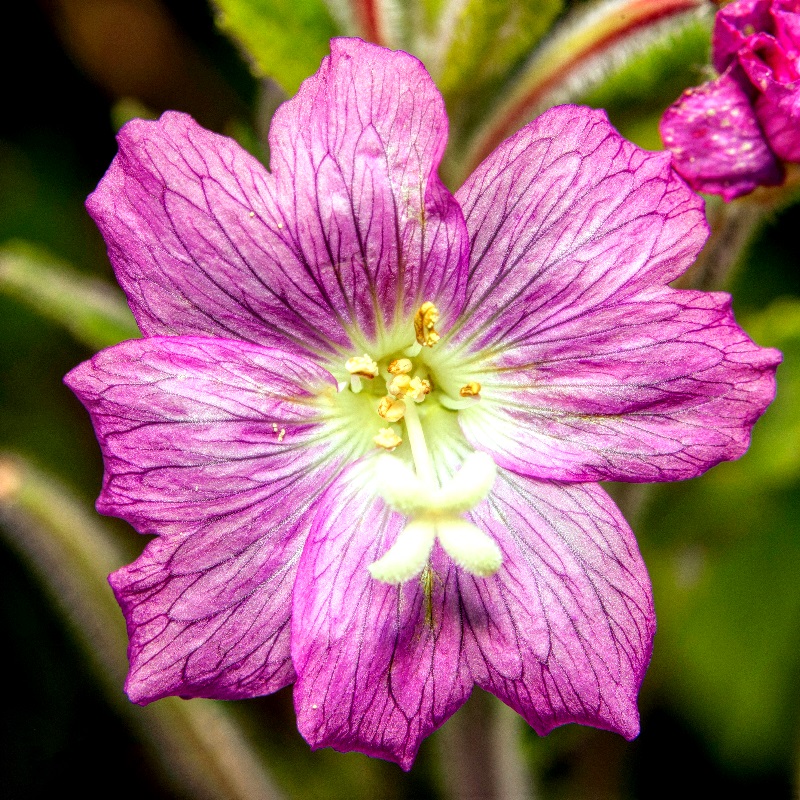}\\
		\vspace{-2mm}
		\caption{LLF ($\sigma_r\!=\!0.4$, $\alpha\!=\!4$, $\beta\!=\!1$)}
	\end{subfigure}
	\begin{subfigure}[c]{0.163\textwidth}
		\centering
		\includegraphics[width=1.14in]{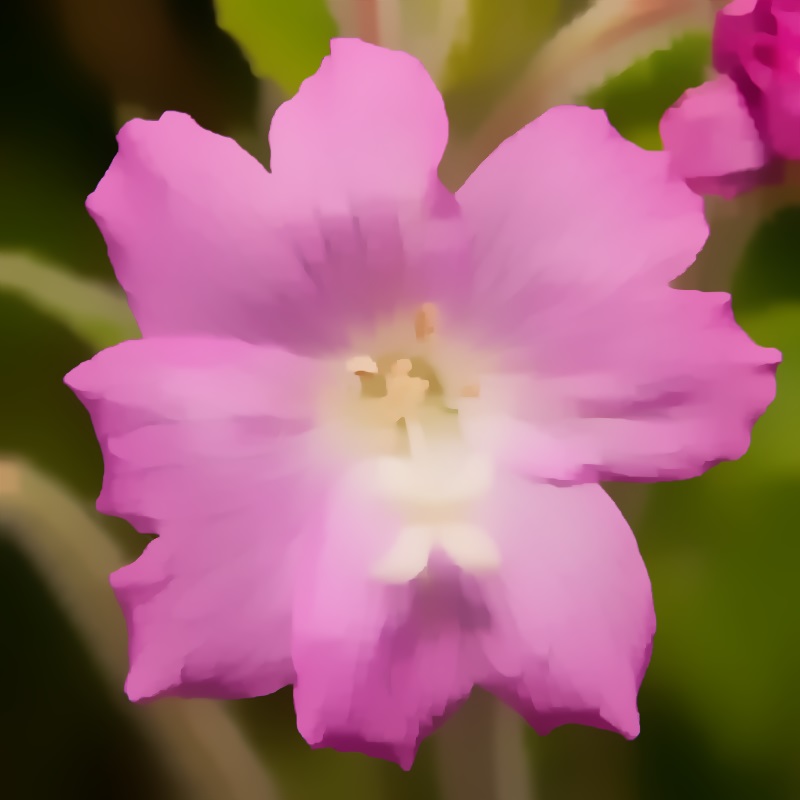}\\ \vspace{2pt}
		\includegraphics[width=1.14in]{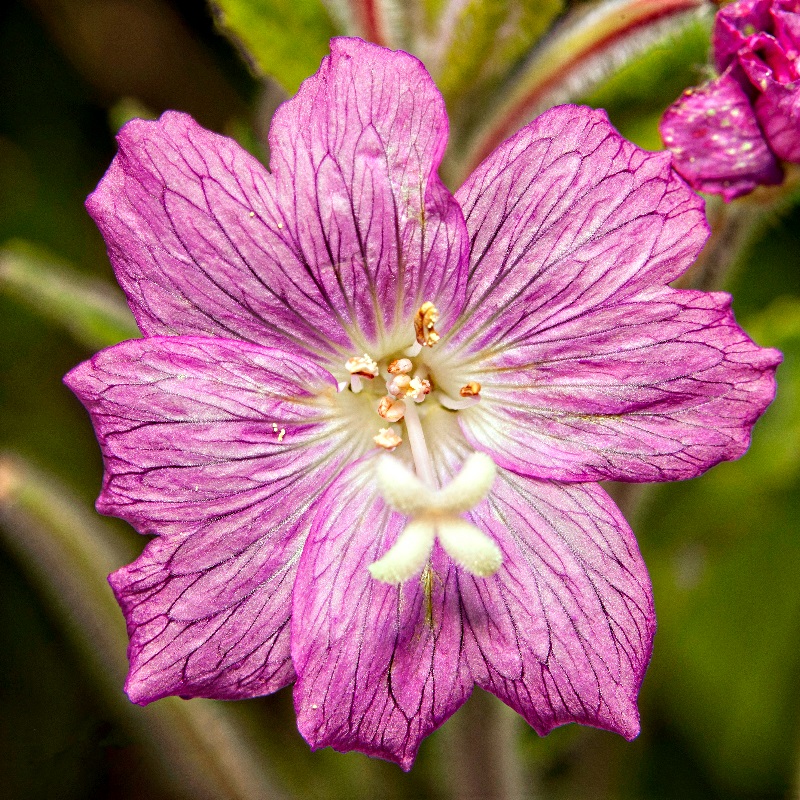}\\
		\vspace{-2mm}
		\caption{RTV ($\lambda\!=\!0.015$, $\sigma\!=\!3$)}
	\end{subfigure}
	\begin{subfigure}[c]{0.163\textwidth}
		\centering
		\includegraphics[width=1.14in]{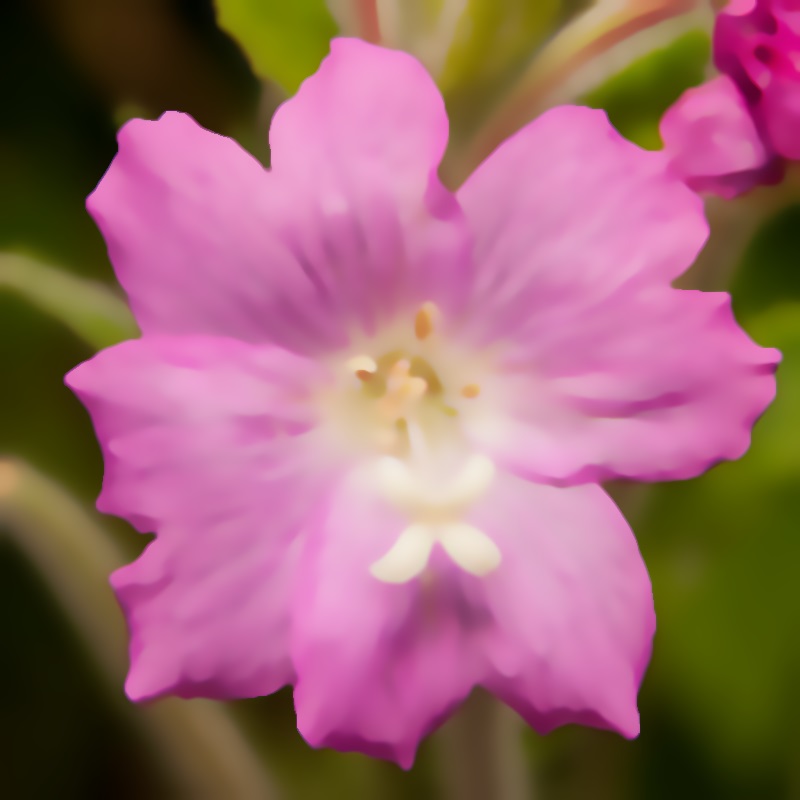}\\ \vspace{2pt}
		\includegraphics[width=1.14in]{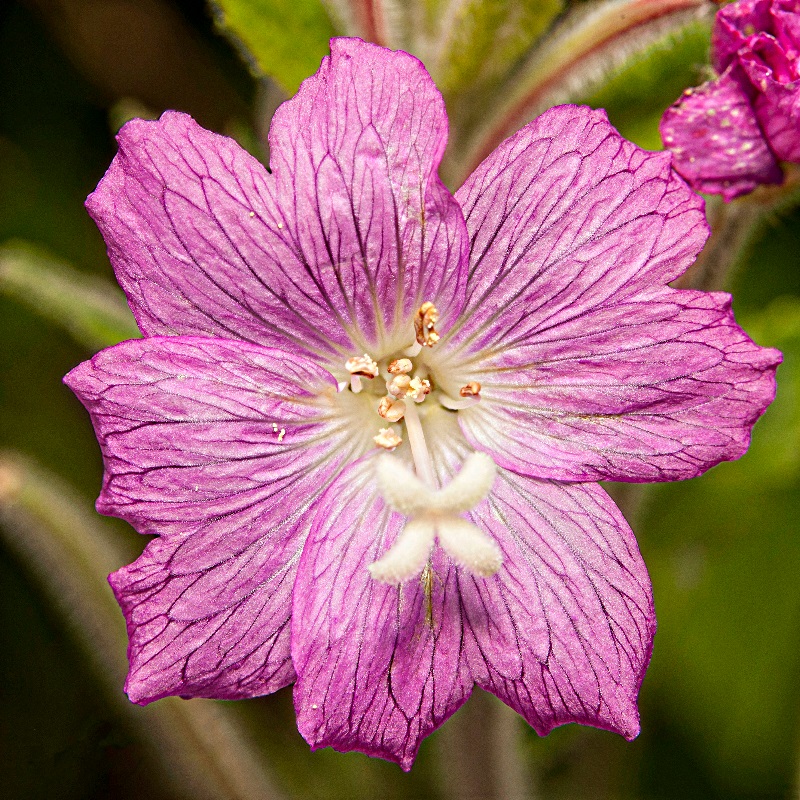}\\
		\vspace{-2mm}
		\caption{BTF ($k\!=\!7$, $n_{itr}\!=\!5$)}
	\end{subfigure}
	\begin{subfigure}[c]{0.163\textwidth}
		\centering
		\includegraphics[width=1.14in]{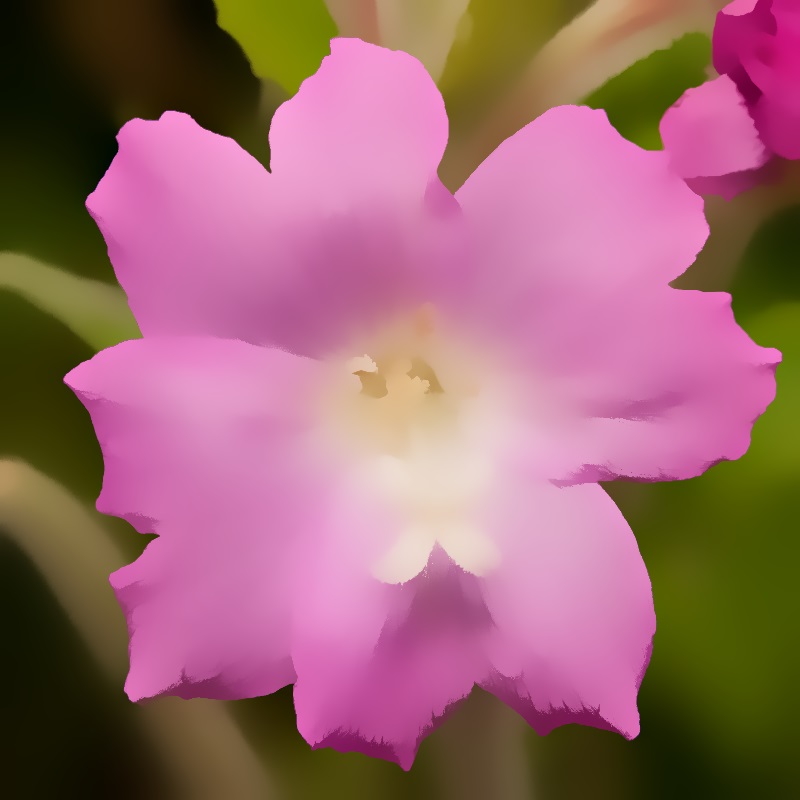}\\ \vspace{2pt}
		\includegraphics[width=1.14in]{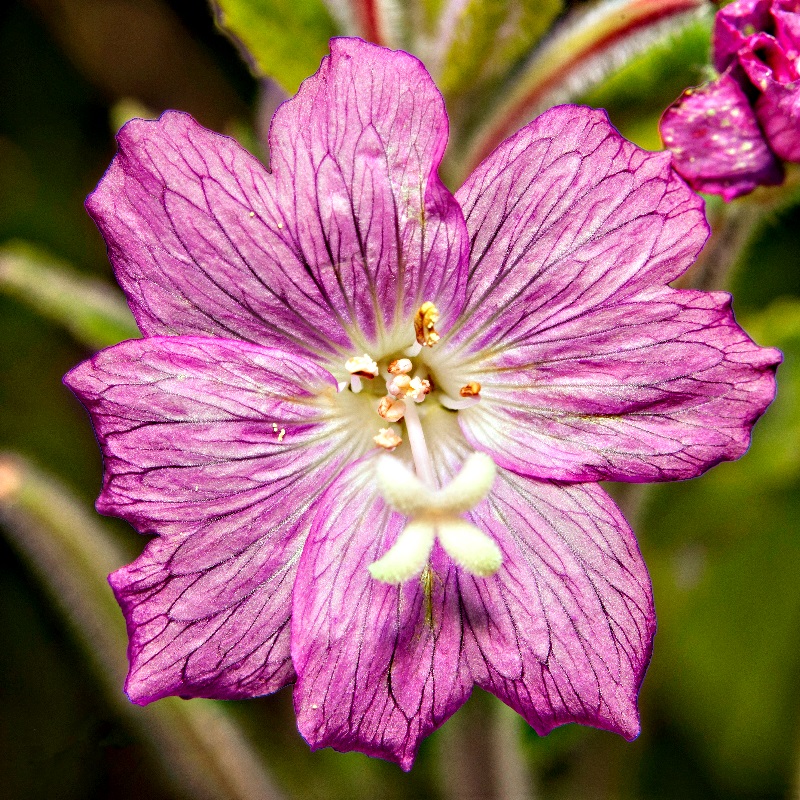}\\
		\vspace{-2mm}
		\caption{Ours ($\sigma_s\!=\!5$, $\sigma_r\!=\!0.07$)}
	\end{subfigure}
	\vspace{-2mm}
	\caption{Detail enhancement of the input image using our approach and four previous methods including WLS \cite{farbman2008edge}, LLF \cite{paris2011local}, RTV \cite{xu2012structure}, and BTF \cite{cho2014bilateral}. Apart from the first column, the top row shows the smoothing results produced by different methods, while the bottom gives the corresponding $2.5 \times$ detail enhancement outputs. Image courtesy of Flickr user Charlene Watt.} 
	\label{fig:detail_enhancement}
\end{figure*}

\paragraph{Effect of different Laplacian pyramid usages} In contrast to the operation in Eq.~\eqref{equ:jbf}, another possible Laplacian pyramid usage is to firstly smooth out the unwanted texture details from $L_{k-1}$ by performing joint bilateral filtering using the initial upsampling output $\hat{R}_{k-1}$ generated from Eq.~\eqref{equ:jbu} as guidance, and then add the filtering result back to $\hat{R}_{k-1}$ for producing $R_{k-1}$, i.e., $R_{k-1} = \hat{R}_{k-1} + \textrm{JBF}(L_{k-1}, \hat{R}_{k-1})$. As shown in Figure~\ref{fig:ablation}(e), this method works well for structure preservation, but is not robust enough for texture removal and may produce result with a small amount of texture residuals.

\paragraph{Difference from prior scale-aware methods} Unlike previous scale-aware texture smoothing methods \cite{zhang2014rolling,du2016two,jeon2016scale} which basically iterate between Gaussian smoothing and joint bilateral filtering in an alternating order at a fixed scale, our method is built upon multi-scale representations in the form of image pyramids and achieves texture smoothing by iteratively upsampling the coarsest level in Gaussian pyramid under the guidance of other levels in both Gaussian and Laplacian pyramids. As shown in Figure~\ref{fig:comp_scale_aware}, our method clearly outperforms these methods in texture removal and structure preservation. Please see the supplementary material for more comparison results.

\paragraph{Necessity of image pyramids} To verify the necessity of image pyramids, we compare our approach with a variant that replaces Gaussian and Laplacian pyramids with a sequence of iteratively Gaussian blurred images (same to the number of coarse levels in Gaussian pyramid) of the original image and a series of difference images between successive Gaussian blurred images, and accordingly modifies Eq.~\eqref{equ:jbu} as joint bilateral filtering without upsampling. As shown in Figure~\ref{fig:no_pyramid}(b), the variant produces result with blurred and distorted structures due to the structure degradation caused by Gaussian smoothing. In contrast, our method achieves result with sharp structures, as the same structure degradation issue arising from Gaussian smoothing in pyramid construction can be effectively alleviated by image downsampling.

\paragraph{Effect of different pyramid depths} Figure~\ref{fig:diff_pyramid_depth} shows how the depth of pyramid affects the results. As can be seen, deeper pyramid with smaller coarsest level helps produce results with stronger texture removal effect, especially for large-scale textures. The reason is that smaller coarsest level summarizes image information at a larger scale, where textures are more likely to be eliminated due to scale differences. However, this trend becomes less obvious when the coarsest level is smaller than $40 \times 40$, since our method is able to remove all the textures based on a pyramid with $40 \times 40$ coarsest level. The above experiment indicates that deeper pyramid can be set for large-scale textures, while relatively shallower pyramid can be used for small-scale textures. As introduced in Section~\ref{sec:implementation}, our experience is that a coarsest level with at least 32 but less than 64 pixels on its long axis is generally sufficient for most images.

\begin{figure}
	\centering
	\captionsetup[subfigure]{labelformat=empty}
	\begin{subfigure}[c]{0.156\textwidth}
		\centering
		\includegraphics[width=\linewidth]{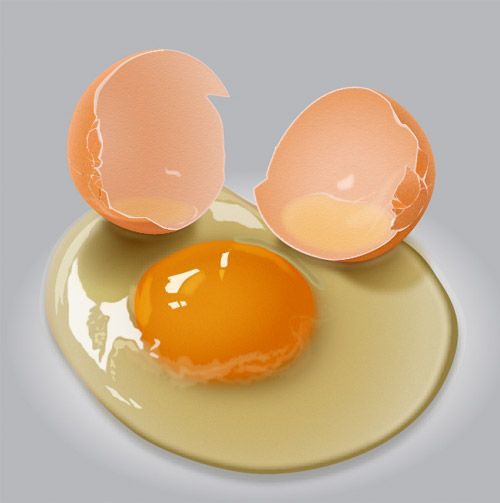} \\  \vspace{2pt}
		\includegraphics[width=\linewidth]{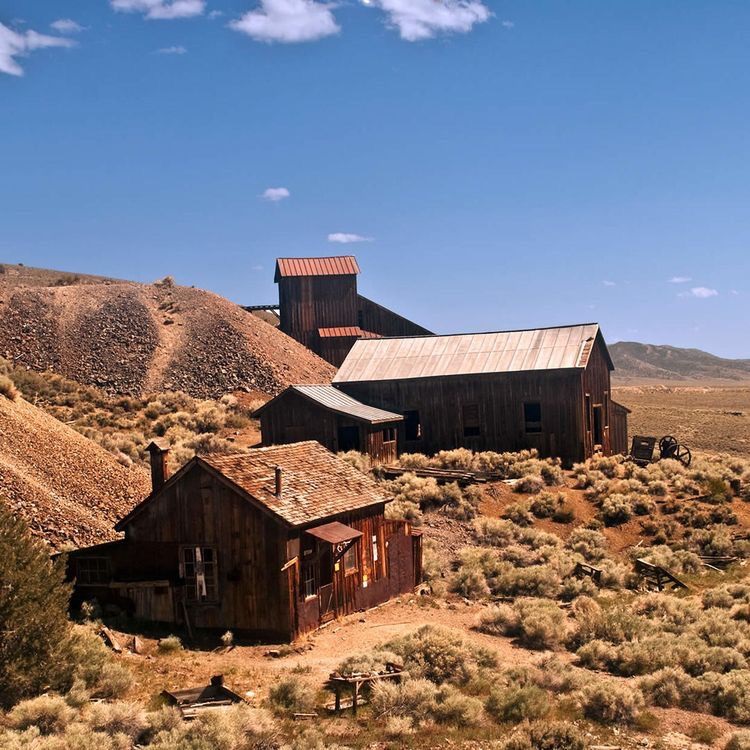} \\
		\vspace{-2mm}
		\caption{Input}
	\end{subfigure}
	\begin{subfigure}[c]{0.156\textwidth}
		\centering
		\includegraphics[width=\linewidth]{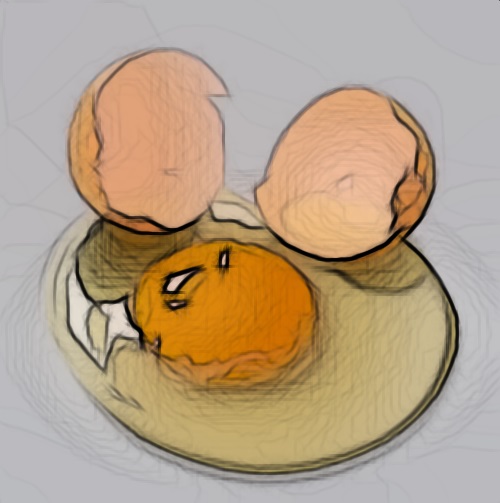} \\  \vspace{2pt}
		\includegraphics[width=\linewidth]{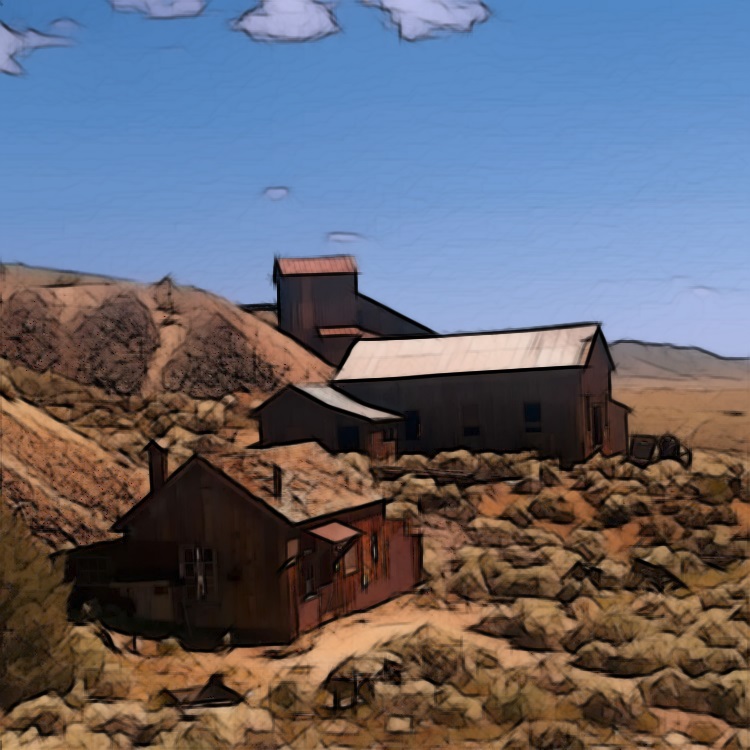}\\
		\vspace{-2mm}
		\caption{WLS ($\lambda\!=\!1$, $\alpha\!=\!1.2$)}
	\end{subfigure}
	\begin{subfigure}[c]{0.156\textwidth}
		\centering
		\includegraphics[width=\linewidth]{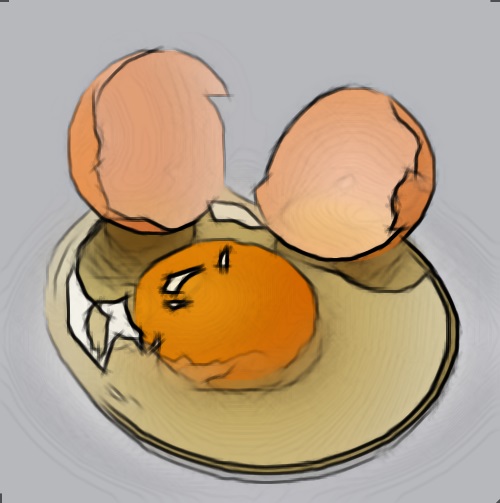} \\ \vspace{2pt}
		\includegraphics[width=\linewidth]{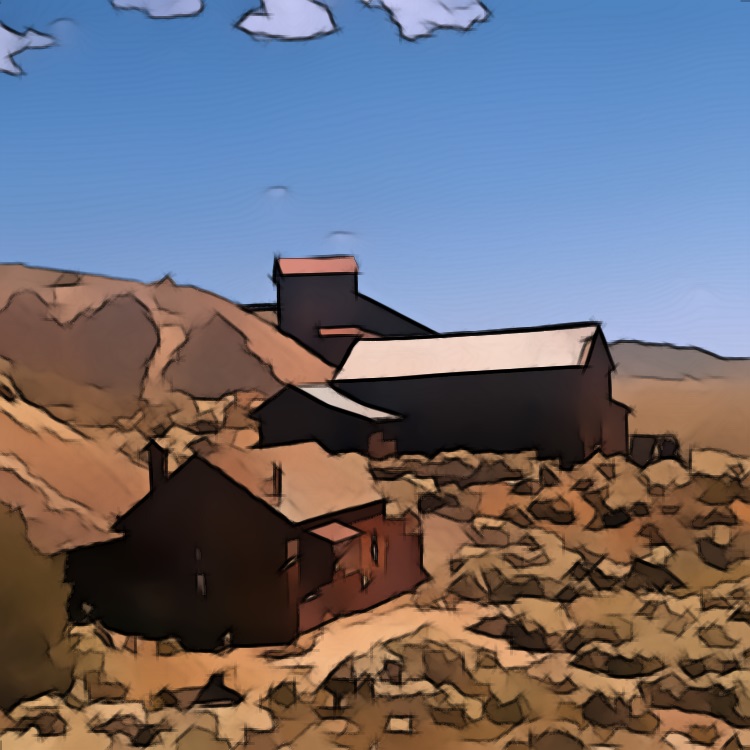}\\
		\vspace{-2mm}
		\caption{Ours ($\sigma_s\!=\!3$, $\sigma_r\!=\!0.03$)}
	\end{subfigure}
	\vspace{-2mm}
	\caption{Image abstraction by WLS \cite{farbman2008edge} and our method. Images courtesy of Vector Diary and Fotospot.}
	\label{fig:abstraction} 
\end{figure}

\paragraph{Effect of multiple smoothing} We show in Figure~\ref{fig:multi_filtering} that multiple smoothing with the same parameters barely changes the result anymore when single smoothing can remove all the textures, which further verifies our effectiveness in structure preservation.

\begin{figure*}
	\centering
	\captionsetup[subfigure]{labelformat=empty}
	\begin{subfigure}[c]{0.246\textwidth}
		\centering
		\includegraphics[width=\linewidth]{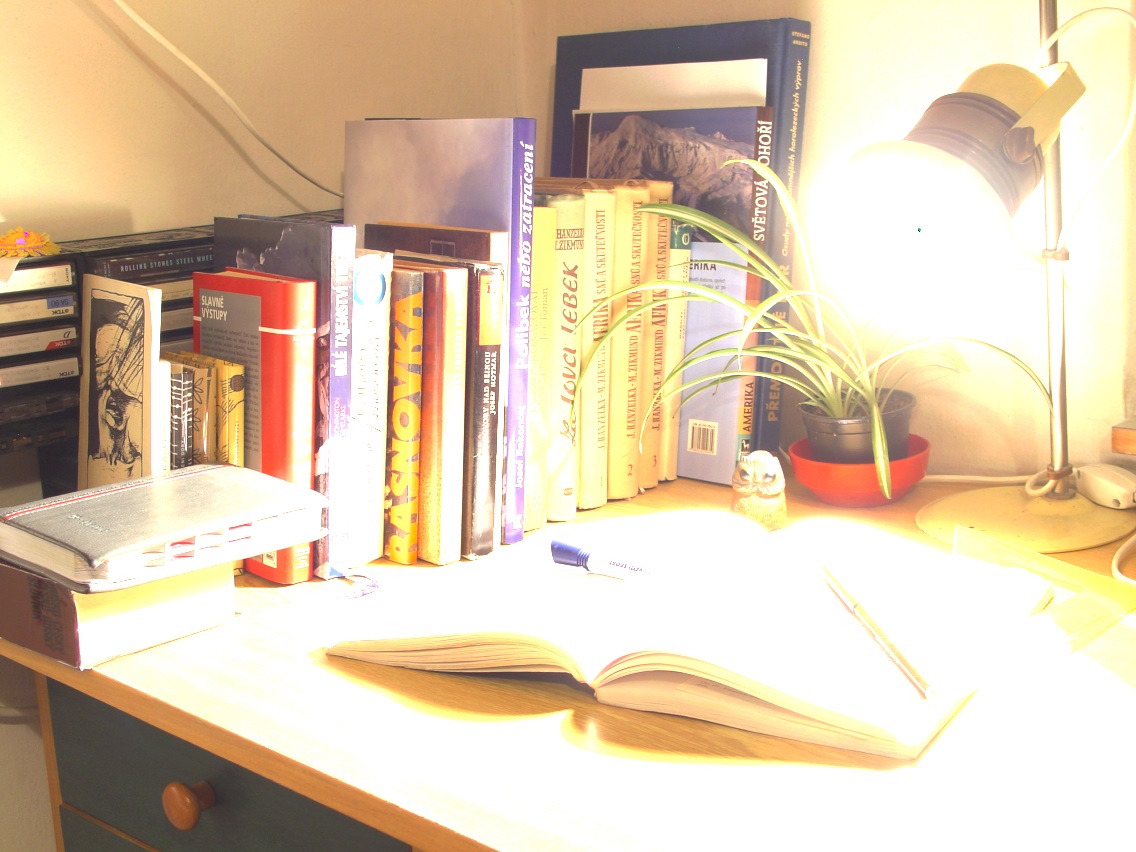}\\
		\vspace{-2mm}
		\caption{Input}
	\end{subfigure}
	\begin{subfigure}[c]{0.246\textwidth}
		\centering
		\includegraphics[width=\linewidth]{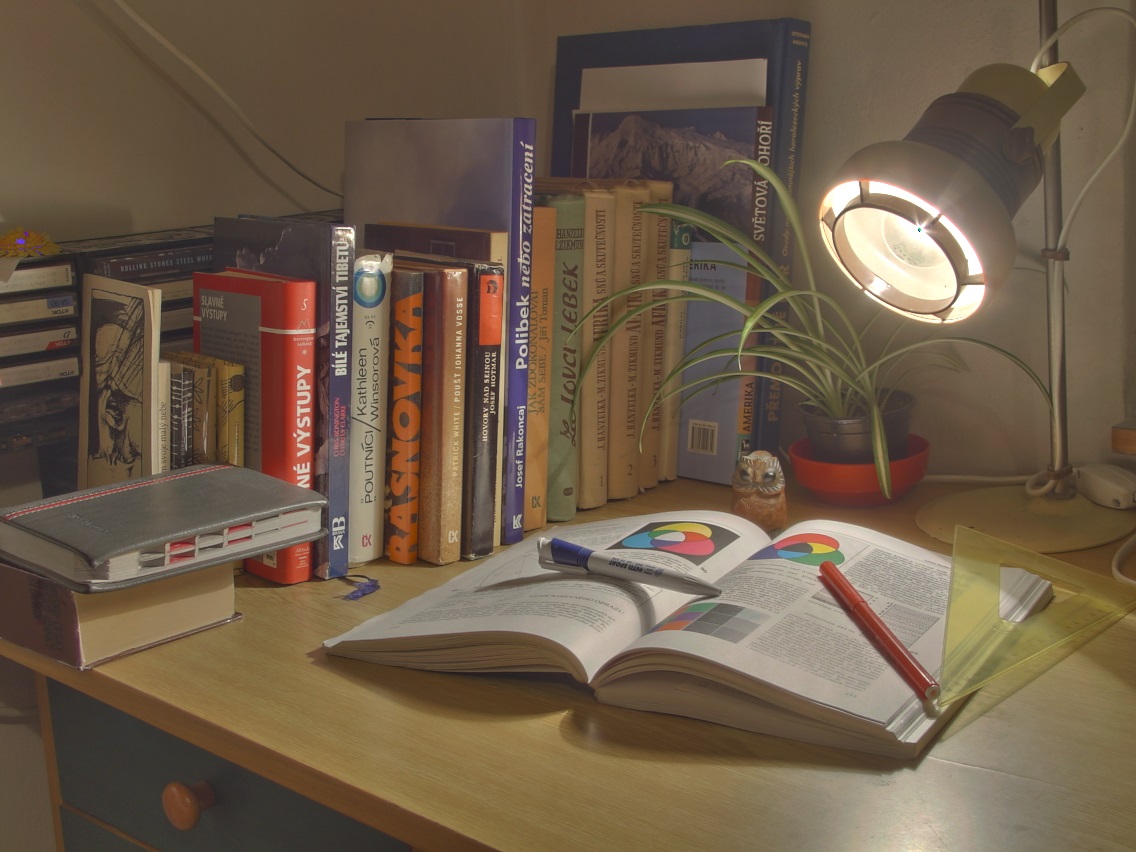}\\
		\vspace{-2mm}
		\caption{WLS ($\lambda=1$, $\alpha=1.2$)}
	\end{subfigure}  
	\begin{subfigure}[c]{0.246\textwidth}
		\centering
		\includegraphics[width=\linewidth]{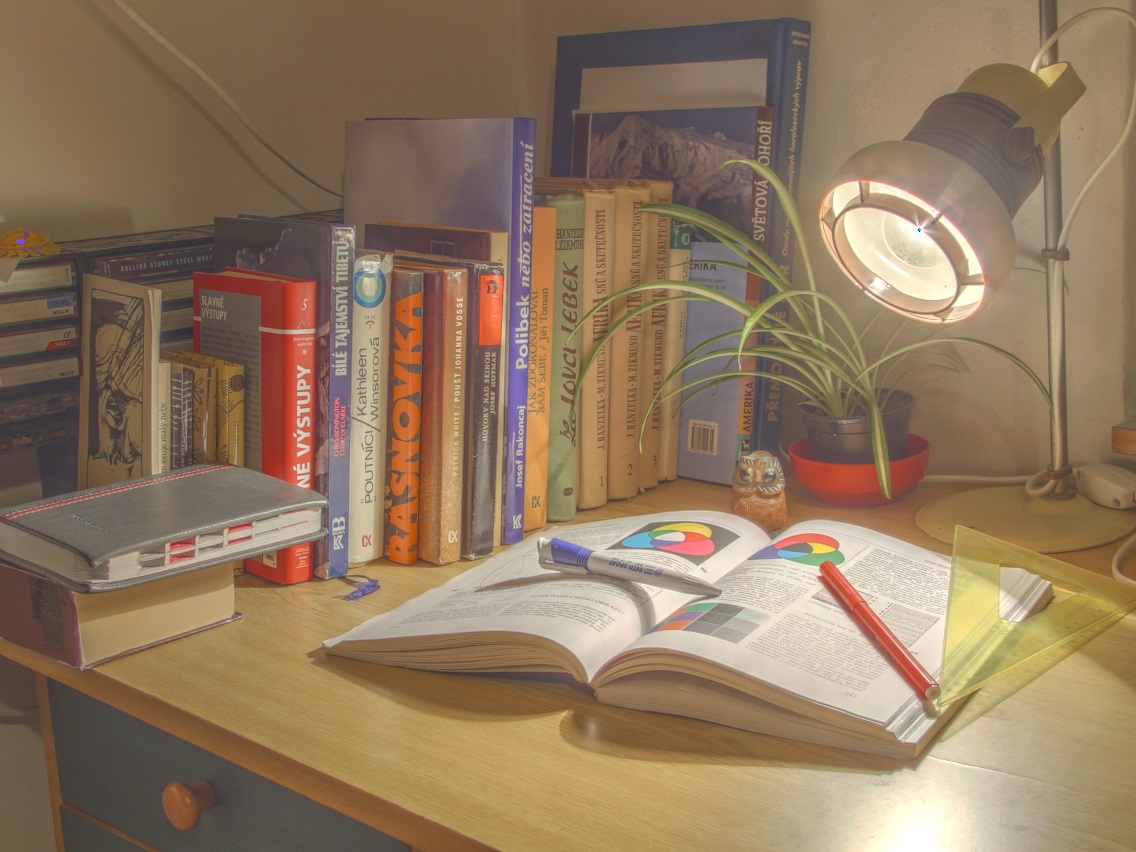}\\
		\vspace{-2mm}
		\caption{LLF ($\sigma_r=0.4$, $\alpha=4$, $\beta=1$)}
	\end{subfigure}
	\begin{subfigure}[c]{0.246\textwidth}
		\centering
		\includegraphics[width=\linewidth]{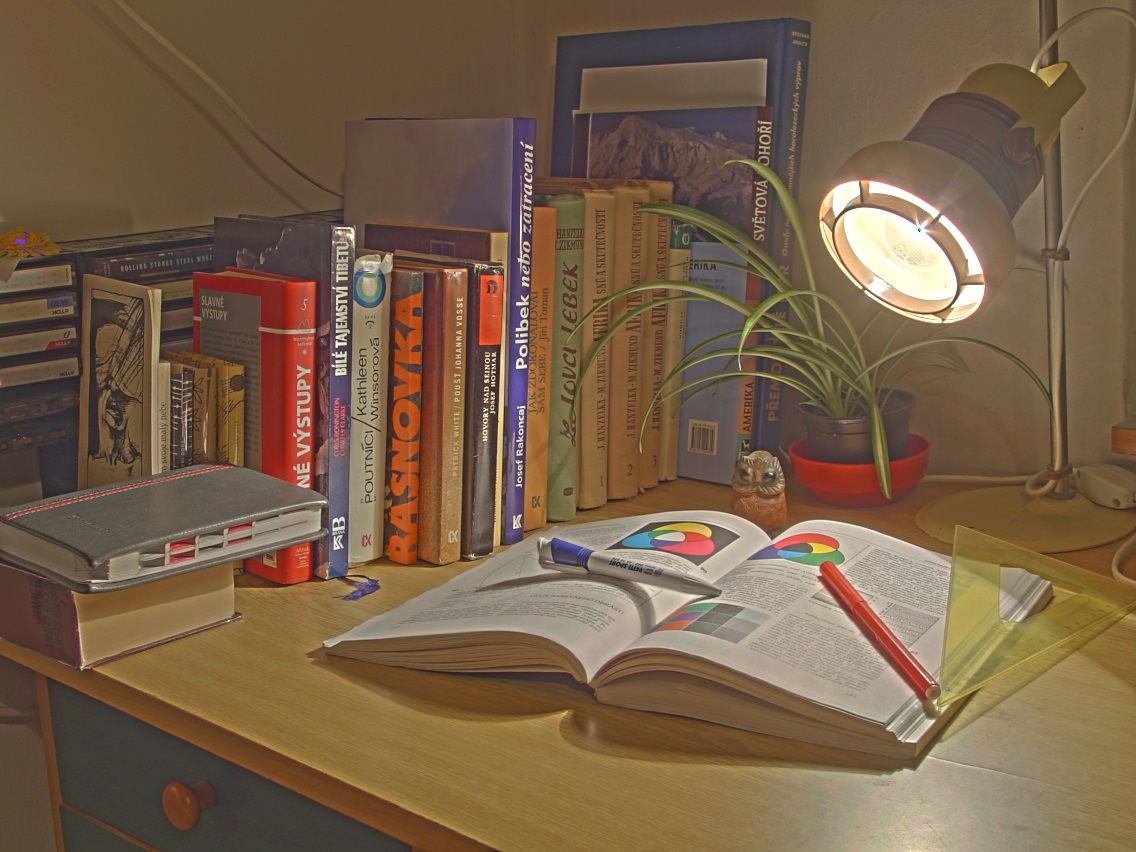}\\
		\vspace{-2mm}
		\caption{Ours ($\sigma_s=3$, $\sigma_r=0.03$)}
	\end{subfigure}
	\vspace{-2mm}
	\caption{HDR Tone mapping result compared with WLS \cite{farbman2008edge} and LLF \cite{paris2011local}. } 
	\label{fig:tone_mapping}
\end{figure*}

\begin{figure}
	\centering
	\begin{subfigure}[c]{0.116\textwidth}
		\centering
		\includegraphics[width=\linewidth]{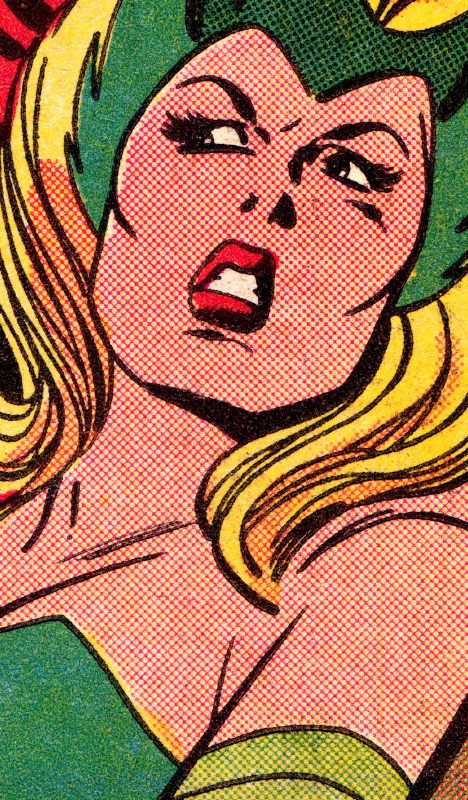}\\ \vspace{2pt}
		\includegraphics[width=\linewidth]{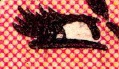}\\
		\vspace{-2mm}
		\caption{}
	\end{subfigure}
	\begin{subfigure}[c]{0.116\textwidth}
		\centering
		\includegraphics[width=\linewidth]{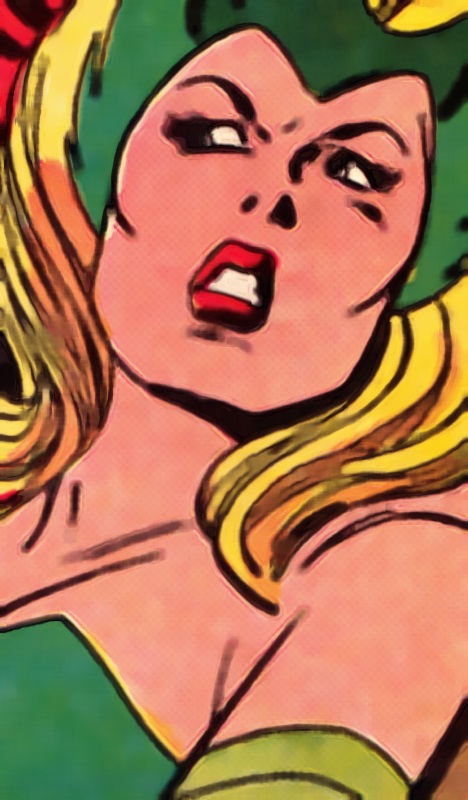}\\  \vspace{2pt}
		\includegraphics[width=\linewidth]{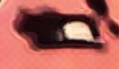}\\
		\vspace{-2mm}
		\caption{}
	\end{subfigure}
	\begin{subfigure}[c]{0.116\textwidth}
		\centering
		\includegraphics[width=\linewidth]{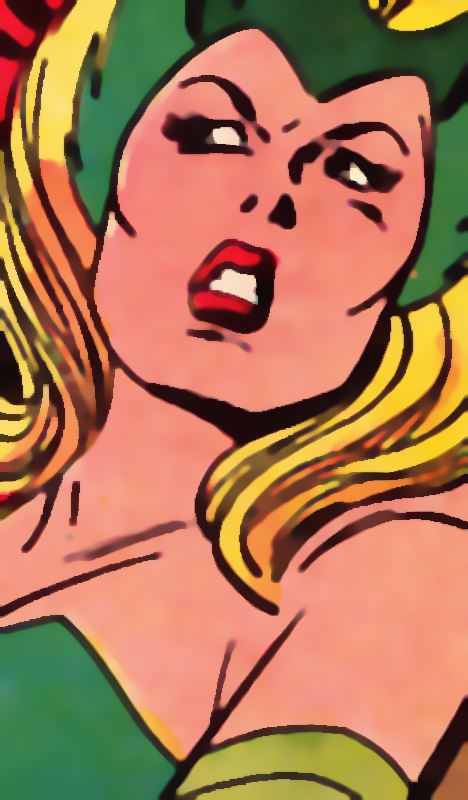}\\ \vspace{2pt}
		\includegraphics[width=\linewidth]{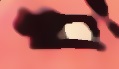} \\
		\vspace{-2mm}
		\caption{}
	\end{subfigure}
	\begin{subfigure}[c]{0.116\textwidth}
		\centering
		\includegraphics[width=\linewidth]{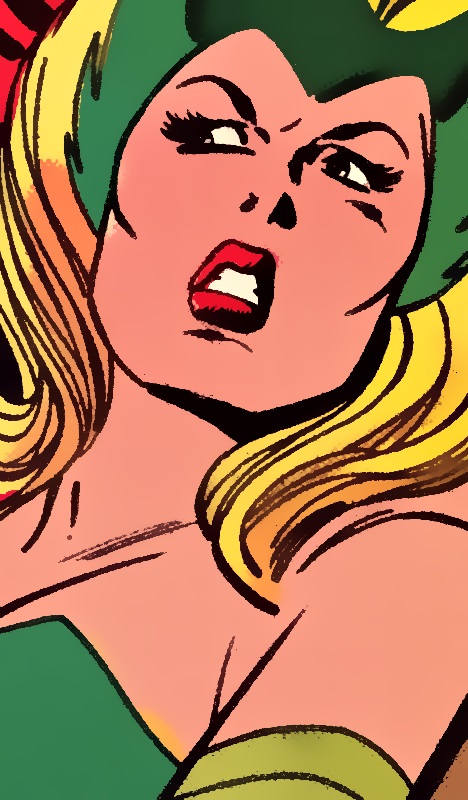}\\ \vspace{2pt}
		\includegraphics[width=\linewidth]{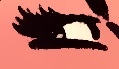} \\
		\vspace{-2mm}
		\caption{}
	\end{subfigure}
	\vspace{-2mm}
	\caption{Inverse halftoning. (a) Input. (b) Result of \cite{karacan2013structure}. (c) Result of \cite{cho2014bilateral} (d) Our result. Parameters: \cite{karacan2013structure} ($k=6$, $\sigma=0.1$, Model 1), \cite{cho2014bilateral} ($k=7$, $n_{itr}=5$), and our method ($\sigma_s=4$, $\sigma_r=0.03$). Source image \copyright Marvel Comics.}  
	\label{fig:halftoning}
\end{figure}

\begin{figure}
	\centering
	\captionsetup[subfigure]{labelformat=empty}
	\begin{subfigure}[c]{0.155\textwidth}
		\centering
		\includegraphics[width=\linewidth]{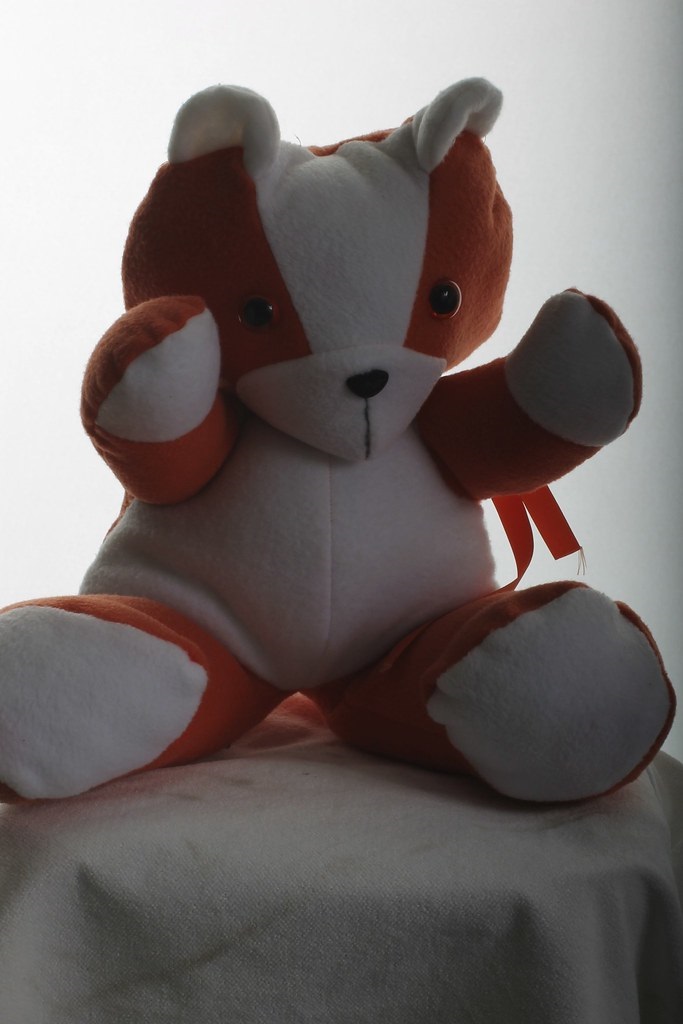} \\ 
		\vspace{-2mm}
		\caption{Input}
	\end{subfigure}
	\begin{subfigure}[c]{0.155\textwidth}
		\centering
		\includegraphics[width=\linewidth]{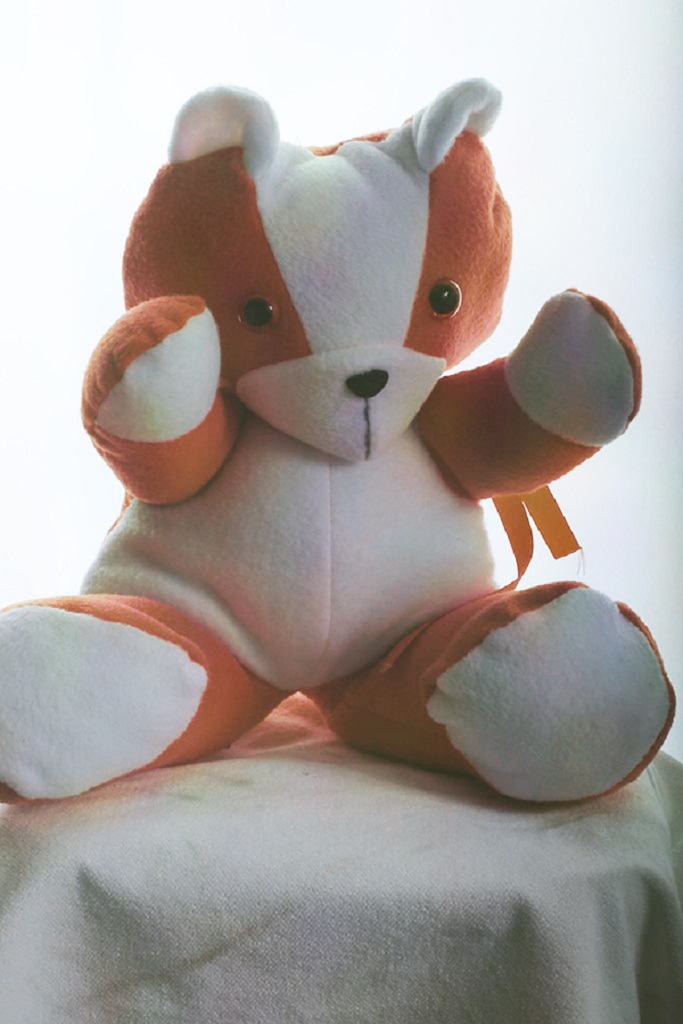} \\ 
		\vspace{-2mm}
		\caption{HDRNet}
	\end{subfigure}
	\begin{subfigure}[c]{0.155\textwidth}
		\centering
		\includegraphics[width=\linewidth]{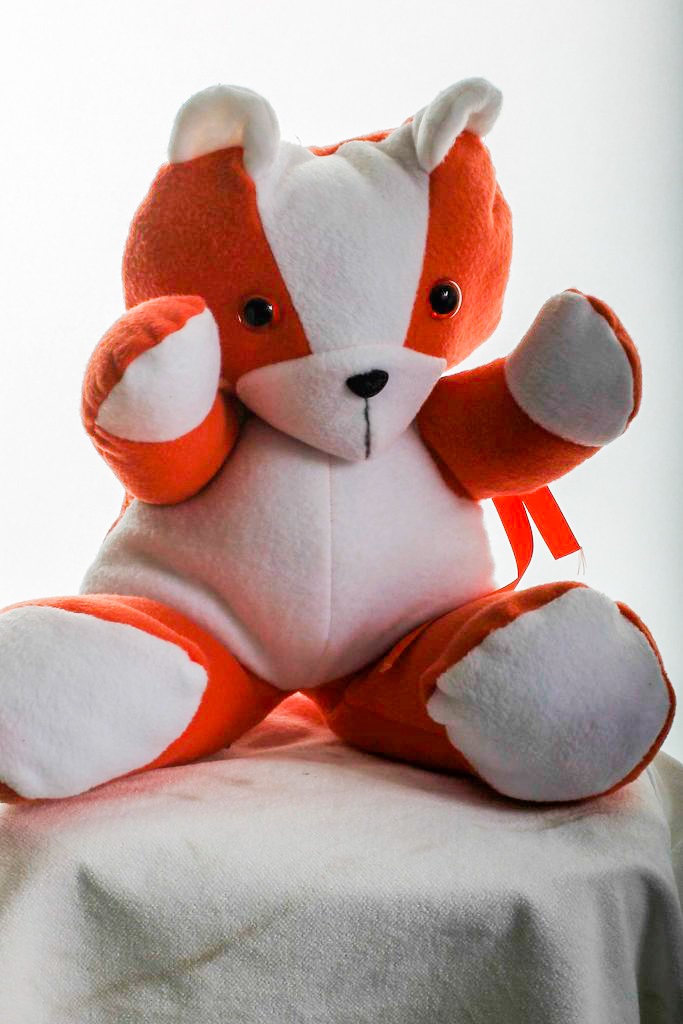} \\
		\vspace{-2mm}
		\caption{Ours ($\sigma_s\!=\!7$, $\sigma_r\!=\!0.07$)}
	\end{subfigure}
	\vspace{-2mm}
	\caption{LDR image enhancement results produced by HDRNet \cite{gharbi2017deep} and our method. Note, result of HDRNet is generated by a pretrained model released by the authors.} 
	\label{fig:low-light-enhancement}
\end{figure}

\paragraph{Effect of noise} As introduced in \cite{zontak2013separating}, the noise level of an image will be gradually reduced as the image scale gets coarser. Our method, which performs texture smoothing by upsampling the coarsest Gaussian pyramid level that is almost noise-free, is thus highly robust to noise, as demonstrated in Figure~\ref{fig:noise}. Please see the supplementary material for more results on noisy images.

\paragraph{Time performance} As the major computation of our algorithm lies in iteratively performing joint bilateral filtering on several low-resolution pyramid levels instead of always operating at original full-resolution, it is relatively efficient compared to existing texture smoothing methods. On a 2.66GHz Intel Core i7 CPU, our unoptimized Matlab implementation that utilizes kernel separation takes about 1 second to process a $1280 \times 720$ image. Thanks to the parallelism of bilateral filtering, our GPU implementation on a NVIDIA GeForce RTX 3090Ti GPU achieves about $200 \times$ speed-up relative to the Matlab implementation, reducing the time cost to 5 milliseconds for the same $1280 \times 720$ image.

\section{Results and Applications}

\paragraph{Comparison with state-of-the-art methods} We compare our method with the state-of-the-art texture smoothing methods \cite{xu2012structure,karacan2013structure,cho2014bilateral,fan2018image} in Figure~\ref{fig:compare_sota}. Note that although \cite{fan2018image} is not specially designed for texture smoothing, it is compared because of its good performance on texture removal. For fair comparison, we produce results of the compared methods using publicly-available implementation or trained model provided by the authors with careful parameter tuning. From the results, we notice two key improvements of our method over the others. First of all, our method can effectively remove complex large-scale and high-contrast textures, such as the blocky texture of varying shapes and sizes in the first image and the dense scatter texture of random colors in the second image. Besides, we are able to avoid blurring or distorting image structures while removing textures. Please see the supplementary material for more comparison results and our comparison with more methods \cite{subr2009edge,zhang2014rolling,zhang2015segment,ham2015robust}.

\paragraph{Applications} Akin to previous work on texture smoothing \cite{xu2012structure,karacan2013structure,cho2014bilateral}, our approach also enables a wide variety of image manipulation applications, including detail enhancement (Figure~\ref{fig:detail_enhancement}), image abstraction (Figure~\ref{fig:abstraction}), HDR tone mapping (Figure~\ref{fig:tone_mapping}), inverse halftoning (Figure~\ref{fig:halftoning}), and LDR image enhancement (Figure~\ref{fig:low-light-enhancement}). As the first four are common applications, here we describe only the implementation of LDR image enhancement. To enhance a poorly lit image $I$, we first follow \cite{guo2016lime} to obtain an initial illumination map $L$ that records the maximum RGB color channel at each pixel. Next, we apply our method to $L$ for producing a smoothed illumination map $L_s$, and then recover an enhanced image $I'$ by $I' = I/L_s^{\gamma}$ based on the Retinex theory, where $\gamma \in (0,1)$ is a enhancement control parameter set as 0.7. For all these applications, our method produces comparable or better results than the compared alternatives, which demonstrates the practicability of our method.

\begin{figure*}
	\centering
	\begin{subfigure}[c]{0.196\textwidth}
		\centering
		\includegraphics[width=\linewidth]{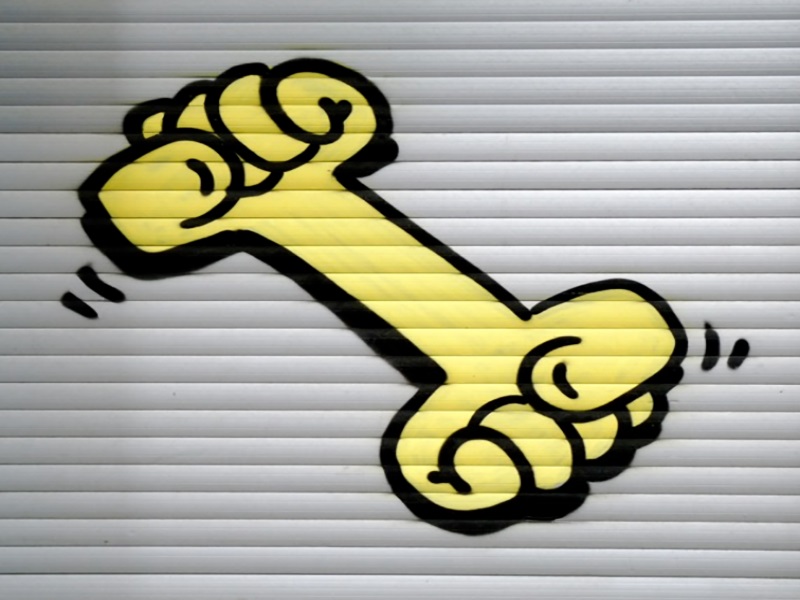} \\  \vspace{2pt}
		\includegraphics[width=\linewidth]{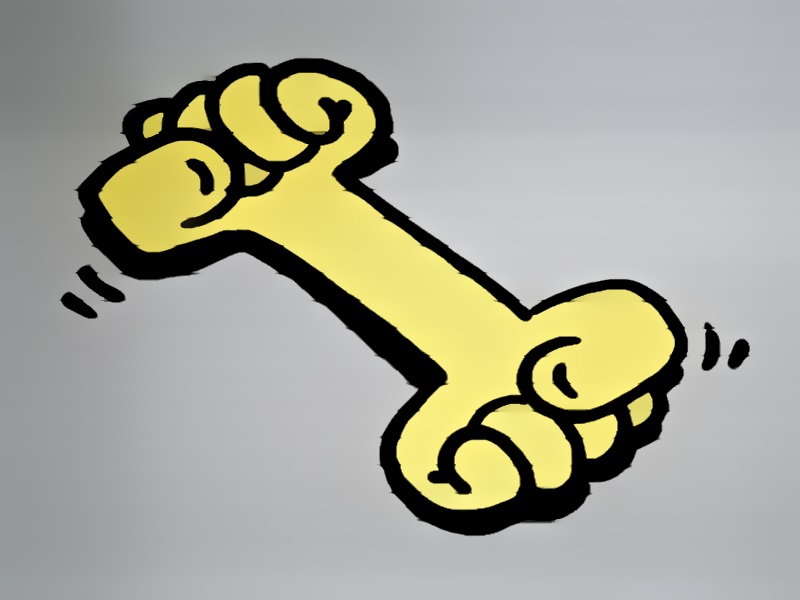} \\
	\end{subfigure}
	\begin{subfigure}[c]{0.196\textwidth}
		\centering
		\includegraphics[width=\linewidth]{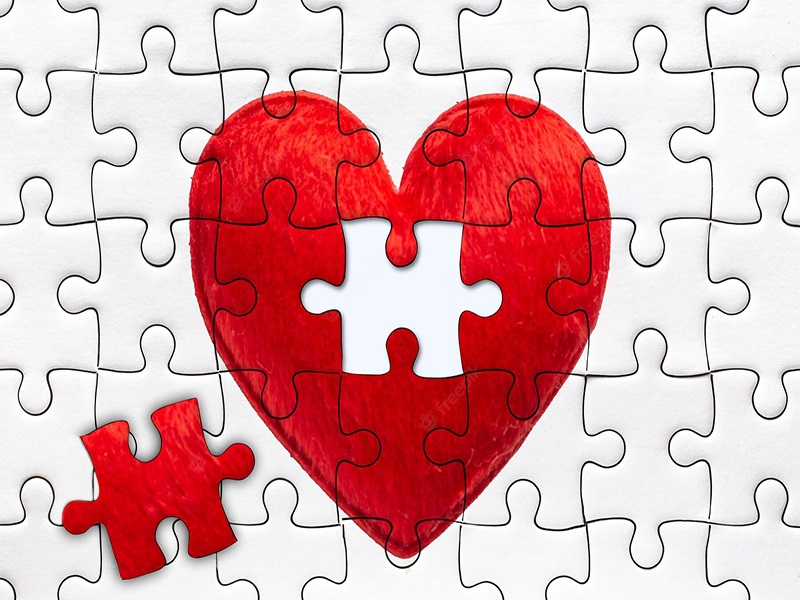} \\  \vspace{2pt}
		\includegraphics[width=\linewidth]{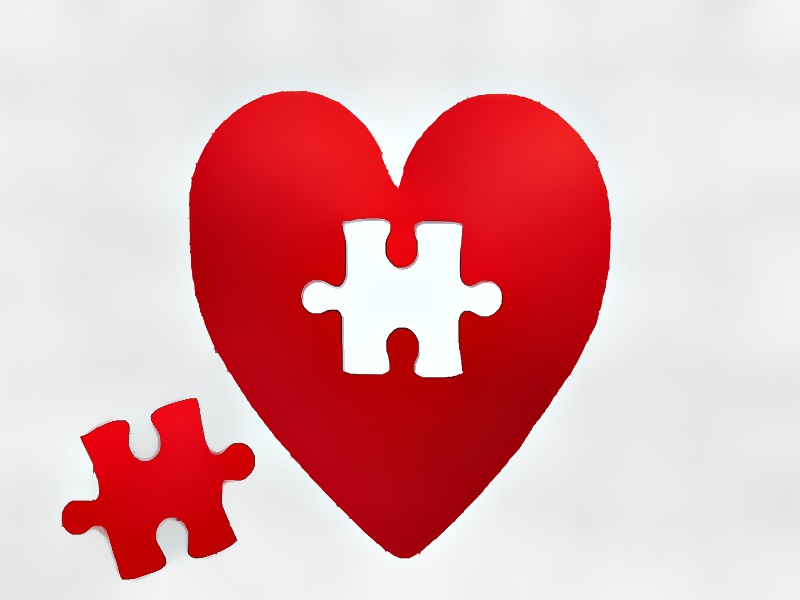} \\
	\end{subfigure}
	\begin{subfigure}[c]{0.196\textwidth}
		\centering
		\includegraphics[width=\linewidth]{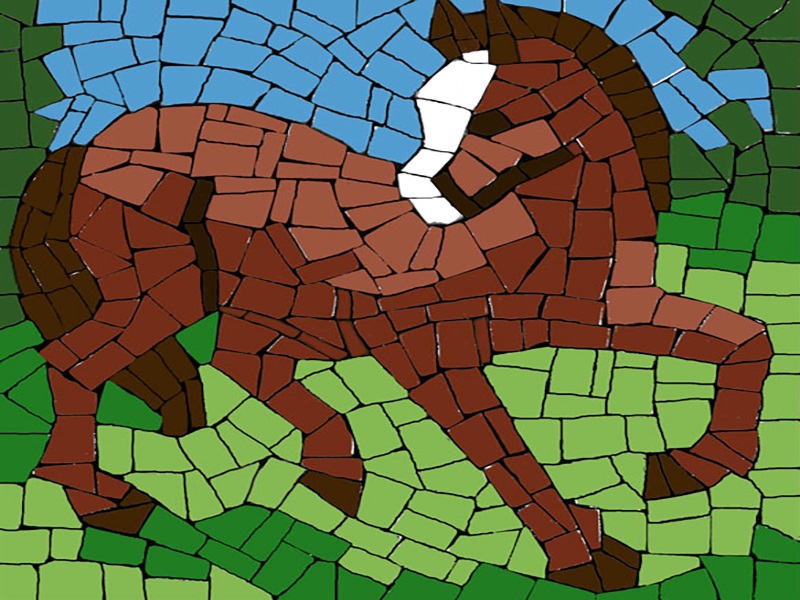} \\  \vspace{2pt}
		\includegraphics[width=\linewidth]{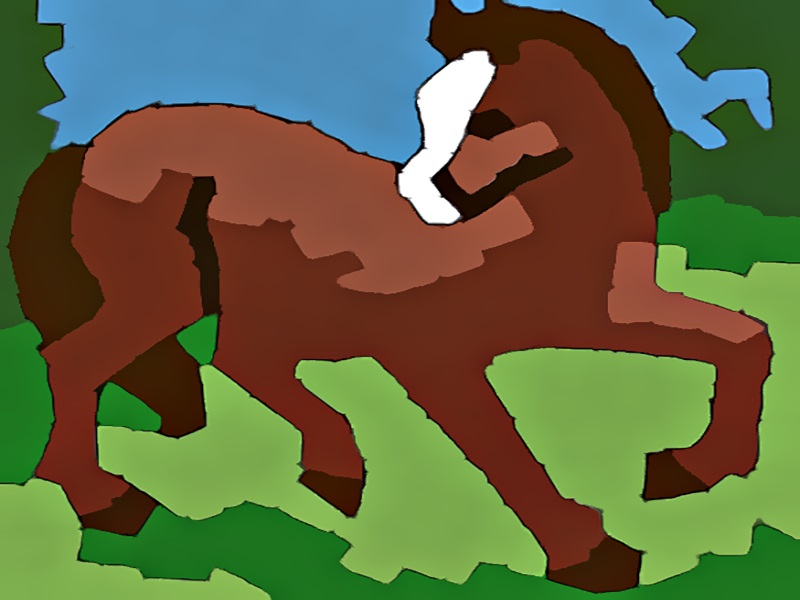} \\
	\end{subfigure}
	\begin{subfigure}[c]{0.196\textwidth}
		\centering
		\includegraphics[width=\linewidth]{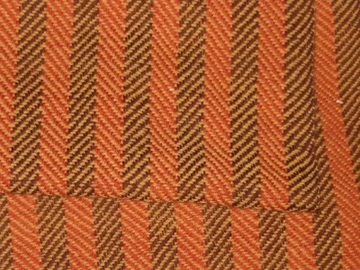} \\  \vspace{2pt}
		\includegraphics[width=\linewidth]{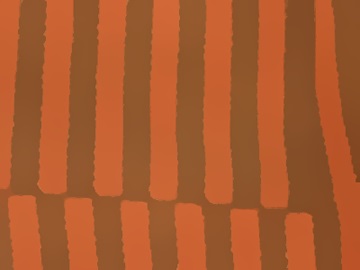} \\
	\end{subfigure}
	\begin{subfigure}[c]{0.196\textwidth}
		\centering
		\includegraphics[width=\linewidth]{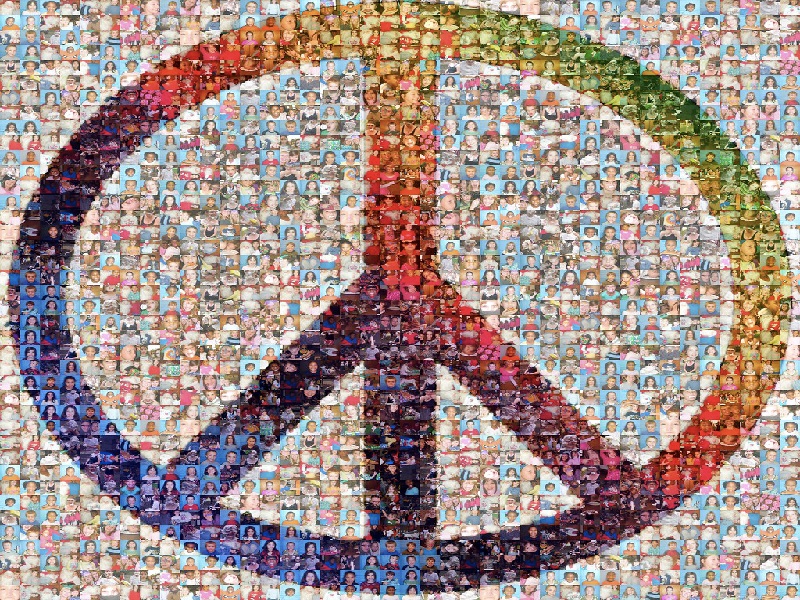} \\  \vspace{2pt}
		\includegraphics[width=\linewidth]{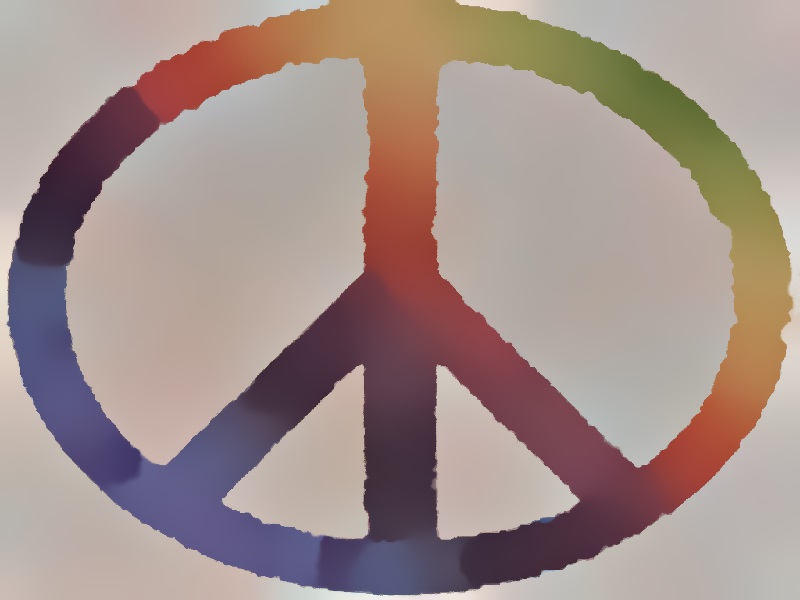} \\
	\end{subfigure}
	\vspace{-2mm}
	\caption{More results produced by our method. From top to bottom are input images and our texture smoothing results. Images courtesy of Flickr user balavenise, Alamy Stock Photo, MosaicArtSupply, \cite{xu2012structure}, and Flickr user Ashley Arend.} 
	\label{fig:more_results}
\end{figure*}

\begin{figure}
	\centering
	\begin{subfigure}[c]{0.236\textwidth}
		\centering
		\begin{tikzpicture}[
		spy using outlines={color=red, rectangle, magnification=3,
			every spy on node/.append style={rectangle,line width=0.15mm},
			every spy in node/.append style={rectangle}}
		]
		\node[inner sep=0,outer sep=0]{\includegraphics[width=\linewidth]{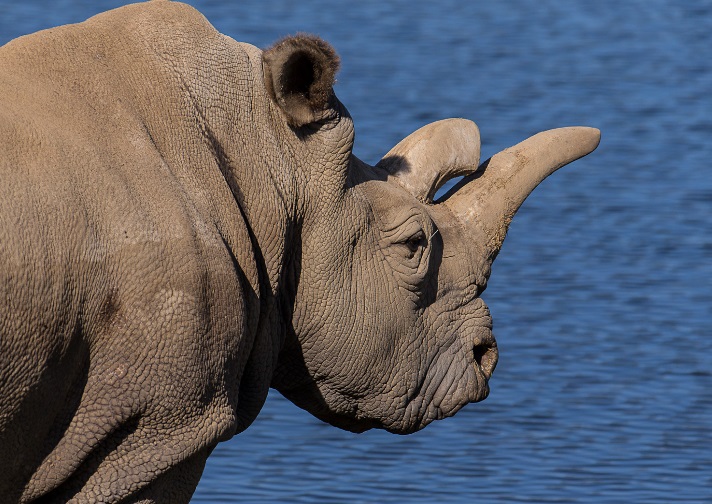}};
		\spy[width=1.0cm, height=1.0cm] on (0.55, 0.4) in node at (1.6, -1.0);
		\end{tikzpicture} \\
	\end{subfigure}
	\begin{subfigure}[c]{0.236\textwidth}
		\centering
		\begin{tikzpicture}[
		spy using outlines={color=red, rectangle, magnification=3,
			every spy on node/.append style={rectangle,line width=0.15mm},
			every spy in node/.append style={rectangle}}
		]
		\node[inner sep=0,outer sep=0]{\includegraphics[width=\linewidth]{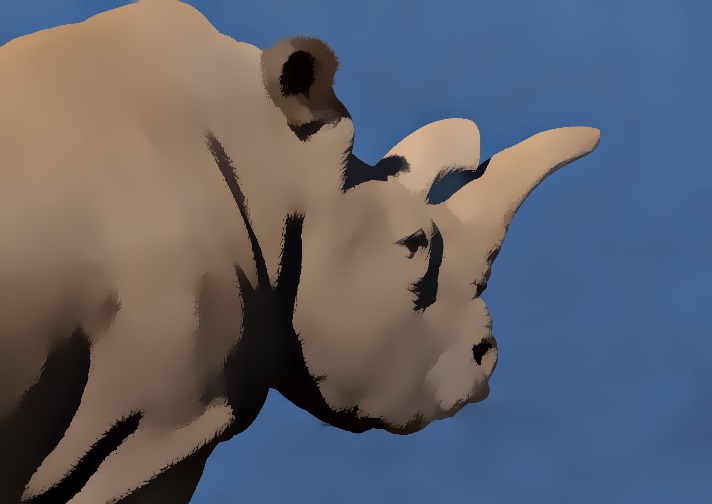}};
		\spy[width=1.0cm, height=1.0cm] on (0.55, 0.4) in node at (1.6, -1.0);
		\end{tikzpicture} \\
	\end{subfigure}
	\vspace{-3mm}
	\caption{Failure case. Our method fails to preserve the small-scale structure surrounded by the rhino horns. Image courtesy of San Diego Zoo.} 
	\label{fig:limitation}
\end{figure}

\paragraph{Additional results.} Figure~\ref{fig:more_results} shows more results produced by our method, where the input images are diverse and contain a broad range of texture types, including: (i) an image with high-contrast line texture (1st column), (ii) an image with large-scale jigsaw texture (2nd column), (iii) an image with irregular tiled texture (3rd column), (iv) an image with complex knitted texture (4th column), and (v) an image with random texture composed of tiny portraits of varying tones and brightness (5th column). As can be seen, our method produces good results for these images, manifesting its effectiveness and robustness in texture smoothing.

\section{Conclusion}
We have presented a new technique for texture smoothing based on standard image pyramids. It is simple, fast and easy to implement, while allowing for surprisingly effective and robust texture removal. In contrast to previous methods, we open up a new perspective to separate texture from structure in the scale space characterized by image pyramids, without relying on any explicit texture-structure separation measures. We believe our work will have a broad impact on image filtering, and will also shed light on the use of pyramids in the domain of image editing and its related applications.

\paragraph{Limitations and future work} Although our method can effectively remove large-scale textures, it may result in loss of small-scale structures that are untraceable in the coarsest Gaussian pyramid level. Figure~\ref{fig:limitation} shows an example where our method fails to retain the small-scale structures surrounded by the rhino horns. Reducing the pyramid depth and the value of $\sigma_s$ can help avoid the issue, but may lead to incomplete texture removal. Note that, for small-scale structures that are not existed in the coarsest level but classified as part of large-scale structures due to image downsampling, our method can recover them with the aid of other levels in Gaussian and Laplacian pyramids, such as the eyelash in Figure~\ref{fig:halftoning} (see the close-ups). Besides, our method inherits the limitation of bilateral filtering and may cause gradient reversal artifacts when a given image is over-smoothed (see Figure~\ref{fig:detail_enhancement}). In the future, we will focus on addressing these limitations. Another promising future work is to extend our method to video texture filtering.

\begin{acks}

We would like to thank the anonymous reviewers for their insightful comments and constructive suggestions. This work is supported by the National Natural Science Foundation of China (U21A20471, 62072191), Guangdong Basic and Applied Basic Research Foundation (2023A1515030002, 2023B1515040025).



\end{acks}

\bibliographystyle{ACM-Reference-Format}
\bibliography{paper}


\begin{thebibliography}{43}


\ifx \showCODEN    \undefined \def \showCODEN     #1{\unskip}     \fi
\ifx \showDOI      \undefined \def \showDOI       #1{#1}\fi
\ifx \showISBNx    \undefined \def \showISBNx     #1{\unskip}     \fi
\ifx \showISBNxiii \undefined \def \showISBNxiii  #1{\unskip}     \fi
\ifx \showISSN     \undefined \def \showISSN      #1{\unskip}     \fi
\ifx \showLCCN     \undefined \def \showLCCN      #1{\unskip}     \fi
\ifx \shownote     \undefined \def \shownote      #1{#1}          \fi
\ifx \showarticletitle \undefined \def \showarticletitle #1{#1}   \fi
\ifx \showURL      \undefined \def \showURL       {\relax}        \fi
\providecommand\bibfield[2]{#2}
\providecommand\bibinfo[2]{#2}
\providecommand\natexlab[1]{#1}
\providecommand\showeprint[2][]{arXiv:#2}

\bibitem[Bao et~al\mbox{.}(2013)]%
        {bao2013tree}
\bibfield{author}{\bibinfo{person}{Linchao Bao}, \bibinfo{person}{Yibing Song},
  \bibinfo{person}{Qingxiong Yang}, \bibinfo{person}{Hao Yuan}, {and}
  \bibinfo{person}{Gang Wang}.} \bibinfo{year}{2013}\natexlab{}.
\newblock \showarticletitle{Tree filtering: Efficient structure-preserving
  smoothing with a minimum spanning tree}.
\newblock \bibinfo{journal}{\emph{IEEE Transactions on Image Processing}}
  \bibinfo{volume}{23}, \bibinfo{number}{2} (\bibinfo{year}{2013}),
  \bibinfo{pages}{555--569}.
\newblock


\bibitem[Bi et~al\mbox{.}(2015)]%
        {bi20151}
\bibfield{author}{\bibinfo{person}{Sai Bi}, \bibinfo{person}{Xiaoguang Han},
  {and} \bibinfo{person}{Yizhou Yu}.} \bibinfo{year}{2015}\natexlab{}.
\newblock \showarticletitle{An L1 image transform for edge-preserving smoothing
  and scene-level intrinsic decomposition}.
\newblock \bibinfo{journal}{\emph{ACM Transactions on Graphics}}
  \bibinfo{volume}{34}, \bibinfo{number}{4} (\bibinfo{year}{2015}),
  \bibinfo{pages}{1--12}.
\newblock


\bibitem[Burt and Adelson(1983)]%
        {burt1983laplacian}
\bibfield{author}{\bibinfo{person}{Peter~J Burt} {and}
  \bibinfo{person}{Edward~H Adelson}.} \bibinfo{year}{1983}\natexlab{}.
\newblock \showarticletitle{The Laplacian Pyramid as a Compact Image Code}.
\newblock \bibinfo{journal}{\emph{IEEE Transactions on Communications}}
  \bibinfo{volume}{3}, \bibinfo{number}{4} (\bibinfo{year}{1983}).
\newblock


\bibitem[Chen et~al\mbox{.}(2007)]%
        {chen2007real}
\bibfield{author}{\bibinfo{person}{Jiawen Chen}, \bibinfo{person}{Sylvain
  Paris}, {and} \bibinfo{person}{Fr{\'e}do Durand}.}
  \bibinfo{year}{2007}\natexlab{}.
\newblock \showarticletitle{Real-time edge-aware image processing with the
  bilateral grid}.
\newblock \bibinfo{journal}{\emph{ACM Transactions on Graphics}}
  \bibinfo{volume}{26}, \bibinfo{number}{3} (\bibinfo{year}{2007}),
  \bibinfo{pages}{103}.
\newblock


\bibitem[Cho et~al\mbox{.}(2014)]%
        {cho2014bilateral}
\bibfield{author}{\bibinfo{person}{Hojin Cho}, \bibinfo{person}{Hyunjoon Lee},
  \bibinfo{person}{Henry Kang}, {and} \bibinfo{person}{Seungyong Lee}.}
  \bibinfo{year}{2014}\natexlab{}.
\newblock \showarticletitle{Bilateral texture filtering}.
\newblock \bibinfo{journal}{\emph{ACM Transactions on Graphics}}
  \bibinfo{volume}{33}, \bibinfo{number}{4} (\bibinfo{year}{2014}),
  \bibinfo{pages}{1--8}.
\newblock


\bibitem[Criminisi et~al\mbox{.}(2010)]%
        {criminisi2010geodesic}
\bibfield{author}{\bibinfo{person}{Antonio Criminisi}, \bibinfo{person}{Toby
  Sharp}, \bibinfo{person}{Carsten Rother}, {and} \bibinfo{person}{Patrick
  P{\'e}rez}.} \bibinfo{year}{2010}\natexlab{}.
\newblock \showarticletitle{Geodesic image and video editing.}
\newblock \bibinfo{journal}{\emph{ACM Transactions on Graphics}}
  \bibinfo{volume}{29}, \bibinfo{number}{5} (\bibinfo{year}{2010}),
  \bibinfo{pages}{134--1}.
\newblock


\bibitem[Du et~al\mbox{.}(2016)]%
        {du2016two}
\bibfield{author}{\bibinfo{person}{Hui Du}, \bibinfo{person}{Xiaogang Jin},
  {and} \bibinfo{person}{Philip~J Willis}.} \bibinfo{year}{2016}\natexlab{}.
\newblock \showarticletitle{Two-level joint local laplacian texture filtering}.
\newblock \bibinfo{journal}{\emph{The Visual Computer}}  \bibinfo{volume}{32}
  (\bibinfo{year}{2016}), \bibinfo{pages}{1537--1548}.
\newblock


\bibitem[Durand and Dorsey(2002)]%
        {durand2002fast}
\bibfield{author}{\bibinfo{person}{Fr{\'e}do Durand} {and}
  \bibinfo{person}{Julie Dorsey}.} \bibinfo{year}{2002}\natexlab{}.
\newblock \showarticletitle{Fast bilateral filtering for the display of
  high-dynamic-range images}.
\newblock \bibinfo{journal}{\emph{ACM Transactions on Graphics}}
  \bibinfo{volume}{21}, \bibinfo{number}{3} (\bibinfo{year}{2002}),
  \bibinfo{pages}{257--266}.
\newblock


\bibitem[Eisemann and Durand(2004)]%
        {eisemann2004flash}
\bibfield{author}{\bibinfo{person}{Elmar Eisemann} {and}
  \bibinfo{person}{Fr{\'e}do Durand}.} \bibinfo{year}{2004}\natexlab{}.
\newblock \showarticletitle{Flash photography enhancement via intrinsic
  relighting}.
\newblock \bibinfo{journal}{\emph{ACM Transactions on Graphics}}
  \bibinfo{volume}{23}, \bibinfo{number}{3} (\bibinfo{year}{2004}),
  \bibinfo{pages}{673--678}.
\newblock


\bibitem[Fan et~al\mbox{.}(2018)]%
        {fan2018image}
\bibfield{author}{\bibinfo{person}{Qingnan Fan}, \bibinfo{person}{Jiaolong
  Yang}, \bibinfo{person}{David Wipf}, \bibinfo{person}{Baoquan Chen}, {and}
  \bibinfo{person}{Xin Tong}.} \bibinfo{year}{2018}\natexlab{}.
\newblock \showarticletitle{Image smoothing via unsupervised learning}.
\newblock \bibinfo{journal}{\emph{ACM Transactions on Graphics}}
  \bibinfo{volume}{37}, \bibinfo{number}{6} (\bibinfo{year}{2018}),
  \bibinfo{pages}{1--14}.
\newblock


\bibitem[Farbman et~al\mbox{.}(2008)]%
        {farbman2008edge}
\bibfield{author}{\bibinfo{person}{Zeev Farbman}, \bibinfo{person}{Raanan
  Fattal}, \bibinfo{person}{Dani Lischinski}, {and} \bibinfo{person}{Richard
  Szeliski}.} \bibinfo{year}{2008}\natexlab{}.
\newblock \showarticletitle{Edge-preserving decompositions for multi-scale tone
  and detail manipulation}.
\newblock \bibinfo{journal}{\emph{ACM Transactions on Graphics}}
  \bibinfo{volume}{27}, \bibinfo{number}{3} (\bibinfo{year}{2008}),
  \bibinfo{pages}{1--10}.
\newblock


\bibitem[Fattal(2009)]%
        {fattal2009edge}
\bibfield{author}{\bibinfo{person}{Raanan Fattal}.}
  \bibinfo{year}{2009}\natexlab{}.
\newblock \showarticletitle{Edge-avoiding wavelets and their applications}.
\newblock \bibinfo{journal}{\emph{ACM Transactions on Graphics}}
  \bibinfo{volume}{28}, \bibinfo{number}{3} (\bibinfo{year}{2009}),
  \bibinfo{pages}{1--10}.
\newblock


\bibitem[Gastal and Oliveira(2011)]%
        {gastal2011domain}
\bibfield{author}{\bibinfo{person}{Eduardo~SL Gastal} {and}
  \bibinfo{person}{Manuel~M Oliveira}.} \bibinfo{year}{2011}\natexlab{}.
\newblock \showarticletitle{Domain transform for edge-aware image and video
  processing}.
\newblock \bibinfo{journal}{\emph{ACM Transactions on Graphics}}
  \bibinfo{volume}{30}, \bibinfo{number}{4} (\bibinfo{year}{2011}),
  \bibinfo{pages}{1--12}.
\newblock


\bibitem[Gastal and Oliveira(2012)]%
        {gastal2012adaptive}
\bibfield{author}{\bibinfo{person}{Eduardo~SL Gastal} {and}
  \bibinfo{person}{Manuel~M Oliveira}.} \bibinfo{year}{2012}\natexlab{}.
\newblock \showarticletitle{Adaptive manifolds for real-time high-dimensional
  filtering}.
\newblock \bibinfo{journal}{\emph{ACM Transactions on Graphics}}
  \bibinfo{volume}{31}, \bibinfo{number}{4} (\bibinfo{year}{2012}),
  \bibinfo{pages}{1--13}.
\newblock


\bibitem[Gharbi et~al\mbox{.}(2017)]%
        {gharbi2017deep}
\bibfield{author}{\bibinfo{person}{Micha{\"e}l Gharbi}, \bibinfo{person}{Jiawen
  Chen}, \bibinfo{person}{Jonathan~T Barron}, \bibinfo{person}{Samuel~W
  Hasinoff}, {and} \bibinfo{person}{Fr{\'e}do Durand}.}
  \bibinfo{year}{2017}\natexlab{}.
\newblock \showarticletitle{Deep bilateral learning for real-time image
  enhancement}.
\newblock \bibinfo{journal}{\emph{ACM Transactions on Graphics}}
  \bibinfo{volume}{36}, \bibinfo{number}{4} (\bibinfo{year}{2017}),
  \bibinfo{pages}{1--12}.
\newblock


\bibitem[Guo et~al\mbox{.}(2016)]%
        {guo2016lime}
\bibfield{author}{\bibinfo{person}{Xiaojie Guo}, \bibinfo{person}{Yu Li}, {and}
  \bibinfo{person}{Haibin Ling}.} \bibinfo{year}{2016}\natexlab{}.
\newblock \showarticletitle{LIME: Low-light image enhancement via illumination
  map estimation}.
\newblock \bibinfo{journal}{\emph{IEEE Transactions on Image Processing}}
  \bibinfo{volume}{26}, \bibinfo{number}{2} (\bibinfo{year}{2016}),
  \bibinfo{pages}{982--993}.
\newblock


\bibitem[Ham et~al\mbox{.}(2015)]%
        {ham2015robust}
\bibfield{author}{\bibinfo{person}{Bumsub Ham}, \bibinfo{person}{Minsu Cho},
  {and} \bibinfo{person}{Jean Ponce}.} \bibinfo{year}{2015}\natexlab{}.
\newblock \showarticletitle{Robust image filtering using joint static and
  dynamic guidance}. In \bibinfo{booktitle}{\emph{CVPR}}.
  \bibinfo{pages}{4823--4831}.
\newblock


\bibitem[Jeon et~al\mbox{.}(2016)]%
        {jeon2016scale}
\bibfield{author}{\bibinfo{person}{Junho Jeon}, \bibinfo{person}{Hyunjoon Lee},
  \bibinfo{person}{Henry Kang}, {and} \bibinfo{person}{Seungyong Lee}.}
  \bibinfo{year}{2016}\natexlab{}.
\newblock \showarticletitle{Scale-aware structure-preserving texture
  filtering}.
\newblock \bibinfo{journal}{\emph{Computer Graphics Forum}}
  \bibinfo{volume}{35}, \bibinfo{number}{7} (\bibinfo{year}{2016}),
  \bibinfo{pages}{77--86}.
\newblock


\bibitem[Karacan et~al\mbox{.}(2013)]%
        {karacan2013structure}
\bibfield{author}{\bibinfo{person}{Levent Karacan}, \bibinfo{person}{Erkut
  Erdem}, {and} \bibinfo{person}{Aykut Erdem}.}
  \bibinfo{year}{2013}\natexlab{}.
\newblock \showarticletitle{Structure-preserving image smoothing via region
  covariances}.
\newblock \bibinfo{journal}{\emph{ACM Transactions on Graphics}}
  \bibinfo{volume}{32}, \bibinfo{number}{6} (\bibinfo{year}{2013}),
  \bibinfo{pages}{1--11}.
\newblock


\bibitem[Kass and Solomon(2010)]%
        {kass2010smoothed}
\bibfield{author}{\bibinfo{person}{Michael Kass} {and} \bibinfo{person}{Justin
  Solomon}.} \bibinfo{year}{2010}\natexlab{}.
\newblock \showarticletitle{Smoothed local histogram filters}.
\newblock \bibinfo{journal}{\emph{ACM Transactions on Graphics}}
  \bibinfo{volume}{29}, \bibinfo{number}{4} (\bibinfo{year}{2010}),
  \bibinfo{pages}{1--10}.
\newblock


\bibitem[Kim et~al\mbox{.}(2018)]%
        {kim2018structure}
\bibfield{author}{\bibinfo{person}{Youngjung Kim}, \bibinfo{person}{Bumsub
  Ham}, \bibinfo{person}{Minh~N Do}, {and} \bibinfo{person}{Kwanghoon Sohn}.}
  \bibinfo{year}{2018}\natexlab{}.
\newblock \showarticletitle{Structure-texture image decomposition using deep
  variational priors}.
\newblock \bibinfo{journal}{\emph{IEEE Transactions on Image Processing}}
  \bibinfo{volume}{28}, \bibinfo{number}{6} (\bibinfo{year}{2018}),
  \bibinfo{pages}{2692--2704}.
\newblock


\bibitem[Kopf et~al\mbox{.}(2007)]%
        {kopf2007joint}
\bibfield{author}{\bibinfo{person}{Johannes Kopf}, \bibinfo{person}{Michael~F
  Cohen}, \bibinfo{person}{Dani Lischinski}, {and} \bibinfo{person}{Matt
  Uyttendaele}.} \bibinfo{year}{2007}\natexlab{}.
\newblock \showarticletitle{Joint bilateral upsampling}.
\newblock \bibinfo{journal}{\emph{ACM Transactions on Graphics}}
  \bibinfo{volume}{26}, \bibinfo{number}{3} (\bibinfo{year}{2007}),
  \bibinfo{pages}{96}.
\newblock


\bibitem[Liu et~al\mbox{.}(2016)]%
        {liu2016learning}
\bibfield{author}{\bibinfo{person}{Sifei Liu}, \bibinfo{person}{Jinshan Pan},
  {and} \bibinfo{person}{Ming-Hsuan Yang}.} \bibinfo{year}{2016}\natexlab{}.
\newblock \showarticletitle{Learning recursive filters for low-level vision via
  a hybrid neural network}. In \bibinfo{booktitle}{\emph{ECCV}}.
  \bibinfo{pages}{560--576}.
\newblock


\bibitem[Liu et~al\mbox{.}(2017)]%
        {liu2017semi}
\bibfield{author}{\bibinfo{person}{Wei Liu}, \bibinfo{person}{Xiaogang Chen},
  \bibinfo{person}{Chuanhua Shen}, \bibinfo{person}{Zhi Liu}, {and}
  \bibinfo{person}{Jie Yang}.} \bibinfo{year}{2017}\natexlab{}.
\newblock \showarticletitle{Semi-global weighted least squares in image
  filtering}. In \bibinfo{booktitle}{\emph{ICCV}}. \bibinfo{pages}{5861--5869}.
\newblock


\bibitem[Liu et~al\mbox{.}(2020)]%
        {liu2020real}
\bibfield{author}{\bibinfo{person}{Wei Liu}, \bibinfo{person}{Pingping Zhang},
  \bibinfo{person}{Xiaolin Huang}, \bibinfo{person}{Jie Yang},
  \bibinfo{person}{Chunhua Shen}, {and} \bibinfo{person}{Ian Reid}.}
  \bibinfo{year}{2020}\natexlab{}.
\newblock \showarticletitle{Real-time image smoothing via iterative least
  squares}.
\newblock \bibinfo{journal}{\emph{ACM Transactions on Graphics}}
  \bibinfo{volume}{39}, \bibinfo{number}{3} (\bibinfo{year}{2020}),
  \bibinfo{pages}{1--24}.
\newblock


\bibitem[Liu et~al\mbox{.}(2021)]%
        {liu2021generalized}
\bibfield{author}{\bibinfo{person}{Wei Liu}, \bibinfo{person}{Pingping Zhang},
  \bibinfo{person}{Yinjie Lei}, \bibinfo{person}{Xiaolin Huang},
  \bibinfo{person}{Jie Yang}, {and} \bibinfo{person}{Michael Kwok-Po Ng}.}
  \bibinfo{year}{2021}\natexlab{}.
\newblock \showarticletitle{A generalized framework for edge-preserving and
  structure-preserving image smoothing}.
\newblock \bibinfo{journal}{\emph{IEEE Transactions on Pattern Analysis and
  Machine Intelligence}} (\bibinfo{year}{2021}).
\newblock


\bibitem[Lu et~al\mbox{.}(2018)]%
        {lu2018deep}
\bibfield{author}{\bibinfo{person}{Kaiyue Lu}, \bibinfo{person}{Shaodi You},
  {and} \bibinfo{person}{Nick Barnes}.} \bibinfo{year}{2018}\natexlab{}.
\newblock \showarticletitle{Deep texture and structure aware filtering network
  for image smoothing}. In \bibinfo{booktitle}{\emph{ECCV}}.
  \bibinfo{pages}{217--233}.
\newblock


\bibitem[Min et~al\mbox{.}(2014)]%
        {min2014fast}
\bibfield{author}{\bibinfo{person}{Dongbo Min}, \bibinfo{person}{Sunghwan
  Choi}, \bibinfo{person}{Jiangbo Lu}, \bibinfo{person}{Bumsub Ham},
  \bibinfo{person}{Kwanghoon Sohn}, {and} \bibinfo{person}{Minh~N Do}.}
  \bibinfo{year}{2014}\natexlab{}.
\newblock \showarticletitle{Fast global image smoothing based on weighted least
  squares}.
\newblock \bibinfo{journal}{\emph{IEEE Transactions on Image Processing}}
  \bibinfo{volume}{23}, \bibinfo{number}{12} (\bibinfo{year}{2014}),
  \bibinfo{pages}{5638--5653}.
\newblock


\bibitem[Paris and Durand(2006)]%
        {paris2006fast}
\bibfield{author}{\bibinfo{person}{Sylvain Paris} {and}
  \bibinfo{person}{Fr{\'e}do Durand}.} \bibinfo{year}{2006}\natexlab{}.
\newblock \showarticletitle{A fast approximation of the bilateral filter using
  a signal processing approach}. In \bibinfo{booktitle}{\emph{ECCV}}.
  \bibinfo{pages}{568--580}.
\newblock


\bibitem[Paris et~al\mbox{.}(2011)]%
        {paris2011local}
\bibfield{author}{\bibinfo{person}{Sylvain Paris}, \bibinfo{person}{Samuel~W
  Hasinoff}, {and} \bibinfo{person}{Jan Kautz}.}
  \bibinfo{year}{2011}\natexlab{}.
\newblock \showarticletitle{Local laplacian filters: edge-aware image
  processing with a laplacian pyramid.}
\newblock \bibinfo{journal}{\emph{ACM Transactions on Graphics}}
  \bibinfo{volume}{30}, \bibinfo{number}{4} (\bibinfo{year}{2011}),
  \bibinfo{pages}{68}.
\newblock


\bibitem[Perona and Malik(1990)]%
        {perona1990scale}
\bibfield{author}{\bibinfo{person}{Pietro Perona} {and}
  \bibinfo{person}{Jitendra Malik}.} \bibinfo{year}{1990}\natexlab{}.
\newblock \showarticletitle{Scale-space and edge detection using anisotropic
  diffusion}.
\newblock \bibinfo{journal}{\emph{IEEE Transactions on Pattern Analysis and
  Machine Intelligence}} \bibinfo{volume}{12}, \bibinfo{number}{7}
  (\bibinfo{year}{1990}), \bibinfo{pages}{629--639}.
\newblock


\bibitem[Petschnigg et~al\mbox{.}(2004)]%
        {petschnigg2004digital}
\bibfield{author}{\bibinfo{person}{Georg Petschnigg}, \bibinfo{person}{Richard
  Szeliski}, \bibinfo{person}{Maneesh Agrawala}, \bibinfo{person}{Michael
  Cohen}, \bibinfo{person}{Hugues Hoppe}, {and} \bibinfo{person}{Kentaro
  Toyama}.} \bibinfo{year}{2004}\natexlab{}.
\newblock \showarticletitle{Digital photography with flash and no-flash image
  pairs}.
\newblock \bibinfo{journal}{\emph{ACM Transactions on Graphics}}
  \bibinfo{volume}{23}, \bibinfo{number}{3} (\bibinfo{year}{2004}),
  \bibinfo{pages}{664--672}.
\newblock


\bibitem[Subr et~al\mbox{.}(2009)]%
        {subr2009edge}
\bibfield{author}{\bibinfo{person}{Kartic Subr}, \bibinfo{person}{Cyril Soler},
  {and} \bibinfo{person}{Fr{\'e}do Durand}.} \bibinfo{year}{2009}\natexlab{}.
\newblock \showarticletitle{Edge-preserving multiscale image decomposition
  based on local extrema}.
\newblock \bibinfo{journal}{\emph{ACM Transactions on Graphics}}
  \bibinfo{volume}{28}, \bibinfo{number}{5} (\bibinfo{year}{2009}),
  \bibinfo{pages}{1--9}.
\newblock


\bibitem[Tomasi and Manduchi(1998)]%
        {tomasi1998bilateral}
\bibfield{author}{\bibinfo{person}{Carlo Tomasi} {and} \bibinfo{person}{Roberto
  Manduchi}.} \bibinfo{year}{1998}\natexlab{}.
\newblock \showarticletitle{Bilateral Filtering for Gray and Color Images}. In
  \bibinfo{booktitle}{\emph{ICCV}}. \bibinfo{pages}{839--839}.
\newblock


\bibitem[Wei et~al\mbox{.}(2018)]%
        {wei2018joint}
\bibfield{author}{\bibinfo{person}{Xing Wei}, \bibinfo{person}{Qingxiong Yang},
  {and} \bibinfo{person}{Yihong Gong}.} \bibinfo{year}{2018}\natexlab{}.
\newblock \showarticletitle{Joint contour filtering}.
\newblock \bibinfo{journal}{\emph{International Journal of Computer Vision}}
  \bibinfo{volume}{126}, \bibinfo{number}{11} (\bibinfo{year}{2018}),
  \bibinfo{pages}{1245--1265}.
\newblock


\bibitem[Weiss(2006)]%
        {weiss2006fast}
\bibfield{author}{\bibinfo{person}{Ben Weiss}.}
  \bibinfo{year}{2006}\natexlab{}.
\newblock \showarticletitle{Fast median and bilateral filtering}.
\newblock \bibinfo{journal}{\emph{ACM Transactions on Graphics}}
  \bibinfo{volume}{25}, \bibinfo{number}{3} (\bibinfo{year}{2006}),
  \bibinfo{pages}{519--526}.
\newblock


\bibitem[Xu et~al\mbox{.}(2011)]%
        {xu2011image}
\bibfield{author}{\bibinfo{person}{Li Xu}, \bibinfo{person}{Cewu Lu},
  \bibinfo{person}{Yi Xu}, {and} \bibinfo{person}{Jiaya Jia}.}
  \bibinfo{year}{2011}\natexlab{}.
\newblock \showarticletitle{Image smoothing via {L0} gradient minimization}.
\newblock \bibinfo{journal}{\emph{ACM Transactions on Graphics}}
  \bibinfo{volume}{30}, \bibinfo{number}{6} (\bibinfo{year}{2011}),
  \bibinfo{pages}{1--12}.
\newblock


\bibitem[Xu et~al\mbox{.}(2015)]%
        {xu2015deep}
\bibfield{author}{\bibinfo{person}{Li Xu}, \bibinfo{person}{Jimmy Ren},
  \bibinfo{person}{Qiong Yan}, \bibinfo{person}{Renjie Liao}, {and}
  \bibinfo{person}{Jiaya Jia}.} \bibinfo{year}{2015}\natexlab{}.
\newblock \showarticletitle{Deep edge-aware filters}. In
  \bibinfo{booktitle}{\emph{ICML}}. \bibinfo{pages}{1669--1678}.
\newblock


\bibitem[Xu et~al\mbox{.}(2012)]%
        {xu2012structure}
\bibfield{author}{\bibinfo{person}{Li Xu}, \bibinfo{person}{Qiong Yan},
  \bibinfo{person}{Yang Xia}, {and} \bibinfo{person}{Jiaya Jia}.}
  \bibinfo{year}{2012}\natexlab{}.
\newblock \showarticletitle{Structure extraction from texture via relative
  total variation}.
\newblock \bibinfo{journal}{\emph{ACM Transactions on Graphics}}
  \bibinfo{volume}{31}, \bibinfo{number}{6} (\bibinfo{year}{2012}),
  \bibinfo{pages}{1--10}.
\newblock


\bibitem[Zhang et~al\mbox{.}(2015)]%
        {zhang2015segment}
\bibfield{author}{\bibinfo{person}{Feihu Zhang}, \bibinfo{person}{Longquan
  Dai}, \bibinfo{person}{Shiming Xiang}, {and} \bibinfo{person}{Xiaopeng
  Zhang}.} \bibinfo{year}{2015}\natexlab{}.
\newblock \showarticletitle{Segment graph based image filtering: fast
  structure-preserving smoothing}. In \bibinfo{booktitle}{\emph{ICCV}}.
  \bibinfo{pages}{361--369}.
\newblock


\bibitem[Zhang et~al\mbox{.}(2014)]%
        {zhang2014rolling}
\bibfield{author}{\bibinfo{person}{Qi Zhang}, \bibinfo{person}{Xiaoyong Shen},
  \bibinfo{person}{Li Xu}, {and} \bibinfo{person}{Jiaya Jia}.}
  \bibinfo{year}{2014}\natexlab{}.
\newblock \showarticletitle{Rolling guidance filter}. In
  \bibinfo{booktitle}{\emph{ECCV}}. \bibinfo{pages}{815--830}.
\newblock


\bibitem[Zhu et~al\mbox{.}(2016)]%
        {zhu2016non}
\bibfield{author}{\bibinfo{person}{Lei Zhu}, \bibinfo{person}{Chi-Wing Fu},
  \bibinfo{person}{Yueming Jin}, \bibinfo{person}{Mingqiang Wei},
  \bibinfo{person}{Jing Qin}, {and} \bibinfo{person}{Pheng-Ann Heng}.}
  \bibinfo{year}{2016}\natexlab{}.
\newblock \showarticletitle{Non-Local Sparse and Low-Rank Regularization for
  Structure-Preserving Image Smoothing}.
\newblock \bibinfo{journal}{\emph{Computer Graphics Forum}}
  \bibinfo{volume}{35}, \bibinfo{number}{7} (\bibinfo{year}{2016}),
  \bibinfo{pages}{217--226}.
\newblock


\bibitem[Zontak et~al\mbox{.}(2013)]%
        {zontak2013separating}
\bibfield{author}{\bibinfo{person}{Maria Zontak}, \bibinfo{person}{Inbar
  Mosseri}, {and} \bibinfo{person}{Michal Irani}.}
  \bibinfo{year}{2013}\natexlab{}.
\newblock \showarticletitle{Separating signal from noise using patch recurrence
  across scales}. In \bibinfo{booktitle}{\emph{CVPR}}.
  \bibinfo{pages}{1195--1202}.
\newblock


\end{thebibliography}



\end{document}